\documentclass{article} 

\usepackage[dvipsnames]{xcolor}
\usepackage{iclr2025_conference,times}

\usepackage{color}

\usepackage{hhline}
\usepackage{amsmath,amsfonts,bm}

\usepackage{colortbl}
\usepackage{array}
\usepackage{graphicx}
\usepackage{url}
\newcommand{\PP}[1]{}
\newcommand{\shyam}[1]{}
\newcommand{\sv}[1]{}
\newcommand{\YSR}[1]{\textcolor{blue}{YsR:#1}}

\title{
LR0.FM: Low-Res Benchmark and  Improving Robustness for Zero-Shot Classification in Foundation Models
}

\author{Priyank Pathak\textsuperscript{1}, Shyam Marjit\textsuperscript{2}, Shruti Vyas\textsuperscript{1} \& Yogesh S Rawat\textsuperscript{1}  \\
\textsuperscript{1}University of Central Florida, \hspace{20pt}\textsuperscript{2}IIIT  Guwahati \\
\texttt{
priyank@ucf.edu,
shyam.marjit@iiitg.ac.in, 
\{shruti, yogesh\}@ucf.edu
}
\vspace{-0.1cm}
}

\usepackage{listings}
\usepackage{hyperref}
\usepackage{makecell}
\usepackage{booktabs}
\usepackage{cleveref}
\usepackage{array}
\usepackage{float}
\usepackage{subcaption}
\usepackage{amssymb}

\newcolumntype{P}[1]{>{\centering\arraybackslash}p{#1}}
 \newcommand*{\x}
 {\!\times\!}
 
\usepackage{multirow}

\definecolor{ao(english)}{rgb}{0.0, 0.5, 0.0}

\newcommand{\STEXT}[1]{{\small{{\text{#1}}}}}

\newcommand{\MTEXT}[1]{ \text{\normalsize{#1}}
}

\usepackage{adjustbox}

\newcommand{\Supp}[0]{\textit{Supplementary}}
\newcommand{\eg}[0]{\textit{e.g.} }
\newcommand{\etc}[0]{\textit{etc.} }
\newcommand{\ie}[0]{\textit{i.e.} }

\definecolor{mygray}{gray}{.9}

\newlength\savewidth
\newcommand{\SPCITE}[1]{{\textcolor{lightgray}{\small{[\citeyear{#1}]}}}}

\definecolor{mypink}{RGB}{255, 105, 180}

\hypersetup{
    colorlinks=true,
    linkcolor=blue,
    urlcolor=mypink,
    citecolor=gray
}

\iclrfinalcopy 
\begin{document}

\maketitle
\vspace{-30pt}
\begin{center}
{
\normalsize
\href{https://ucf-crcv.github.io/lr0.fm}{https://ucf-crcv.github.io/lr0.fm}
}
\end{center}    
\vspace{-2pt}

\begin{figure}[ht]
  \centering
\includegraphics[width=0.92\linewidth]{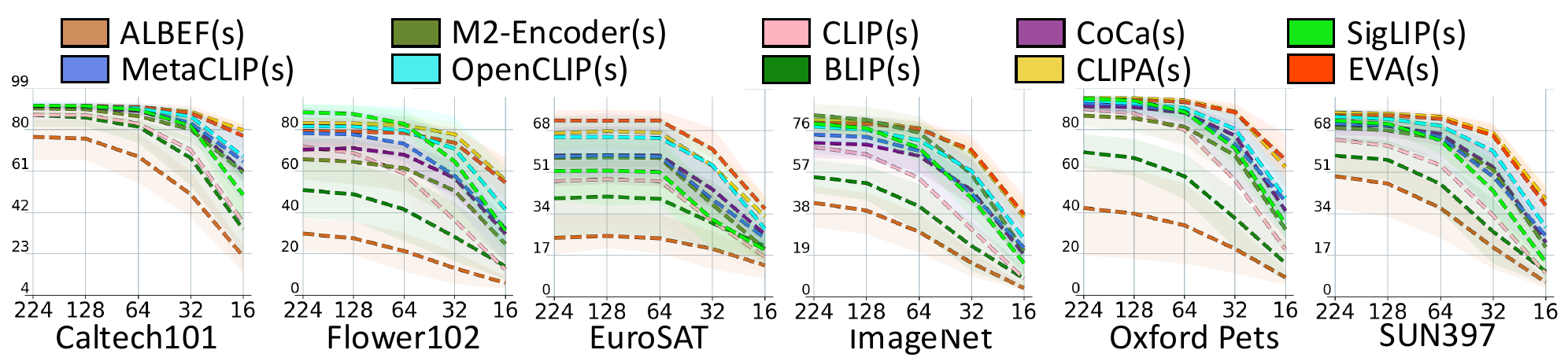}
  \caption{\textbf{Top-1 zero-shot classification accuracy (y-axis) vs resolution (x-axis)}: 
Backbones for foundation models are merged as shade, with average performance across backbones in the dark.
}
\vspace{-3pt}  
\label{fig:acc_drop}
\vspace{-0.2pt}  
\end{figure}

\begin{abstract}
Visual-language foundation Models (FMs) exhibit remarkable zero-shot generalization across diverse tasks, largely attributed to extensive pre-training on large-scale datasets. 
However, their robustness on low-resolution/pixelated (LR) images, a common challenge in real-world scenarios, remains underexplored. 
We introduce \textbf{LR0.FM}, a comprehensive benchmark evaluating the impact of low resolution on the zero-shot classification performance of \textit{10} FM(s) across \textit{66} backbones and \textit{15} datasets. 
We propose a novel metric, \textbf{Weighted Aggregated Robustness}, to address the limitations of existing metrics and better evaluate model performance across resolutions and datasets.
Our key findings show that: 
(i) model size positively correlates with robustness to resolution degradation, 
(ii) pre-training dataset quality is more important than its size, and 
(iii) fine-tuned and higher resolution models are less robust against LR. 
Our analysis further reveals that the model makes semantically reasonable predictions at LR, and the lack of fine-grained details in input adversely impacts the model's initial layers more than the deeper layers. We use these insights and introduce a simple strategy, \textbf{LR-TK0}, to enhance the robustness of models without compromising their pre-trained weights. We demonstrate the effectiveness of \textbf{LR-TK0} for robustness against low-resolution across several datasets and its generalization capability across backbones and other approaches. \textit{Code is available at this \href{https://github.com/shyammarjit/LR0.FM}{link.}} 
\end{abstract}
\vspace{-5pt}

\section{Introduction}

Vision-Language Foundation Models (FMs), such as CLIP~\citep{radford2021learning}, LLaMA~\citep{touvron2023llama}, and other variants, have shown extraordinary generalization capabilities across a wide range of downstream tasks, including image classification~\citep{ilharco_gabriel_2021_5143773}, object detection~\citep{zhong2022regionclip}, and semantic segmentation~\citep{xu2022odise}. These models benefit from large-scale, multi-modal pre-training on diverse datasets like DataComp-1B~\citep{datacomp} and LAION-5B~\citep{schuhmann2022laion}, 
enabling them with zero-shot capabilities.
Although these models excel on high-resolution benchmarks, their performance with low-resolution (LR) pixelated images, a common real-world challenge, remains adequately underexplored.

Low-resolution images frequently arise in various practical scenarios, such as surveillance footage~\citep{davila2023mevid}, satellite imagery~\citep{Patil2017ClassificationOL}, and privacy-protected pixelated data~\citep{10.1145/3394171.3413972} \textit{etc}. In these cases, details crucial for accurate classification may be obscured by artifacts like pixelation and compression, leading to substantial performance degradation. 
For instance, small objects (faces) within larger images~\citep{cheng2019low} pose unique challenges, often requiring models to rely on limited visual cues. Given the widespread presence of LR images in real-world applications, it is crucial to understand how robust FMs are in these settings.

\begin{figure}[!tb]
\centering
\subfloat
{
\includegraphics[height=4.12cm]{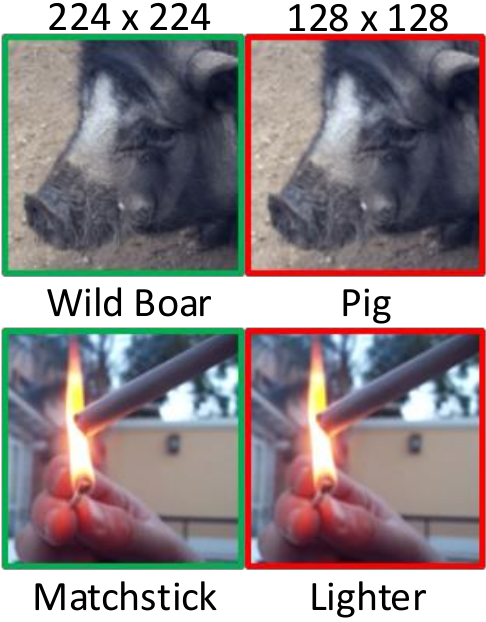}
\label{fig:128_ex}
}
\hfill
\subfloat
{
\includegraphics[height=4.12cm]{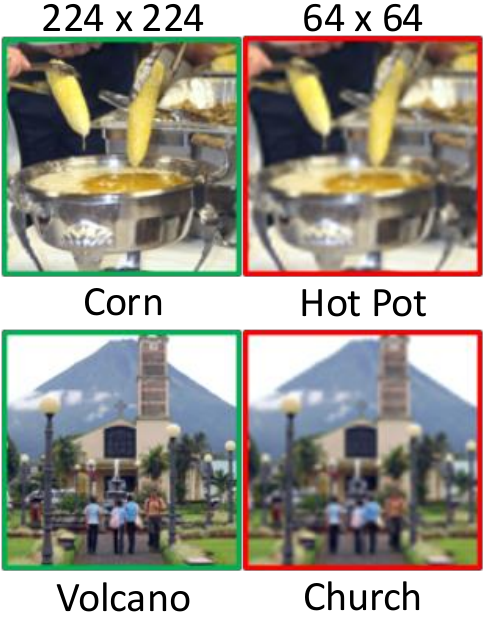}
\label{fig:32_ex}
}
\hfill
\subfloat
{
\includegraphics[height=4.12cm]{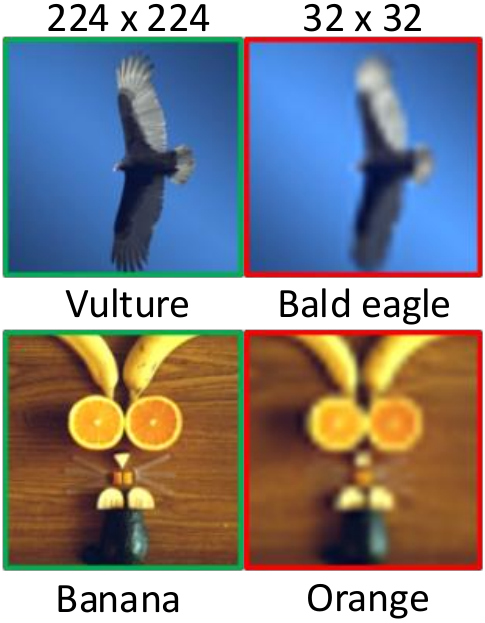}
\label{fig:32_ex}
}
\hfill
\subfloat
{
\includegraphics[height=4.12cm]{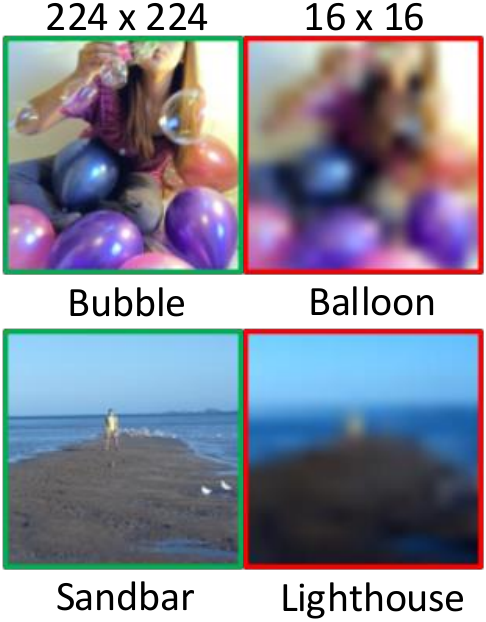}
\label{fig:32_ex}
}
\vspace{-2pt}
\caption{\textbf{Zero-Shot misclassifications:} 
EVA-CLIP \SPCITE{sun2023eva} correct classification at $224\x224$ (green) \& misclassification at lower resolution (red). However, ImageNet labels-based mispredictions are semantically reasonable (humans), indicating viability of pre-trained weights at low resolution.
}
\label{fig:img_ex}
\vspace{-9pt}
\end{figure}
Motivated by this, we present an in-depth benchmarking study of FMs, focusing on their zero-shot classification performance under LR conditions. 
We introduce LR0.FM, a comprehensive benchmark that evaluates \textbf{10 foundation models} across \textbf{66 backbones} and \textbf{15} diverse image classification \textbf{datasets}, ranging from large-scale datasets like ImageNet~\citep{deng2009imagenet} to fine-grained and texture-specific datasets like 
Oxford Pets~\citep{parkhi2012cats} and DTD~\citep{cimpoi2014describing}. 
Our study systematically examines the effects of resolution degradation, revealing key insights into how model size, pre-training dataset quality, and fine-tuning impact robustness in LR scenarios.

Metrics for measuring  robustness ($\gamma$, ~\cite{schiappa2024robustness}) and its averaging across datasets (SAR)
have some limitations; 1) They can produce misleadingly high scores when models perform poorly on challenging datasets, and 2) They tend to ignore certain datasets, skewing the overall comparison. To address these, we propose a new metric, \textbf{Weighted Aggregated Robustness (WAR)}, which provides a more balanced evaluation by considering performance drops across datasets more fairly. 

Our analysis reveals several interesting insights. Larger models tend to maintain robustness better when faced with LR inputs, while the quality of the pre-training dataset is more crucial than its size in preserving performance. 
Furthermore, fine-tuned models and those with higher-resolution inputs significantly underperform against resolution drop. 
We also observe that although models struggle at low-resolution 
(\cref{fig:acc_drop}) and loss of fine-grained details (\cref{fig:img_ex}: \eg Vulture vs Bald Eagle, Bubble vs Balloon \textit{etc.}), their predictions often remain semantically reasonable, even at extreme resolutions (\cref{fig:img_ex}: \eg Orange vs Banana, Church vs Volcano \textit{etc.}). 
\textit{Supplementary} demonstrates more examples (including real-world) where such mispredictions are made.    
This suggests a solution for low-resolution does not require extensive modifications to the model and its pre-trained weights. 

Based on these insights, we propose a simple yet effective solution, \textbf{LR-TK0: LR-Zero-Shot Tokens}, which introduces low-resolution-specific tokens to enhance robustness without altering the pre-trained model weights. Our method preserves the model's semantic reasoning capabilities while compensating for the loss of fine-grained detail, offering a feature super-resolution-like approach ~\citep{chen2024robustsam}. By training on synthetic diffusion-based high-resolution images, LR-TK0 improves performance in low-resolution zero-shot classification tasks, making FMs more robust for practical, real-world applications.


In summary, we make the following contributions in this work,
\vspace{-8pt}
\begin{enumerate}
\itemsep-0.1em 
    \item We present \textbf{\textit{LR0.FM}}, a comprehensive benchmarking of Visual-Language Foundation Models (FMs) on zero-shot classification of low-resolution images, providing several key insights. To the best of our knowledge, no prior work has explored this aspect of FMs.
    \item We introduce a simple and effective method, \textbf{\textit{LR-TK0}}, to enhance model robustness against low-resolution inputs without altering the pre-trained weights.
    \item We introduce Weighted Aggregated Robustness (\textbf{WAR}), a novel robustness metric for evaluating models under challenging conditions, overcoming the limitations of existing metrics. 
\end{enumerate}

\section{Related Works}
\label{sec:related_works}
\noindent \textbf{Foundation Models (FM):} 
Large-scale models~\citep{kirillov2023segment, girdhar2023imagebind}, pre-trained on massive datasets, demonstrate generalization across numerous downstream tasks. 
For example, CLIP~\citep{radford2021learning} embeds $\!\sim\!400$ million image-text pairs in a shared feature space for zero-shot image classification and image-text retrieval. It is also effective in other domains like video-text retrieval~\citep{luo2022clip4clip}, and video and audio understanding~\citep{lin2022frozen, guzhov2022audioclip}.
Joint vision-text learning has also succeeded in tasks such as self-supervision~\citep{ miech2020end}, few-shot~\citep{alayrac2022flamingo}, multi-modal retrieval~\citep{yu2022coca} \etc
However, the robustness of these models against \textit{real-world challenges} \eg harmful images~\citep{qu2024unsafebench}, image quality~\citep{wu2023q}, text quality~\citep{Xu2024}, \etc requires further exploration. \vspace{3pt}\\
\noindent \textbf{Zero Shot:} Zero Shot/Open-set/In-the-wild image classification predicts an unseen class by matching the image with labels~\citep{sun2023eva}.
In the past, traditional models have been tested for their zero-shot capabilities~\citep{chao2016empirical, xian2017zero}, however, FMs are better suited for this task. 
Benchmarking their zero-shot capabilities is a relatively newer area of research~\citep{schiappa2023probing, Schulter_2023_ICCV}.
To assess the performance comprehensively, 
we have expanded the pool of models from traditional 10-11 FM backbones \eg 4 backbones \citep{NEURIPS2022_3c4688b6}, 9 backbones \citep{liu2024few}, 6 backbones (\citep{zhang2024progressive}) \etc to \textit{66 backbones}. \vspace{3pt}\\
\noindent \textbf{Low Resolution (LR):}
LR images are captured in various practical scenarios and are sometimes used intentionally for computational cost reduction (RECLIP~\citep{li2023reclip}). LR benchmarks mostly focus on face recognition~\citep{9432821, li2018face}, with some work in zero-shot/unconstrained recognition~\citep{8600370, cheng2019low}. 
Super Resolution~\citep{ohtani2024rethinking, gao2023implicit} are often domain-specific or restores only $\ge64\x64$. 
However, there is a lack of study on the robustness of FM(s) against real-world challenges~\citep{Xu2024}, with no previous work on very LR. 
We benchmark FM(s) against LR images and propose a lightweight solution for improving robustness, without training on any of the target datasets~\citep{chen2024robustsam}.
\vspace{-7pt}
\begin{table}[!t]
\centering
\caption{\textbf{Benchmark Models (66 Backbones):}
Pre-training is image-text pairs from datasets like DataComp-1B (DC-1B) \citep{datacomp}, Conceptual Captions (CC) \citep{sharma2018conceptual}, Conceptual 12M (C-12M) \citep{changpinyo2021conceptual}.
Text Encoders are mostly modified vanilla transformers (Tran.)\citep{NIPS2017_3f5ee243}.
Vision backbones use (modified) ViTs~\citep{dosovitskiy2020image}.
\vspace{-4pt}
}
\renewcommand{\arraystretch}{1.2}
\label{tab:model_desc}  
\scalebox{0.9}{
\centering
\setlength\tabcolsep{2pt}
\begin{tabular}{p{2.7cm}|
P{2cm}| P{5.8cm}| P{1.7cm}| P{2.3cm}
}
\specialrule{1pt}{0pt}{0pt}
\rowcolor{mygray} 
Models & \#Backbones & 
\multicolumn{2}{P{7.64cm}|}{Pre-training (Dataset / Size Billion:B \& Million:M)}  
& Text Encoder \\
\hline\hline
CLIP \SPCITE{radford2021learning}
& 4 ViTs \& 5 ResNets & WIT-400M 
\SPCITE{radford2021learning} & 400M &  Tran. \SPCITE{radford2019language} \\ 
\hline 
OpenCLIP \SPCITE{ilharco_gabriel_2021_5143773}
& 8 ViTs & \STEXT{DC-1B}, LAION-2B\SPCITE{schuhmann2022laion}, DFN-5B\SPCITE{fang2023data} & 1B-5B & Tran.\SPCITE{ ilharco_gabriel_2021_5143773} \\ 
\hline 
MetaCLIP \SPCITE{xu2023metaclip} & 8 ViTs & \textit{Self} & 400M-2.5B & OpenCLIP \\ 
\hline 
CLIPA \hspace{-0.6cm} (v1\&v2)  
\SPCITE{li2023clipa, li2023clipav2} & 7 ViTs & DC-1B, LAION-2B \SPCITE{schuhmann2022laion} & 1B-2B & Autoregressive Tran. \SPCITE{NIPS2017_3f5ee243} \\
\hline 
SigLIP
\SPCITE{Zhai_2023_ICCV} & 8 ViTs & WebLI \SPCITE{chen2022pali} & 10B & Tran.\\
\hline 
CoCa
\SPCITE{yu2022coca} & 3 ViTs & LAION-2B \SPCITE{schuhmann2022laion}, 
COCO \SPCITE{lin2014microsoft} & 2B & Tran. Decoder \\ 
\hline 
M\textsuperscript{2}-Encoder\SPCITE{guo2024m2encoder}& 
3M\textsuperscript{2}-Encoder & BM-6B \SPCITE{guo2024m2encoder} & 6B & Magneto \SPCITE{Magneto}\\
\hline 
ALBEF \SPCITE{ALBEF} & 4 ALBEF (ViT) & COCO \SPCITE{lin2014microsoft}, Visual Genome \SPCITE{krishna2017visual}, CC, SBU Captions \SPCITE{ordonez2011im2text}, C-12M & 4M-14M & BERT \SPCITE{Devlin2019BERTPO} \\
\hline 
BLIP \SPCITE{li2022blip} & 8 ViTs & ALBEF \SPCITE{ALBEF}, LAION-400M \SPCITE{schuhmann2021laion} & 14M-129M & BERT \SPCITE{Devlin2019BERTPO}\\
\hline
EVA-CLIP(\&18B) \SPCITE{sun2023eva, EVA-CLIP-18B}
& 8 EVA(s) (ViT(s)) & 
LAION-400M\SPCITE{schuhmann2021laion}, LAION-2B\SPCITE{schuhmann2022laion}, Merged-2B \SPCITE{sun2023eva} & 400M-2B & OpenCLIP\\
\specialrule{1pt}{0pt}{0pt}
\end{tabular}
}
\vspace{-6pt}
\end{table}

\vspace{-3pt}
\section{Benchmarking Setup}
\vspace{-2pt}
\noindent \textbf{Model:}
\Cref{tab:model_desc} lists all 10 Foundation models used in our benchmarking\footnote{ 
EVA-CLIP~\citep{sun2023eva} \& EVA-CLIP-18B~\citep{EVA-CLIP-18B} merged into one.}.
CLIP, OpenCLIP, MetaCLIP, CLIPA, and SigLIP use the same ViT model with different pre-training datasets and slight architectural modifications (\eg layer norm position, token masking \textit{etc.}).
M\textsuperscript{2}-Encoder (built on top of CoCa), ALBEF, and BLIP use modified cross attention between text and vision transformers.  
EVA-CLIP is a family of models equipped with recent advancements \eg architectural modifications, token dropping, training via distillation \etc surpassing all existing works. 
Backbones are referred to using their publicly available pre-trained weights, \eg CLIP-ViT L (400M), which means: CLIP model ViT-L architecture, pre-trained on 400 million datasets. `B' would indicate a billion. \vspace{4pt}\\  
\noindent \textbf{Dataset:}
\begin{figure}[!tb]
\centering
\subfloat
{
\includegraphics[height=2.9cm]{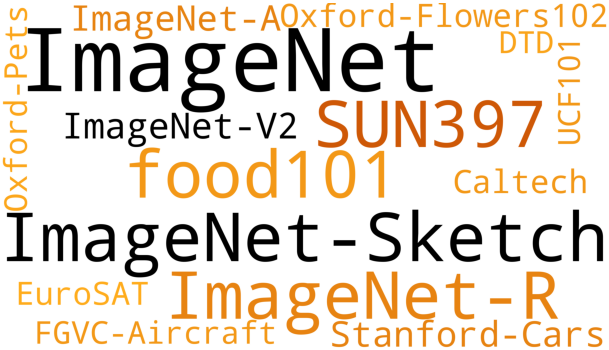}
}
\hfill 
\subfloat
{
\includegraphics[height=2.9cm]{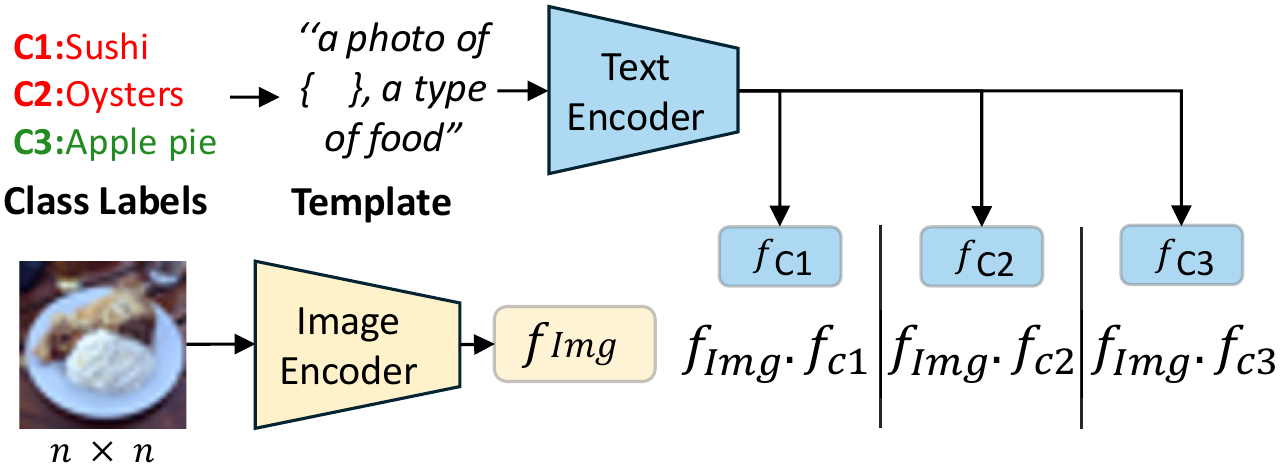}
}
\vspace{-2pt}
\caption{ 
\emph{Left}: \textbf{Dataset}: Size $\propto\log$ \# test images, and color gradient $\propto$ \# of test classes 
{\color{orange}\textbf{orange}} is 10 \& {\color{black}\textbf{black}} is 1000 classes). 
\emph{Right}: \textbf{Zero Shot Evaluation}: 
Food-101 image ($32\x32$) generates image embeddings $f_{Img}$, while class labels are filled in templates (1 shown) generating text embeddings (averaged across templates).
The dot product of $f_{Img}$ with text features gives classification logits.
}
\vspace{-7pt}
\label{fig:eval_pipeline}
\label{fig:dataset}
\label{fig:dataset_eval_pipeline}
\end{figure}
\Cref{fig:eval_pipeline} (left) highlights benchmarking datasets size and the number of classes for: 
Imagenet \SPCITE{deng2009imagenet}, 
ImageNet-A \SPCITE{hendrycks2021nae}, 
ImageNet-V2 \SPCITE{recht2019imagenet}, 
ImageNet-R \SPCITE{hendrycks2021many}, 
ImageNet-Sketch (ImageNet-SK) \SPCITE{wang2019learning}, 
Caltech101 \SPCITE{griffin2007caltech}, 
DTD split-1 (DTD) \SPCITE{cimpoi2014describing}, 
Food101 \SPCITE{bossard2014food}, 
SUN397 \SPCITE{zhou2014learning}
Stanford Cars (Cars) \SPCITE{kramberger2020lsun}, 
FGVC Aircraft (Aircraft) \SPCITE{maji2013fine},
Oxford Pets (Pets) \SPCITE{parkhi2012cats}, 
Oxford Flowers102 (Flowers102) \SPCITE{liu2016flower}, 
EuroSAT \SPCITE{helber2019eurosat}, 
UCF101 \SPCITE{soomro2012ucf101}. 
Details in \Supp. \vspace{4pt}\\         
\noindent\textbf{Zero-Shot Image Classification}
We adopt CLIP~\citep{radford2021learning} evaluation protocol for all the models as shown in \cref{fig:eval_pipeline} (right).
Image encoder generates embeddings for images, while test labels are used with dataset-specific templates (multiple templates, \Supp) \eg ``a photo of a [label]". Model’s Text encoder generates final text embeddings (averaged across all templates) for the class label.
The dot product of visual and text embeddings produces class logits, with the highest logit score determining the predicted class. 
Accuracy is computed using Top-1 match. \vspace{4pt} \\ 
\noindent \textbf{Low Resolution:}
Models are evaluated on their pre-trained resolution, namely $224\x224$ $256\x256$, $378\x378$ \etc 
Low resolution is simulated by downsampling HR images to $16\x16$, $32\x32$, $64\x64$, and $128\x128$ using bicubic interpolation, followed by model specific preprocessing similar to their HR counterparts, \eg resizing to $224\x224$, center crop, \etc 
Performance degradation starts below $64\x64$ (\cref{fig:acc_drop}), so we focus mainly on $16\x16$ and $32\x32$.
This downsampling mimics pixelation as seen in low-resolution cameras (\eg self-driving cars) and distant images (\eg CCTV), \etc \vspace{4pt}\\
\noindent \textbf{Evaluation Metrics}:
\label{sec:eval_metrics}
We represent top-1 accuracy on the dataset `$D$' with a resolution \underline{$n\x n$} as $A^{D}_{n}\in [0,1]$, \eg HR accuracy $A_{HR}^{D}$ $\ge A^{D}_{n}$ (LR accuracy), where HR is model specific $\in$ \{224, 256, 372, 384, 512\}. 
Top-1 scores averaged across datasets is \textbf{ACC-n}. 
Robustness against artifacts~\citep{schiappa2024robustness} is measured by relative robustness 
($\gamma_{n}^{D}=1-(A^{D}_{HR}- A^{D}_{n})/A^{D}_{HR}$).
$\gamma_{n}^{D}$ is dataset-specific, and it is common to average scores across datasets for model comparison, denoted by \textbf{Simple Aggregated Robustness (SAR-n)}.
\textit{Higher number indicates more robustness}.
However, there are two significant issues with $\gamma_{n}^{D}$ and SAR-n:

\textbf{\textit{Problem A) Misleading high robustness:}} If the model performs poorly on a challenging dataset \ie performance close to random predictions, then downsampling will likely maintain this random prediction with minimal drop in accuracy, giving abnormally high robustness score. \textit{Ex.} `ALBEF (4M)' for Aircraft dataset, ($A_{\text{rand}}^{\text{aircraft}}\!=\!1\%$), $A_{HR}^{\text{aircraft}}\!=\!2.7\%, A_{16}^{\text{aircraft}}\!=\!1\%$,
\underline{$\gamma_{16}^{\text{aircraft}}\!=\!37\%$}, \ie random predictions yields $\!\sim\!40\%$ robustness ($40\%$ robustness is among the highest, more in \Supp).

\begin{figure}[!t]
\centering 
\subfloat
{
\includegraphics[height=2.8cm]{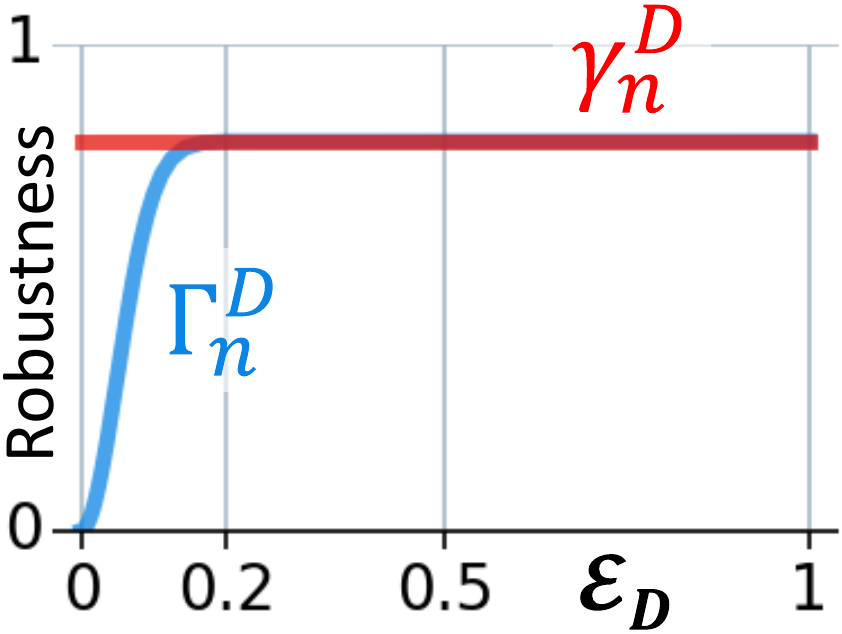}
}
\hfill
\subfloat
{
\includegraphics[height=2.8cm]{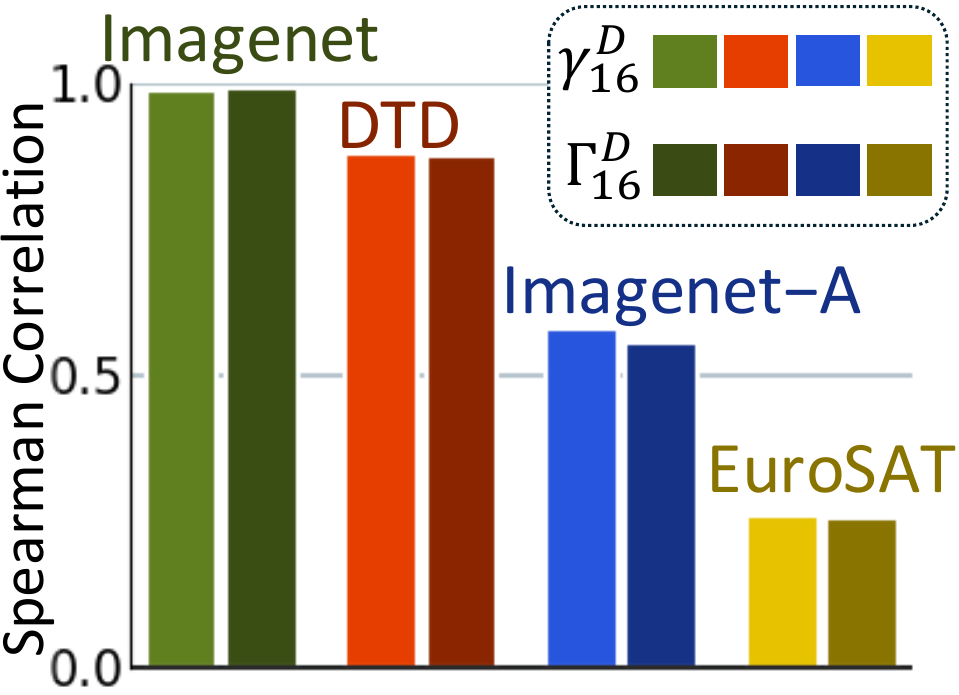}
}
\hfill
\subfloat
{
\includegraphics[height=2.8cm]{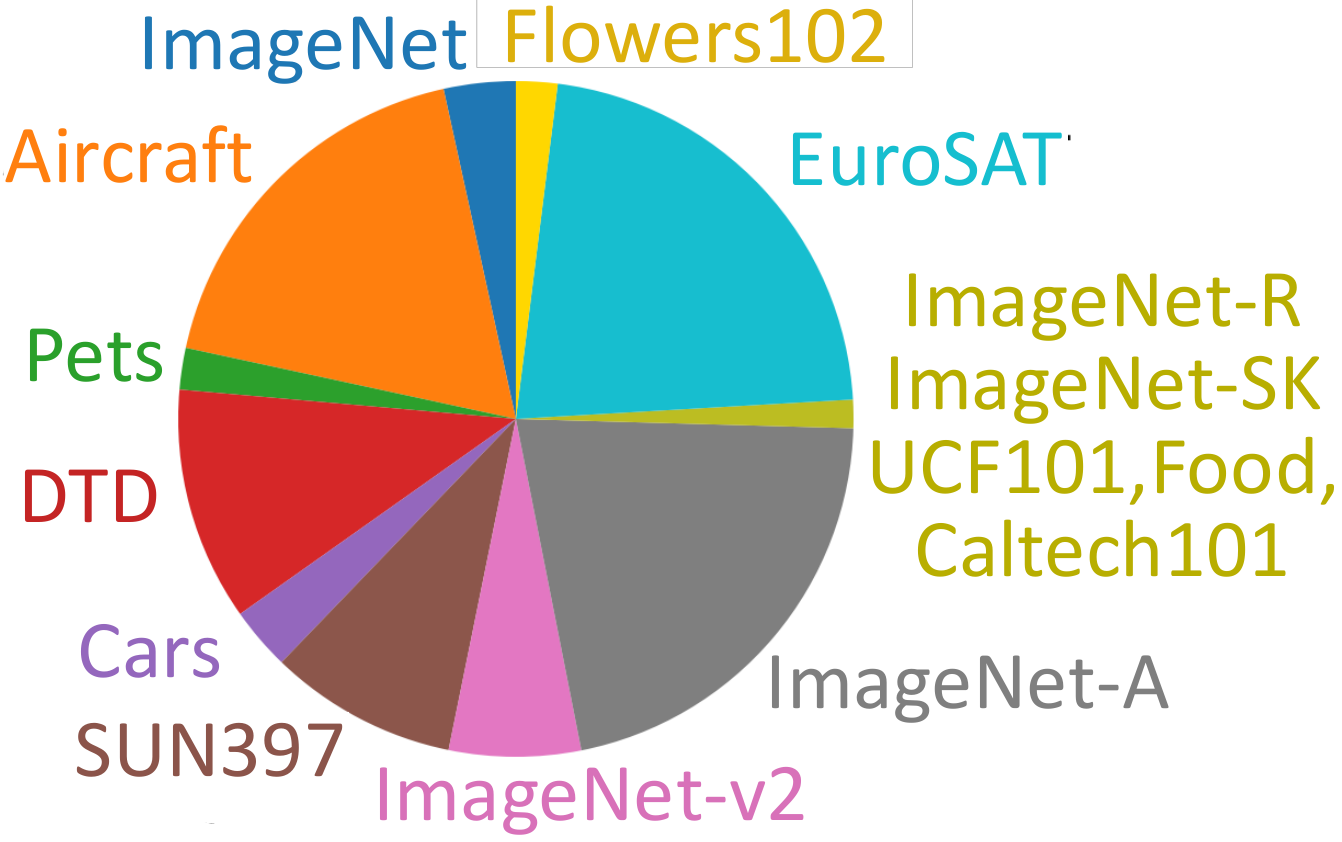}
}
\vspace*{-3pt}
\caption{ 
\textit{Left}: {\textbf{Improved} 
 \boldmath{$\Gamma^{D}_{n}$}} \textbf{vs traditional} {\boldmath{$\gamma^{D}_{n}$}}:
$\!\Gamma^{D}_{n}\!\approx\!\gamma^{D}_{n}\!$ except near random predictions ($\mathcal{E}_D\!\rightarrow\!0$). 
\textit{Mid}: \textbf{Correlation} 
between the ordering of models after averaging of robustness (SAR) across datasets ($\gamma^{D}_{16}$ \& $\Gamma^{D}_{16}$) with dataset's true ordering. 
SAR final ranking ignores datasets like EuroSAT (0.26). 
\textit{Right}: \textbf{Optimized dataset weights} for WAR-16. \textit{Supplementary} contains numeric value. 
}
\label{fig:datasets_weights}
\label{fig:corr_both_gamma}
\label{fig:new_gamma_old_gamma}
\label{fig:hyperparam_war}
\vspace{-3pt}
\end{figure}

\textbf{\textit{Solution: Improved Relative Robustness $\Gamma^{D}_{n}$}:} 
A naive solution is to calculate relative robustness only for correct predictions at the HR resolution. 
However, tracking predictions for each model across all datasets might not be scalable, especially if the dataset contains millions of images. We propose \textit{zero-ing out robustness near random predictions}. We first define \textit{accuracy gap} for the model on a dataset with `C’ classes as \underline{$\boldmath{\mathcal{E}_D\!=\!A^{D}_{HR}\!-\!A^{D}_{rand}}$}, with $\mathcal{E}_D\!\in\![0,1]$, and 
$A^{D}_{rand}\!=\!1/C$ represents random prediction accuracy\footnote{Random guessing one of the `C' class yields $1/C$ accuracy, referred to as $A^{D}_{rand}$ in this work.}.
If $A^{D}_{HR}\!>>\!A^{D}_{rand}$, $\mathcal{E}_D$ will be high. Conversely, if $A^{D}_{HR} \simeq$ random prediction, $\mathcal{E}_D\!\rightarrow\!0$.
Using $\mathcal{E}_D$, we compute \textbf{improved relative robustness} $\Gamma^{D}_{n}$ as 
\vspace{-2pt}
\begin{align}
    \Gamma^{D}_{n} &= \gamma^{D}_{n} \times (1 -  e^{- \alpha (\mathcal{E}_D)^2}) \hspace{0.5cm} \mid \hspace{0.5cm} \alpha >> 1 \hspace{0.5cm} \& \hspace{0.5cm} 0 \le \mathcal{E}_D \le 1 \label{eq:improved_robsutness}
\end{align}
when $\mathcal{E}_D\!\sim\!0$ \ie near random predictions, $\Gamma^{D}_{n}\!\sim\!0$, otherwise $\Gamma^{D}_{n} \approx \gamma^{D}_{n}$, as shown in \cref{fig:new_gamma_old_gamma} (left). 
Hyperparameter
$\alpha$  
is the rate at which 
$\Gamma^{D}_{n}$  declines as accuracy approaches random prediction.
We chose $\alpha=200$ as a middle between 100 (the drop at $\mathcal{E}_D\sim0.2$) and 500 (the drop at $\mathcal{E}_D\sim 0$).

\textbf{\textit{Problem B) SAR overlooks datasets:}} 
When comparing models, their robustness scores are averaged across datasets (SAR). 
Ideally, model ranking, after averaging robustness across datasets, 
should stay consistent with
their rankings on individual datasets.
However, \cref{fig:corr_both_gamma} (mid) shows the rankings of 66 models after averaging correlate (Spearman Rank) highly with ImageNet (0.99) and DTD (0.88), but only moderately with ImageNet-A (0.56) and weakly / not with EuroSAT (0.26). 
Most datasets follow the ImageNet trend, influencing the final model rankings and minimizing the impact of datasets like ImageNet-A and EuroSAT (behave differently) as if these datasets aren't present.

\textbf{\textit{Solution: Weighted Aggregated Robustness:}}
Averaging the robustness scores gives each dataset score of 1. 
We propose adjusting the dataset weights so that the \textit{model rankings after aggregation reflect each dataset fairly} (\cref{fig:datasets_weights} (right)). 
Weights are optimized such that the correlation (Spearman) between the model rankings after the weighted average and individual dataset rankings are maximized. 
The weighted sum of robustness is:
\textbf{WAR-n}$=\sum_d^{\text{Datasets}} | \Gamma^{d}_{n} \x w^d_n |  / \sum_d^{\text{Datasets}} | w^d_n |$, where $w^d_n$ is dataset weight, and $\Gamma^{d}_{n}$ is dataset-specific improved robustness score for the resolution $n\x n$.

We use Ax tool~\citep{bakshy2018} for optimizing the weights of the dataset $w^d_{16}\!\in\![0.1,1]$ such that the Spearman correlation (SC) between the final model ranking obtained after the weighted averaging and individual dataset ranking is maximized on  
empirically found (more in \Supp):
\vspace{-0.4cm}
\begin{align}
0.95\x
\big(\MTEXT{SC(Imagenet)}\!+\!\MTEXT{SC(ImageNet-V2)}\!+\! \MTEXT{SC(DTD)}\big) + \MTEXT{SC(ImageNet-A)}\!+\!\MTEXT{SC(EuroSAT)}
\end{align} \vspace{-0.65cm}

Optimizing $w^d_{16}$ may give minimal weights to some datasets, thus WAR-n may not reflect the true robustness and is more apt for model comparisons, representing all the datasets.  
Hence we use both Weighted Aggregated Robustness (WAR) using improved relative robustness $\Gamma^{D}_{n}$ (\cref{eq:improved_robsutness}) and simple averaging (SAR) using traditional robustness $\gamma^{D}_{n}$ for evaluating models. 
\textit{Note}, \textbf{$\gamma^{D}_{n}$ and $\Gamma^{D}_{n}$ measure dataset robustness while SAR and WAR measure averaged robustness across the datasets}.

\begin{figure}[!t]
\centering
\subfloat
{
\includegraphics[height=4.25cm]{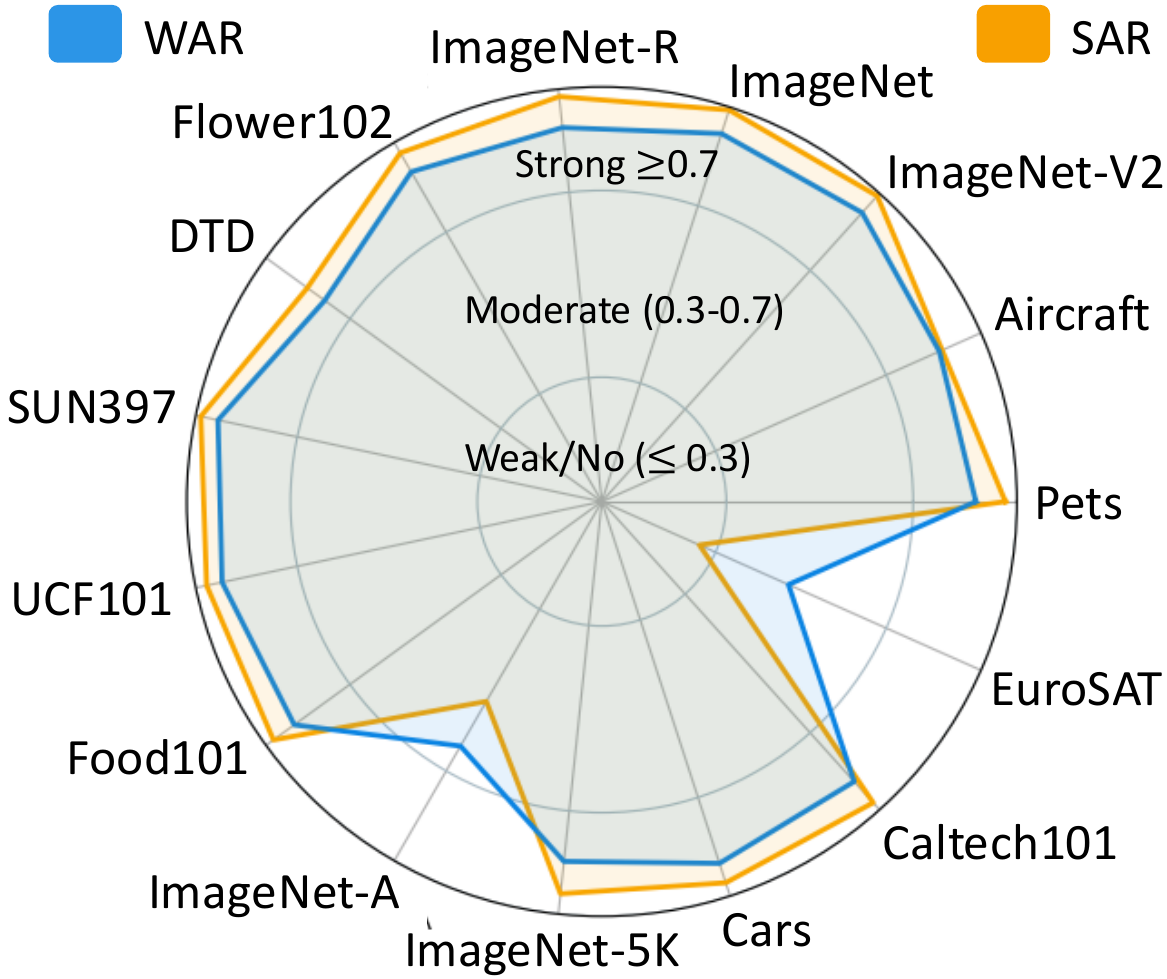} 
}
\hfill
\subfloat
{
\includegraphics[height=4.15cm]{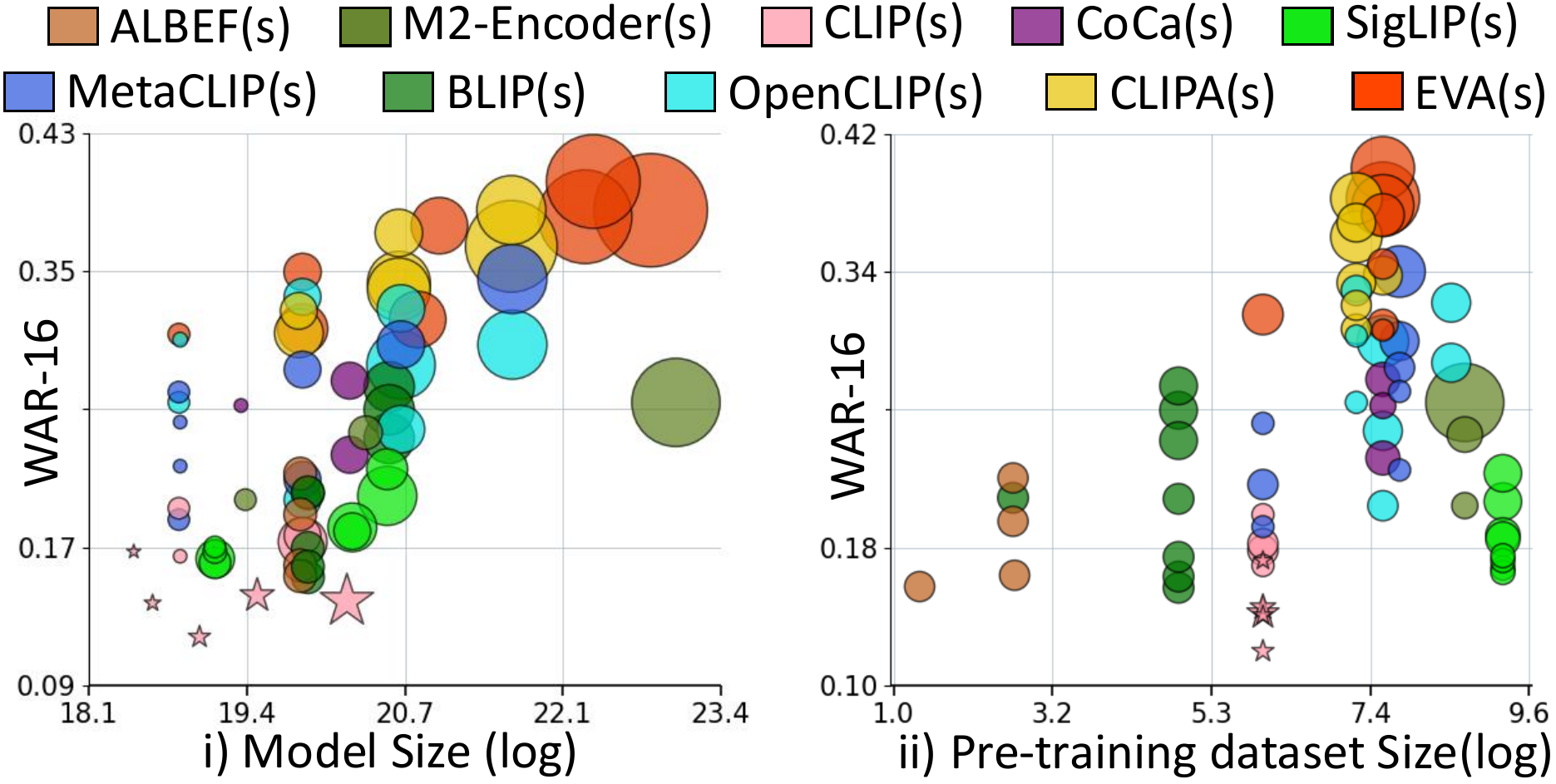}
}
\vspace{-3pt}
\caption{ 
Evaluations at $\!16\!\x\!16\!$. \textit{Left:} 
\textbf{SAR vs WAR:}
WAR improves the correlation (between the ordering of models after aggregation with individual datasets) for EuroSAT (0.26 $\rightarrow$ 0.49 and ImageNet-A ($0.56\rightarrow 0.68$), both computed via $\Gamma^{D}_{16}$.
\textit{Right}:
\textbf{i) Model Size \& ii) Pre-training dataset size} positively impacts robustness.
(i) Dot size $\propto$ GFLOPs, no impact on robustness (ii) Dot size $\propto$ Model Size, positively impact robustness. 
ResNets ($\star$), and transformers ($\bigcirc$).
}
\label{fig:robustness_metrics}
\label{fig:model_arch}
\label{fig:model_arch2}
\vspace{-7pt}
\end{figure}

\begin{figure}[!t]
\centering 
\subfloat
{\includegraphics[height=2.8cm]{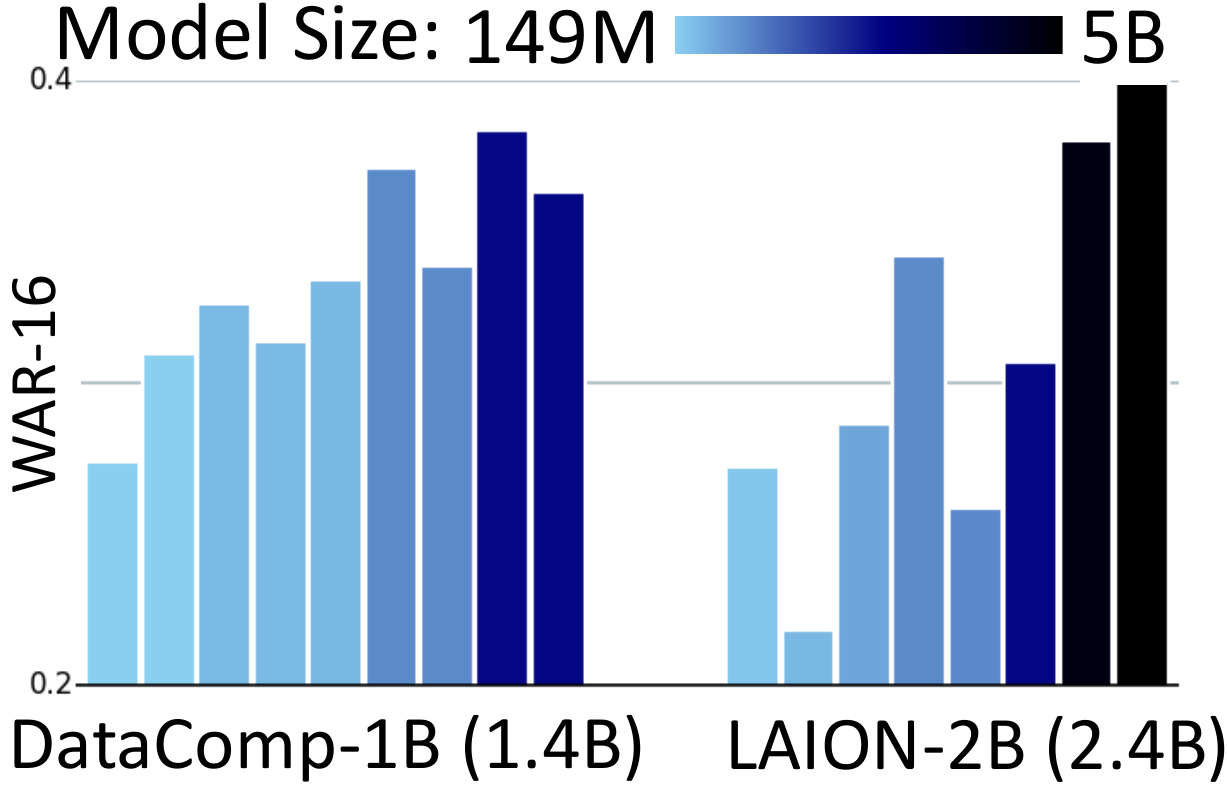}}
\hfill
\subfloat
{\includegraphics[height=2.8cm]{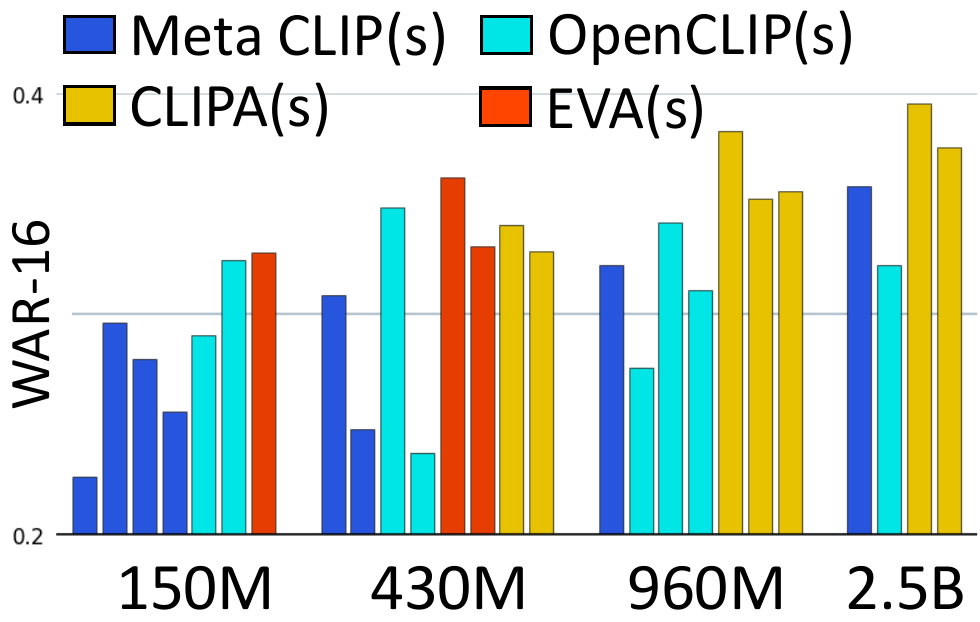}}
\hfill
\subfloat
{\includegraphics[height=2.8cm]{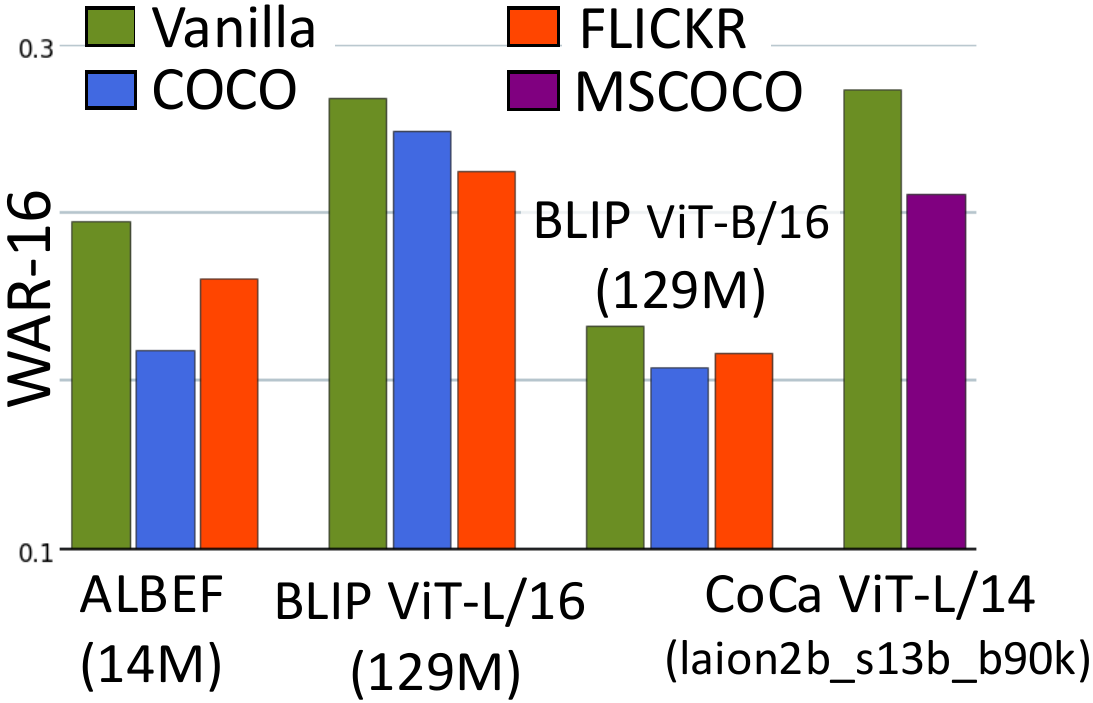}}
\vspace{-4pt}
\caption{ 
\textit{Left:} \textbf{DataComp-1B vs LAION-2B}: Smaller DataComp-1B pre-training helps robustness. Models are ordered via size.  
\textit{Mid:} \textbf{Model comparison w/o Size}: Models binned into size buckets ($\pm30$M).
\textit{Right:} \textbf{Fine-tuning degrades} robustness.  (\textit{left} \& \textit{mid}): bigger models are more robust.
}
\label{fig:model_size_wise_comparison} 
\label{fig:DC1B_L2B}
\label{fig:deepanalysis}
\vspace{-10pt}
\end{figure}
\begin{figure}[!t]
\centering 
\subfloat
{\includegraphics[height=3.2cm]{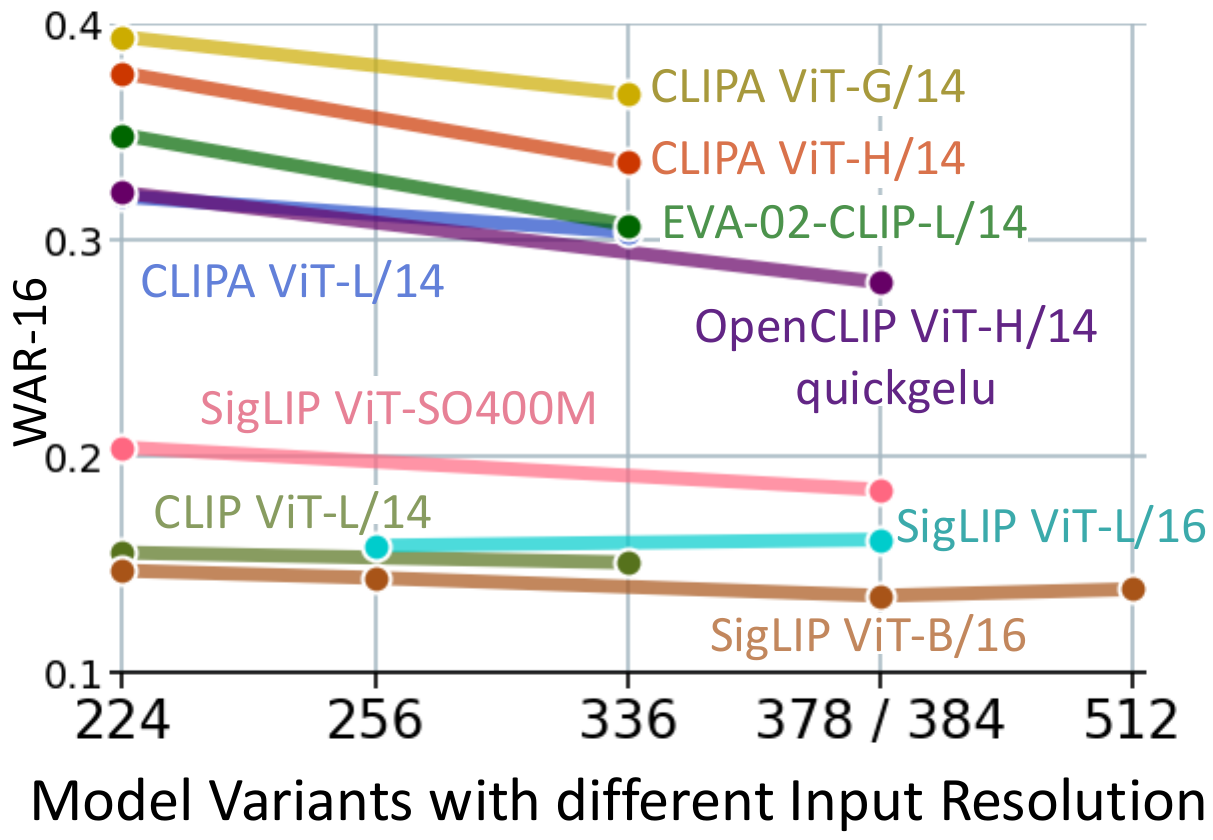}}
\hfill
\subfloat
{\includegraphics[height=3.2cm]{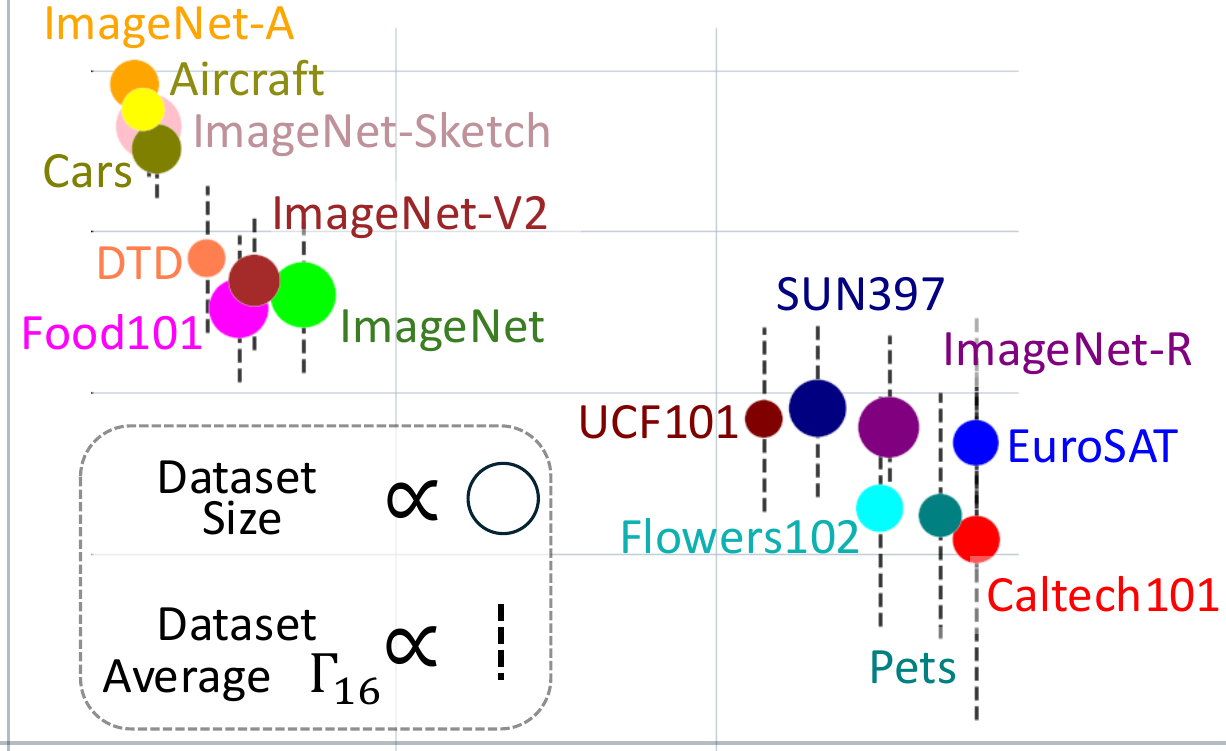}}
\hfill
\subfloat{\includegraphics[height=3.6cm]{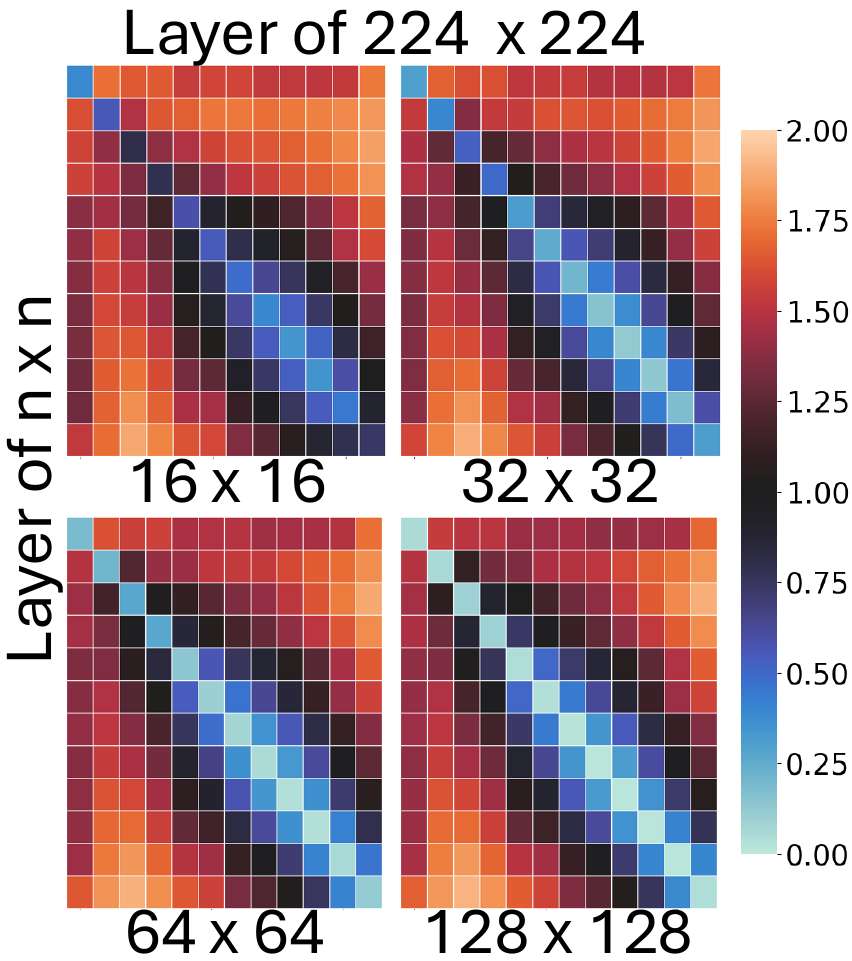}}
\vspace{-3pt}
\caption{ 
\textit{Left}: \textbf{High Input Resolution Model} are less robust. 
\textit{Mid}: \textbf{t-SNE of Dataset robustness} Dataset represented via 66 models robustness ($\Gamma^{D}_{16}$), indicates 3 clusters. 
\textit{Right}: \textbf{Layers-wise features L2 similarity}: 
$n\x n$ model layers similarity w/ $224\x224$ ones, for EVA02-B-16. For a given heatmap (\eg $16\x16$), the lower right indicates the similarity of deeper layers (brighter means more similar), while the upper left represents non-similar shallow layers (dull means less similar).  
}
\label{fig:feat_heatmap}
\label{fig:deepanalysis2}
\vspace{-11pt}
\end{figure}

\section{Benchmarking Analysis}

\noindent\textbf{Proposed WAR Metrics}: 
Spearman correlation between the rankings of 66 models, calculated using SAR and WAR averaging of relative robustness $\Gamma^{D}_{16}$ (across all datasets), and the individual dataset rankings is shown in \underline{\cref{fig:robustness_metrics}(left)}.
WAR shows a slight decrease in avg. correlation (SAR-16 \textit{0.89} vs WAR-16 \textit{0.87}), but it also improves the representation of EuroSAT \& ImageNet-A.
The correlation score for EuroSAT increased from a weak/no correlation of 0.26 to a moderate 0.49. \vspace{5pt} \\
\noindent \textbf{Model Architecture / Pretraining:}
\underline{\Cref{fig:model_arch} (right, (i))} shows, on average, 
\textbf{larger model} (x-axis) are \textbf{more robust}. 
Among the models, 
CLIP-ResNets (stars) are the least robust (compared to transformers (dots)) while EVA, MetaCLIP, CLIPA, and OpenCLIP exhibit the highest robustness against the LR.
Higher GFLOP (size of dots) weakly impacts robustness with too many exceptions. \vspace{5pt}\\
\noindent \textbf{Pretraining `Quality over Quantity'}: 
\underline{\Cref{fig:model_arch} (right (ii))} shows pre-training dataset size weakly correlates with robustness, with exceptions like SigLIP (10B), and M2-encoder (6B) performing worse. 
Models pre-trained on DataComp-1B generally outperform those pre-trained on LAION-2B, despite having over 500M fewer image-text pairs (\underline{\cref{fig:DC1B_L2B} (left)}). 
This suggests that the \textbf{model} and \textbf{quality of pre-training} have a greater impact on robustness \textbf{than the quantity of pre-training}. \vspace{5pt}\\
\noindent \textbf{Model Specific}:
We remove architectural size advantages by categorizing top-performing models into parameter buckets as shown in \underline{\cref{fig:model_size_wise_comparison} (mid)}. 
For smaller models (150M and 430M parameters), OpenCLIP matches EVA and outperforms MetaCLIP and CLIPA, despite these two being built on top of OpenCLIP. 
However, for larger models, this trend reverses, with EVA-CLIP remaining superior for comparable sizes.
Two factors contribute to performance discrepancies within models of the same parameter size:
\textbf{(1) Fine-tuning:} 
ALBEF and BLIP fine-tuned variants 
are less robust on EuroSAT and Aircraft, reducing their overall robustness (\underline{\cref{fig:deepanalysis} (right)})
\textbf{(2) Higher input resolution:} 
Models with higher input resolutions (\eg $336\x336$) are generally less robust than their $224\x224$ counterparts, likely due to increased interpolation from $16\x16$ to higher resolutions (\underline{\cref{fig:deepanalysis2} (left)}). \vspace{5pt} \\  
\noindent \textbf{Dataset Specific}:
Relative robustness of 66 models on each dataset forms its robustness vector representations. 
Representing these vectors using t-SNE (\underline{\cref{fig:deepanalysis2} (mid)}), reveal three major dataset clusters:
high-robustness (long bars) (\eg Caltech101), 
weakly robust (medium bars) (\eg ImageNet), and 
least robust (smallest bar) (\eg, ImageNet-A).
This indicates that \textbf{low-resolution performance varies by dataset}, which warrants a deeper dive into dataset-specific robustness, \textit{left as future work}.

\begin{figure}[!t]
  \centering
  \includegraphics[width=0.98\linewidth]{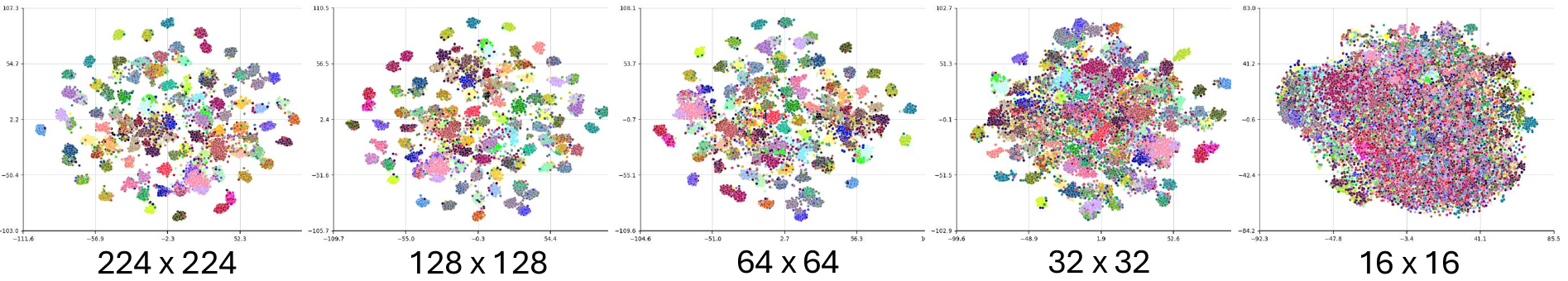}
  \caption{\textbf{Feats t-SNE}: EVA-02-CLIP-B/16 test features for Food-101, colored using class labels. With low resolutions ($16\x16$, and $32\x32$), features become indistinguishable, thereby overlapping.
  }
  \label{fig:tsne_feats}
  \vspace{-3pt}
\end{figure}

\noindent \textbf{Inside Model}:
\underline{\Cref{fig:acc_drop}} shows the accuracy of all models first drops at $64\x64$, with a more significant decline after $32\x32$.
EVA-B/16 features t-SNE (\underline{\cref{fig:tsne_feats}}) shows \textbf{features become indistinguishable as resolution decreases}.
Inside the model, \underline{\Cref{fig:feat_heatmap} (right)} shows the pairwise similarity (L2 distance)~\citep{kornblith2019similarity}
between layers of models trained at different resolutions with the $224\x224$. Diagonal elements ($i^{th}$ layer of $n\x n$ model similarity with $i^{th}$ layer of a model trained at $224\x224$), is \textbf{more similar towards the deeper end} (lower right, the similarity is brighter), \textbf{than the initial layers} (upper left, the similarity is dull).  
Additionally, 
model similarity increases with resolution, while layers remain differentiable at all resolutions (dull non-diagonal values).
\vspace{-5pt}

\begin{figure}[!t]
\centering
\subfloat{\includegraphics[height=3.5cm]{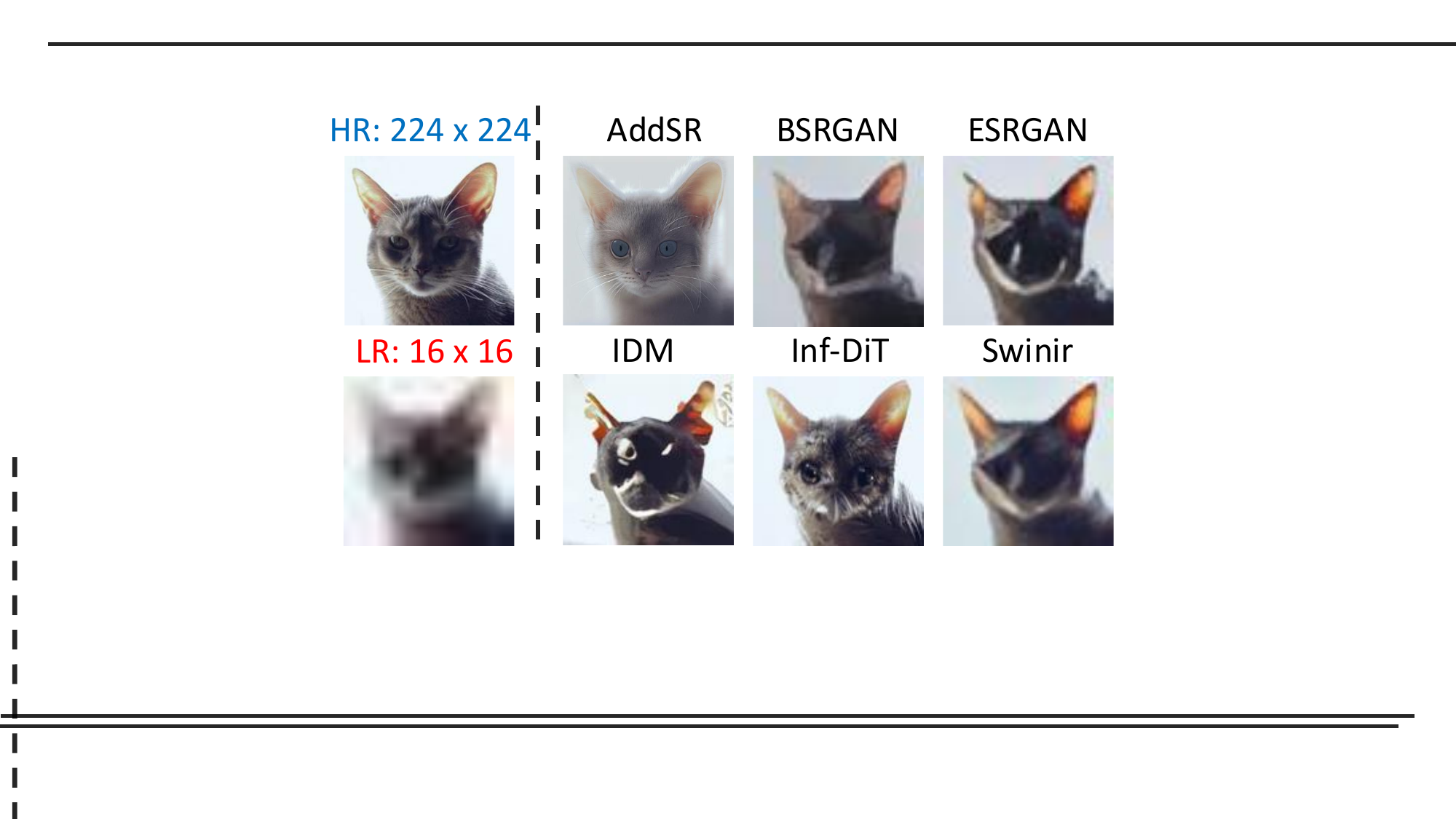}}
\hfill
\subfloat{\includegraphics[height=3.5cm]{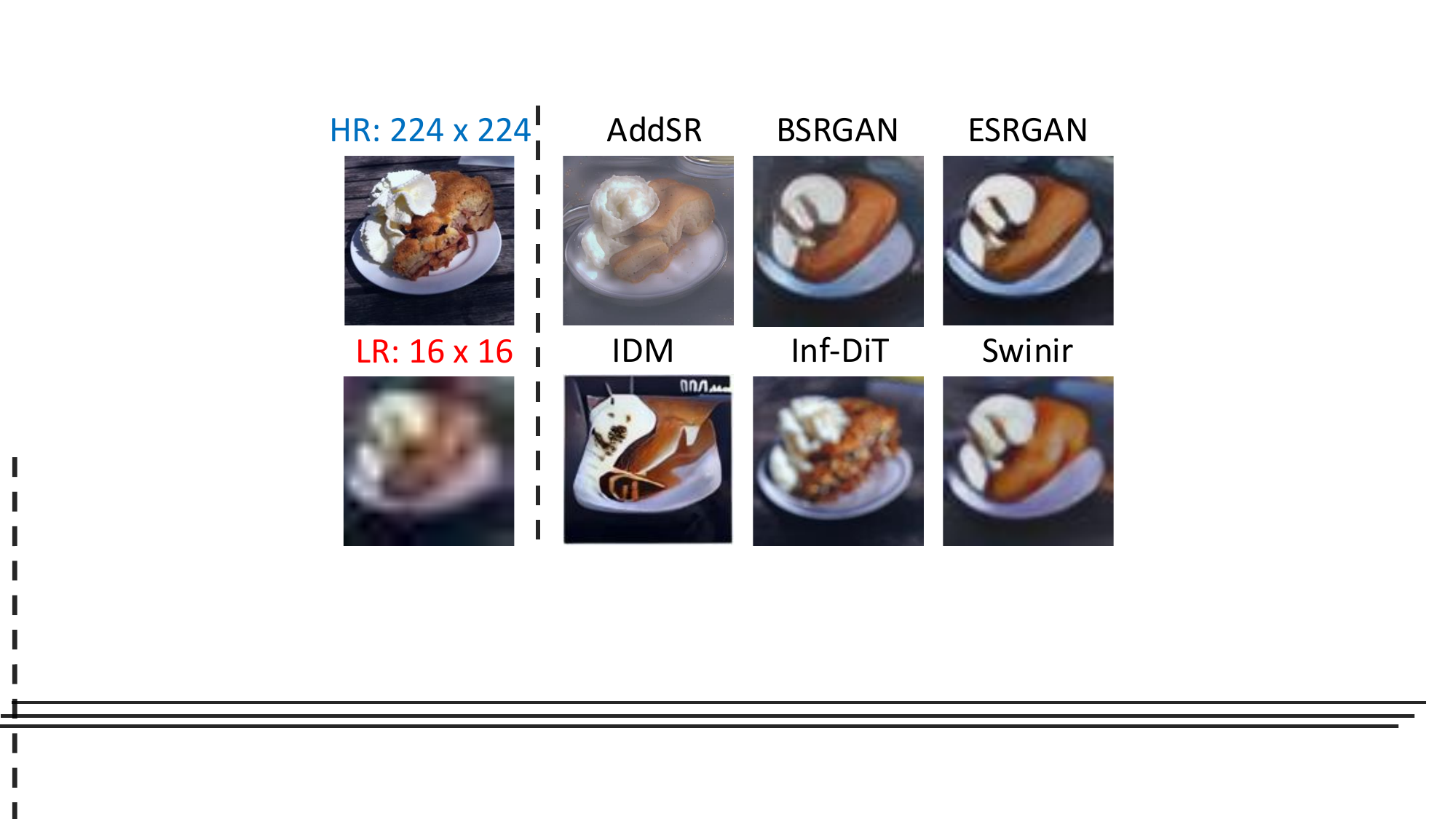}}
\caption{
\textbf{Super resolution at {\boldsymbol{$16\x16$}}}: Image from Pets (\textit{left}) and Food102 (\textit{right}). Models include AddSR~\SPCITE{xie2024addsr}, BSRGAN~\SPCITE{zhang2021designing}, ESRGAN~\SPCITE{wang2018esrgan}, IDM~\SPCITE{gao2023implicit}, Inf-DiT~\SPCITE{yang2024inf}, and Swinir~\SPCITE{liang2021swinir}.
}
\label{fig:sr_images}
\vspace{-6pt}
\end{figure}

\section{Proposed Method: LR-TK0}
\vspace{-4pt}
\Cref{fig:img_ex} reveals two key insights: i) LR lacks fine-grained details ii) FM(s) make semantically reasonable predictions even at $16\x16$, 
highlighting the importance of preserving semantic capabilities (pre-training).
While super-resolution (SR) methods could restore lost details without affecting models, zero-shot SR for very low resolutions $(\le64\x64)$, doesn't work well in practice, as shown in \cref{fig:sr_images}, where SR models fail to reconstruct out-of-domain images at $16\x16$.
To enhance model robustness against low resolution, our solution \textbf{LR-TK0} adds 
trainable LR tokens on top of frozen transformers (preserving the pre-trained weights).
These LR tokens learn to bridge the gap between the high-resolution (HR) and low-resolution (LR) domains, via self-supervised distillation (\cref{sec:model_training}).
We train these tokens on synthetically generated diffusion-based images (\Cref{sec:training_data}) in a task-agnostic setting, ensuring the model is not exposed to any of the 15 target datasets.
\vspace{-5pt}

\subsection{LR tokens}
\begin{figure}[!t]
\centering
\subfloat
{\includegraphics[height=3.8cm]{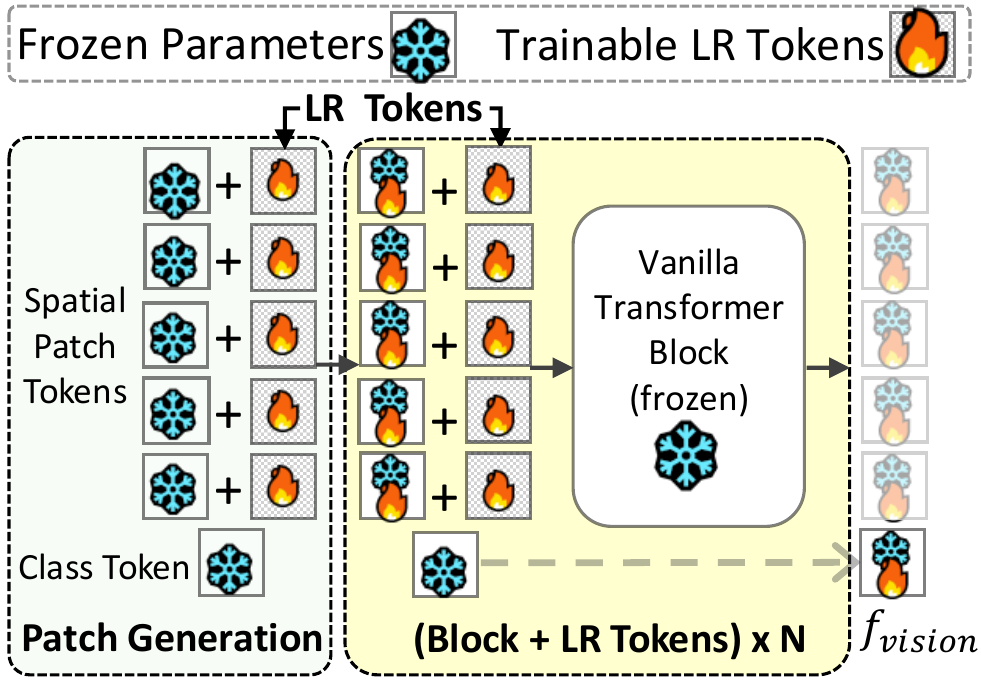}}
\hfill
\subfloat
{
\includegraphics[height=3.8cm]{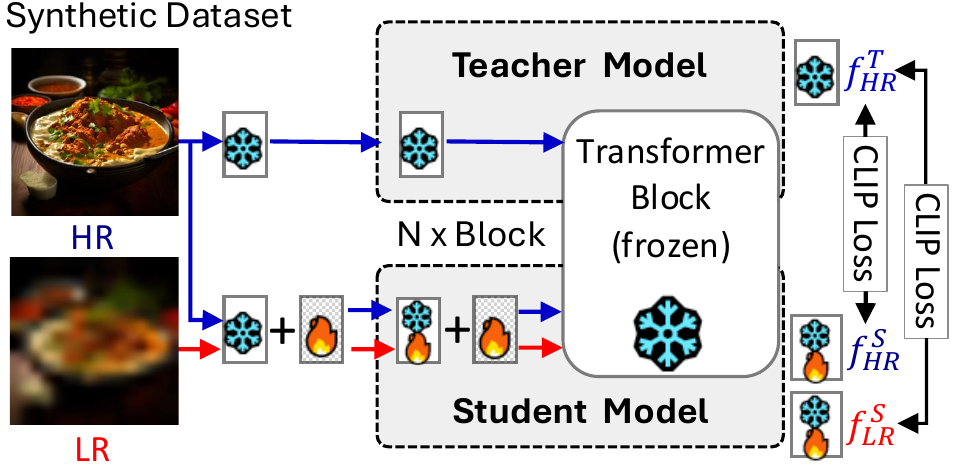}
}
\caption{
Fire (\& ice) icons represent trainable  (\& frozen) parameters.
\textit{Left:} \textbf{LR tokens} are \underline{added} to the frozen spatial patches (white) after patch generation, before each frozen transformer block, and class token as a final feature. 
\textit{Right:} \textbf{LR-TK0:}
Multi-scale training (only 1 shown for simplicity). 
Teacher (w/o LR tokens) generates $f^T_{HR}$  (HR), Student (w/ LR tokens) generates both $f^S_{HR},f^S_{LR}$.
}
\label{fig:pseudo_distillation}
\label{fig:model}
\end{figure}

\vspace{-6pt}
\label{sec:model_training}
To preserve the zero-shot capabilities of the model; \textit{pre-trained weights of the model are frozen}.
Instead, additional trainable tokens, referred to as ``LR Tokens", are added on top of the spatial tokens after RGB to patch tokens conversion (patchification) and before each transformer block. As shown in \cref{fig:model} (left) \# LR tokens = \# Spatial tokens $\x$ (N+1) blocks. 
These tokens aim to compensate for the loss of details in low resolution, thereby enhancing the model’s interpretability of LR images.
Contrary to prompt learning~\citep{jia2022visual}, where task-specific tokens are concatenated to the spatial tokens, ours are added/merged.  
\Cref{fig:feat_heatmap} (right)  indicates LR feature at the initial layer deviates more than the later ones, thus LR tokens are added at every block.  

\noindent \textbf{LR-TK0 Technique}:
We adopt the multi-scale paradigm~\citep{chen2019learning} \ie training multiple low resolutions per HR image, given its success in the LR domain.
Model without LR tokens (frozen pre-trained weights) acts as a teacher generating feature representations for HR images, as true embedding $f^\text{\textbf{T}}_{HR}$. 
In contrast, LR tokens (\& pre-trained model) act as student, generating embeddings for both HR ($f^\text{\textbf{S}}_{HR}$) and LR image(s) ($f^\text{\textbf{S}}_{LR}$) as shown in  \cref{fig:pseudo_distillation} (right).
$f^\text{\textbf{S}}_{HR},f^\text{\textbf{S}}_{LR}(s)$ are matched with $f^\text{\textbf{T}}_{HR}$ using a contrastive loss~\citep{radford2021learning}, similar to text and image alignment. 
Anchoring HR-LR features around frozen teacher avoids direct matching of HR-LR embeddings, preventing pulling the HR features towards LR ones (converging into one)~\citep{9137263}. 
This also ensures features w/ and w/o spatial tokens remain similar (regularization). 
Feature matching doesn't require any labels for these synthetic images, aka \textbf{unsupervised}.
It also \textbf{task agnostic}, \ie doesn't involve any model task-related characteristics (classification in this case). 
\vspace{-15pt}

\begin{figure}[!t]
\centering
\subfloat{
\includegraphics[height=2cm]{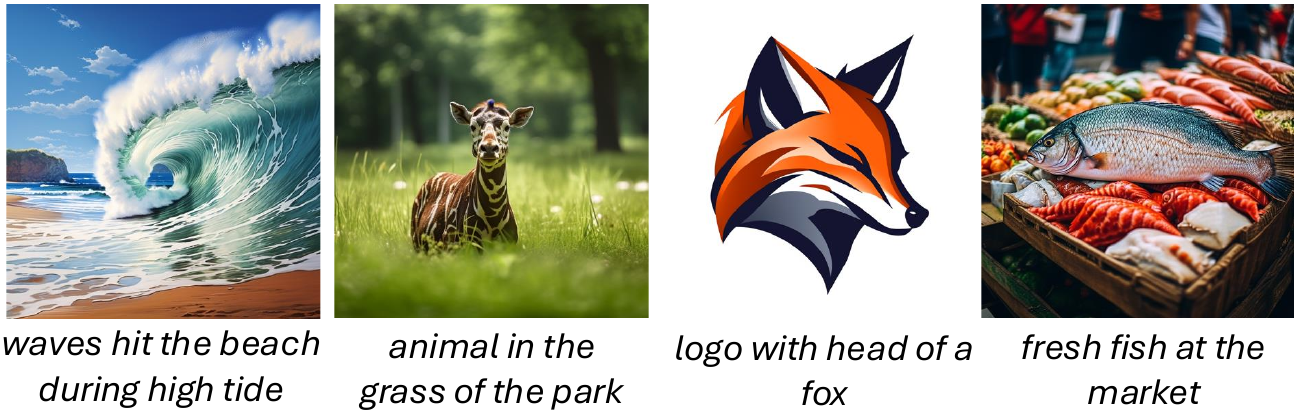}
\vspace{-3pt}
}
\hfill
\subfloat
{
\includegraphics[height=2cm]{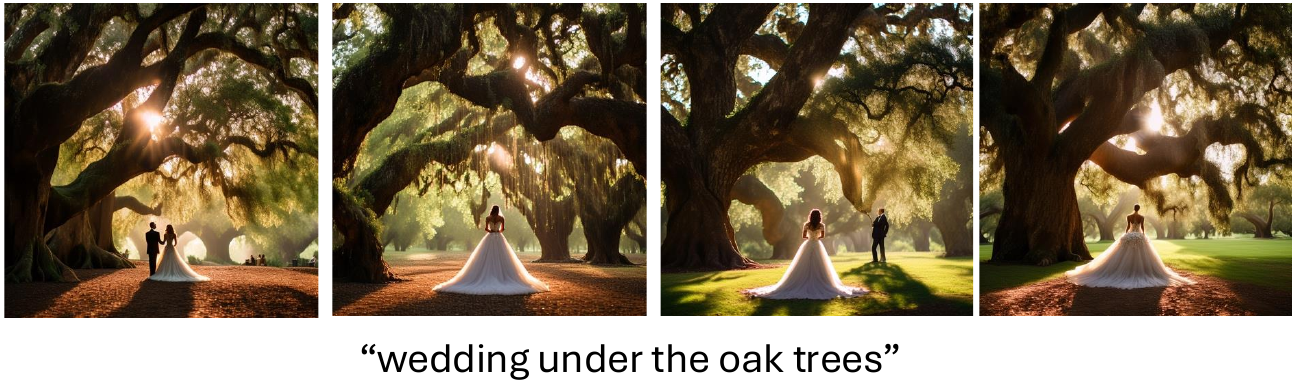}
\vspace{-3pt}
}
\caption{ 
\textbf{Synthetic Images}: (\textit{Left}) Images generated using PIXART-$\alpha$~\SPCITE{chen2023pixartalpha} using randomly sampled captions from Conceptual Captions ~\SPCITE{sharma2018conceptual}.
(\textit{Right}) Multiple images per caption.
}
\label{fig:Sr_images}
\vspace{-5pt}
\end{figure}

\subsection{Synthetic HR Dataset}
\label{sec:training_data}
We use the diffusion model PIXART-$\alpha$~\citep{chen2023pixartalpha} to generate synthetic HR images, via 7,000 randomly sampled captions from Conceptual Captions~\citep{sharma2018conceptual}. 
We expand our training set by creating multiple images (variations, human observation) per caption as shown in \cref{fig:Sr_images}. 
Conceptual Captions are commonly used in pretraining many zero-shot models (\cref{tab:model_desc}), and using synthetic diffusion-based images helps LR tokens capture a wide range of domains, ensuring generalized training.
\textbf{Random captions avoid targeting any specific dataset}. 
To our knowledge, our work is the first to train a model on synthetic diffusion images for zero-shot evaluation, contrary to training on a subset of target datasets~\citep{chen2024robustsam}.
Following the multi-scale paradigm, we downsample HR images to a randomly sampled spatial resolution (height = width) from three LR resolution buckets {[$16$,$32$]}, {[$32$, $64$]}, {[$64$, $128$]}, forming  HR-LR image pairs. \vspace{4pt} \\ 
\noindent \textbf{Zero-Shot:}
If \underline{7,000 (or fewer)} concepts/captions can consistently enhance model performance across \underline{15 datasets}, it suggests that the model is likely learning the relationship between HR and LR features rather than exploiting shortcuts. 
This is supported by greater improvements at LR ($16\x16$) compared to HR ($128\x128$). 
If the model somehow cheats the zero-shot evaluation using diffusion-generated images, we would expect similar or better performance improvements at HRs.
\vspace{-5pt}

\section{Proposed Method: Experimentation \& Ablation}
\vspace{-4pt}
\noindent
\textbf{Implementation Details}
\label{sec:implementation}
Models are trained with 7K captions (\& 30 images/captions) in a multi-scale paradigm. 
EVA is trained for 200 epochs, while MetaCLIP and OpenCLIP are for 10 epochs. 
Evaluation metrics (\cref{sec:eval_metrics}): \underline{SAR} (simple averaging of $\gamma^{D}_{n}$), \underline{WAR} (weighted averaging of $\Gamma^{D}_{n}$), and \underline{Acc} (average top-1). Higher number means better performance. 
Vanilla model's HR accuracy computes the accuracy gap $\mathcal{E}_D$, and dataset weights derived for $16\x16$ used for all resolutions (more in \Supp). 
\textbf{`EVA-02-CLIP-B/16' (EVA-B/16)}, is used for all our model-level analysis. 
\vspace{-8pt}

\subsection{Results}
\label{sec:results}
\begin{figure}[!tb]
  \centering
  \includegraphics[width=1\linewidth]{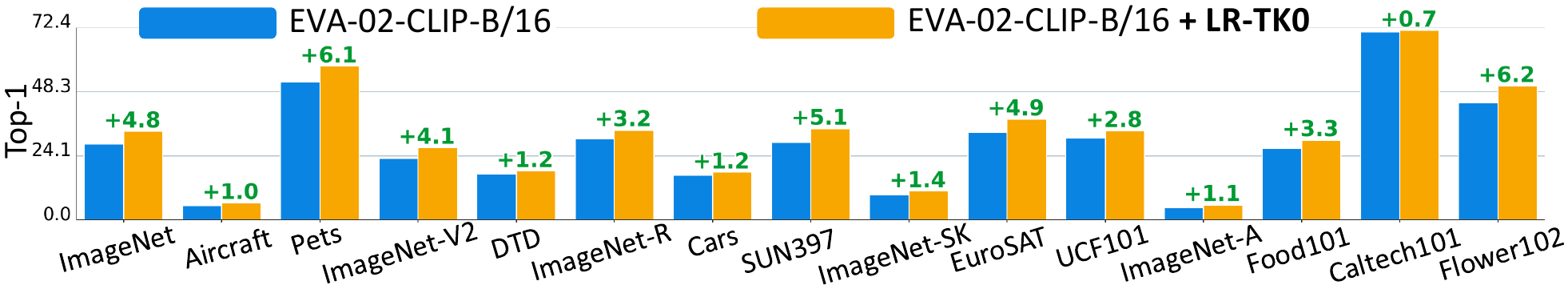} 
  \vspace{-11pt}
  \caption{\textbf{Baseline vs LR-TK0}: 
  Top-1 accuracy for EVA-B/16 on $16\x16$. (more in \Supp)
  }  
\label{fig:performance_dataset}
\vspace{-5pt}
\end{figure}

\textbf{\Cref{tab:proposed_main_results}} shows our LR tokens consistently enhance robustness at low resolutions ($16\x16$ \& $32\x32$), particularly for MetaCLIP. 
While the low resolution is often seen as a domain shift problem~\citep{9098036},
leading to potential declines in HR performance, our multi-scale training and HR teacher distillation minimize accuracy drops at higher resolutions (1-2\% accuracy drop).
Also, LR tokens have a minimal parameter gain ($+3\%$).
\textbf{\Cref{fig:performance_dataset}} shows Top-1 accuracy for EVA-B/16 with and without our LR-TK0, at $16\x16$, with max improvement on Flower-102 (6.2\%).
\textbf{\Cref{tab:comparison_with_SR}} compares EVA-B/16 with super-resolution (SR) methods, with SR methods performing poorly in zero-shot settings for very low resolutions (\cref{fig:sr_images}).
In contrast, our approach is better suited for zero-shot scenarios. 
Diffusion-based SR method IDM is too computationally expensive to evaluate on large datasets like ImageNet (results in \Supp).
\textbf{\Cref{tab:vpt_robust_sam}} applies our LR-TK0 technique to visual prompt tuning which concatenates tokens (instead of adding) only before the first block.
RobustSAM (segmentation models) modified for image classification (\Supp).
\vspace{-2pt}

\begin{table}[!t]
\caption{\textbf{LR-TK0 improvement on Foundation models:} `Meta-B/16': MetaCLIP-ViT-B/16 (2.5B), 
 `OC-B/16': OpenCLIP-ViT-B/16. 
 Higher number $\propto$ better performance. 
}
\vspace{-3pt}
\label{tab:proposed_main_results}
\renewcommand{\arraystretch}{1.2}
\setlength\tabcolsep{2.5pt}
\scalebox{0.83}{
\centering
\begin{tabular}{
c|c|
c|c|c||    c|c|c||    c|c|c||   c|c|c|| c|c|c
}
\specialrule{1pt}{0pt}{0pt}
\rowcolor{mygray} 
 &  \# 
& \multicolumn{3}{c||}{$16\x16$}
& \multicolumn{3}{c||}{$32\x32$}
& \multicolumn{3}{c||}{$64\x64$}
& \multicolumn{3}{c||}{$128\x128$}
& \multicolumn{3}{c}{$224\x224$}
\\
[-.4pt]\hhline{~~ --- --- --- --- ---}
\rowcolor{mygray} 
\multirow{-2}{*}{Model} & Param 
& SAR & WAR & Acc
& SAR & WAR & Acc
& SAR & WAR & Acc
& SAR & WAR & Acc
& SAR & WAR & Acc
\\
\hline\hline
EVA-B/16 & 149.7M & 
38.0 & 30.7 & 28.1 
& 74.4 & 64.8 & 53.5 
& 92.4 & 85.8 & 65.2 
& 98.4 & 96.1 & 68.8
& 100 & 100 & 69.6 \\  
\rowcolor{mygray} 
\textbf{+LR-TK0} & 155.2M  & 
42.4 & 35.4 & 31.3 
& 75.3 & 66.4 & 54.1 
& 91.8 & 85.9 & 64.8 
& 97.8& 95.5 & 68.3 
& 99.1 & 98.7 & 69.0 
\\ 
\hline \hline
Meta-B/16  & 149.6M  & 
32.1 & 27.2 & 23.4 & 
65.3 & 54.4 & 47.0  
& 89.5 & 83.6 & 62.9 
& 98.5 & 96.7 & 68.5 
& 100 & 100.0 & 69.4 
\\
\rowcolor{mygray} 
\textbf{+LR-TK0} & 151.6M
& 41.9 & 38.9 &  30.2 
& 71.7 & 66.0 & 51.0
& 89.3 & 85.4 & 62.6 
& 96.7 & 95.4 & 67.3
& 97.6 & 97.4 & 67.9
\\
\hline \hline 
OC-B/16 & 149.6M
& 33.4 & 26.5 & 24.8 
& 68.6 & 59.5 & 49.8  
& 89.2 & 84.1 & 63.6 
& 96.8 & 94.8 & 68.3 
& 100 & 100 & 70.4
\\
\rowcolor{mygray} 
\textbf{+LR-TK0} & 151.6M &   
37.4 & 34.4 & 27.4 
& 69.0 & 63.0 & 49.9
& 88.8  & 84.2 & 63.4
& 96.8  &  95.1 & 68.4      
& 99.0 &  99.0 & 69.8 
\\ 
\specialrule{1pt}{0pt}{0pt}
\end{tabular}
}
\vspace{-3pt}
\end{table}

\begin{table}[!t]
\begin{minipage}{.47\linewidth}
\caption{\textbf{Comparison with SR methods}: 
EVA-B/16 results, SR-specific pre-processing.
}
\label{tab:comparison_with_other_zeroshot}
\label{tab:comparison_with_SR}
\setlength\tabcolsep{2.2pt}
\renewcommand{\arraystretch}{1.2}
\scalebox{0.98}{
\centering
\begin{tabular}{
l|
c|c|c|    c|c|c 
}
\specialrule{1pt}{0pt}{0pt}
\rowcolor{mygray} 
& \multicolumn{3}{c|}{$16\x16$}
& \multicolumn{3}{c}{$32\x32$}
\\[-.4pt] \hhline{~--- ---}
\rowcolor{mygray} 
\multirow{-2}{*}{Method}
& SAR & WAR & Acc
& SAR & WAR & Acc
\\
\hline\hline
Baseline 
& 34.1 & 26.8 & 25.0 
& 71.8 & 59.0 & 51.2 
\\ 
\hline
BSRGAN
& 12.4 & 12.2 & 8.8    
& 37.3 & 28.7 & 26.9 
\\
ESRGAN
& 14.2 & 15.1 & 10.0 
& 40.3 & 32.6 & 28.9 
\\ 
Swinir
& 17.9 & 17.6 & 12.7 
& 47.7 & 38.3 & 34.3
\\ 
AddSR
& 20.5 & 16.8 & 15.0 
& 48.3  & 36.0 &  35.2 \\
Inf-DiT
& 29.0 & 25.3 & 20.9
& 67.7  & 58.6 &  48.0
\\ 
\hline
\rowcolor{mygray} 
Our 
& \textbf{38.9} & \textbf{29.5} & \textbf{28.4} 
& \textbf{73.1} & \textbf{62.0} & \textbf{52.0}
\\
\specialrule{1pt}{0pt}{0pt}
\end{tabular}
}
\end{minipage}
\hfill
\begin{minipage}{.49\linewidth}
\setlength\tabcolsep{1.4pt}
\caption{\textbf{Generalization of LR-TK0 with other Zero-Shot Techniques}: 
Visual prompt Tuning (VPT)~\SPCITE{jia2022visual} concatenates 50 learnable tokens to spatial tokens.
RobustSAM~\SPCITE{chen2024robustsam} is an image segmentation model modified for classification.
}
\label{tab:vpt_robust_sam}
\setlength\tabcolsep{2.5pt}
\scalebox{0.95}{
\begin{tabular}{
l| c|c| c|c | c || c 
}
\specialrule{1pt}{0pt}{0pt}
\rowcolor{mygray} 
& \multicolumn{5}{c||}{WAR}
& SAR
\\[-.4pt] \hhline{~--- ---}
\rowcolor{mygray} 
\multirow{-2}{*}{LR-TK0}
& 16 &  32 & 64 & 128 & 224 
& 16 
\\
\hline\hline
Baseline 
& 30.7 & 64.8 & 85.8 & 96.1 & 100 & 38.0 \\
+VPT
& 35.5 & 64.1 & 84.6 & 94.5 & 97.8 & 42.6 \\
+RobustSAM
& 32.2 & 61.5 & 82.7 & 92.4 & 93.0 & 37.8 \\
+LR Tokens 
& 35.4 & 66.4 & 85.9 & 95.5 & 98.7 & 42.4 \\
\specialrule{1pt}{0pt}{0pt}
\end{tabular}
}
\end{minipage}
\vspace{-10pt}
\end{table}

\subsection{Ablation Study}
\noindent \textbf{Design Choices:} \textbf{\Cref{tab:design_choices}} shows 
not freezing the pre-trained weights (\ie fine-tuning the last 4 blocks at 1/100 of the default learning rate) with and without LR tokens (first two rows) degrades the performance, indicating the necessity of preserving pre-trained weights. 
Our design choice is task agnostic \ie model's classification plays no role in learning the HR-LR relationship but classifying LR images into captions (as class labels, \textit{task-oriented}) has more or less the same performance.
\textbf{\Cref{tab:design_choices2}} shows benefit of multi-scale training
(3 buckets, faster to train). 

\begin{table}[!tb]
\renewcommand{\arraystretch}{1.2}
\setlength\tabcolsep{1.5pt}
\begin{minipage}{.55\linewidth}
\caption{\textbf{Ablation:} EVA-B/16 trained with 7K captions \\and 50 images/caption. `CL': use of classifier. \textit{Not} frozen means fine-tuning end-to-end. 
}
\vspace{-2pt}
\label{tab:design_choices}  
\centering
\scalebox{0.95}{
\begin{tabular}{
c|c|c|c|
c|c| c
}  
\specialrule{1pt}{0pt}{0pt}
\rowcolor{mygray} 
Frozen & LR Tk. & CL 
& SAR-16 & WAR-16 & SAR-32 & WAR-32 \\
\hline\hline 
\multicolumn{3}{c|}{Baseline (frozen)} 
& 38.0 & 30.7  &  74.4  &  64.8 \\
\hline 
  &  &    
& 31.1 & 24.5 &  67.2  &  56.6 \\
  & \checkmark &   
& 32.8 & 27.8 &  68.1  & 58.3 \\
\rowcolor{mygray} 
\checkmark & \checkmark &    
& 42.3 & 35.2 &  75.3& 66.4 \\
\checkmark & \checkmark &  \checkmark   
&  42.0 & 34.7 &  75.2 & 65.9 \\ 
\specialrule{1pt}{0pt}{0pt}
\end{tabular}
} 
\end{minipage}%
\quad
\begin{minipage}{.41\linewidth}
\setlength\tabcolsep{5pt}
\centering
\caption{\textbf{Multi-Scale (MS) Buckets}: `+' indicates Cumulative addition. E.g. [64,128] has [16,32] and [32,64] buckets.
}
\vspace{-2pt}
\label{tab:design_choices2}  
\scalebox{0.9}{
\begin{tabular}{@{}l@{ }|
c|c
}
\rowcolor{mygray} 
\specialrule{1pt}{0pt}{0pt}
MS Buckets & WAR-16 & WAR-32 \\
\hline\hline
Baseline & 30.74 & 64.81 \\ 
\hline
{[$16,32$]} 
& 34.01 & 64.77 \\
+ {[$32,64$]} 
& 35.28 &  66.10 \\
\rowcolor{mygray} 
+ {[$64,128$]} 
&  35.45 &  \textbf{66.40} \\ 
+ {[$128,224$]} 
& \textbf{35.73} &  65.91 \\
 \specialrule{1pt}{0pt}{0pt}
\end{tabular}
}
\end{minipage}
\vspace{-6pt}
\end{table}

\begin{figure*}[!tb]
\centering
\begin{minipage}{0.3\textwidth}
\centering
\includegraphics[height=3.25cm]{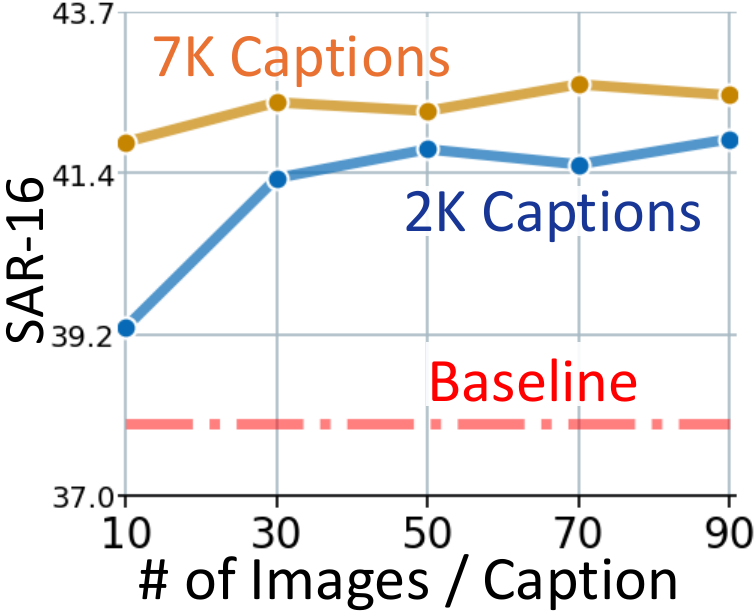}
\vspace{-2pt}
\caption{\textbf{\#Images/Caption:} Robustness vs. Size of diffusion generated dataset.
}
\label{fig:img_per_caption}
\end{minipage}
\hfill
\begin{minipage}{0.31\textwidth}
\centering
\includegraphics[height=3cm]{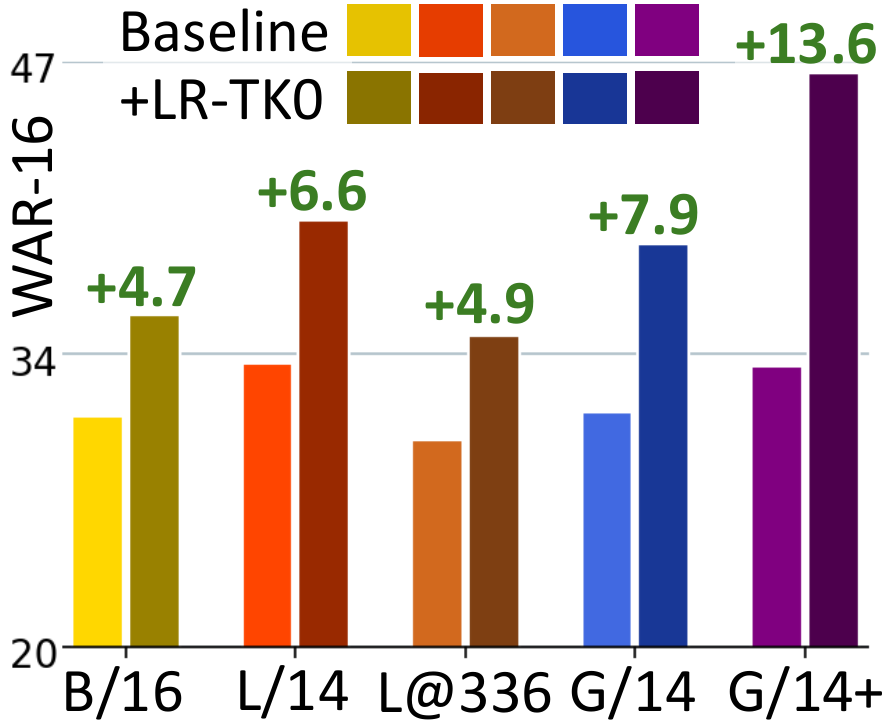}
\vspace{-2pt}
\caption{\textbf{LR-TK0 improves all EVA backbones}: L@336 is L/14 with 336 input}
\label{fig:eva_backbones}
\end{minipage}
\hfill
\begin{minipage}{0.31\textwidth}
\centering
\includegraphics[height=3.1cm]{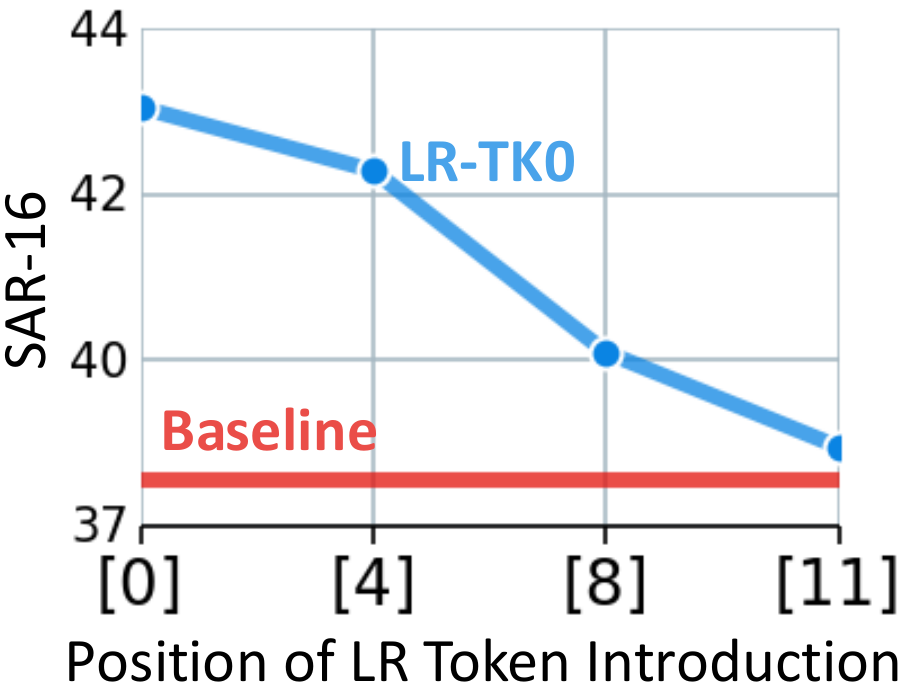}
\vspace{-2pt}
\caption{ 
\boldsymbol{$[i]$} LR tokens introduced starting from $i^{th}$ block (\& none after patchification). 
}
\label{fig:pos_lr_tokens}  
\end{minipage}
 \vspace{-6pt}
\end{figure*}

\begin{figure}[!t]
\centering
\subfloat{
\includegraphics[width=0.48\linewidth]{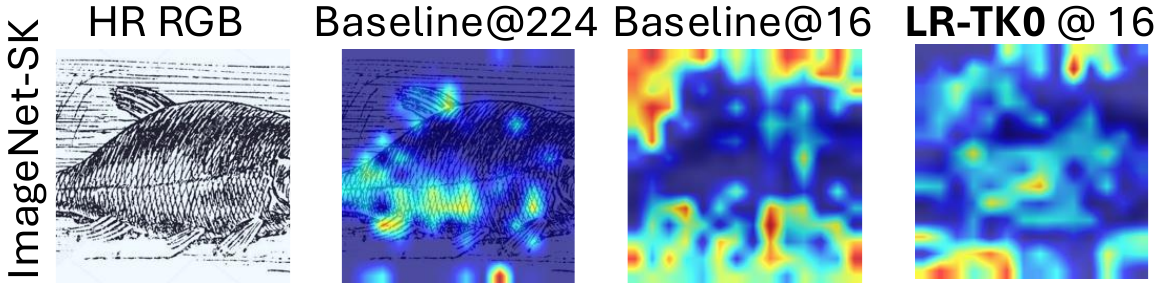}
}
\hfill
\subfloat
{
\includegraphics[width=0.48\linewidth]{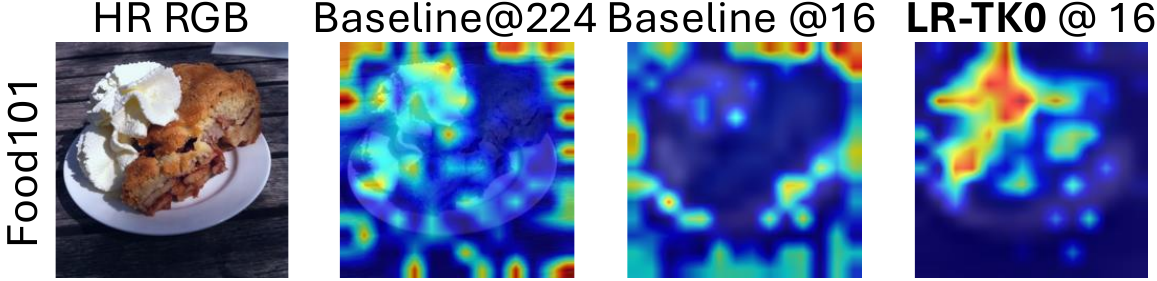}
}
\caption{ 
\textbf{LR token Grad-CAM}: Baseline (EVA-B/16) attention is scattered at $16\x16$ (compared to $224\x224$).  LR-TK0 focuses on the object, likely capturing fine-grained details. @: input resolution. 
}
\label{fig:gram_cam_results}
 \vspace{-8pt}
\end{figure}

\noindent \textbf{\# Images/Caption:}
\textbf{\Cref{fig:img_per_caption}} shows 
multiple images per caption \& even 2000 captions consistently improve performance across 15 datasets, hinting at bridging the gap between HR-LR  domains. \vspace{4pt} \\
\noindent \textbf{EVA backbones:} 
\textbf{\Cref{fig:eva_backbones}}  shows 
LR tokens enhance various EVA backbones, namely, Base (B/16), Large (L/14 \& L@336), and G (G/14 \& G/14+).
Larger backbones, B$<$L$<$G, benefit from 
more tokens (via more layers).
Model with 
$336\x336$ input underperforms (validation, \cref{fig:deepanalysis2} (left)). \vspace{4pt} \\
\noindent \textbf{Position of LR Tokens:}
 \textbf{\Cref{fig:pos_lr_tokens}} shows introducing tokens in the earlier layer (starting from $[i]$-th block, and subsequent layers) is more helpful than later. This helps validate the observation in~\cref{fig:feat_heatmap} (right), \ie 
 initial layers suffer more at low resolution than deeper ones, validating the choice of fixing (introducing tokens) at initial layers than just at final features. \vspace{4pt} \\
\noindent \textbf{Grad-CAM results:} 
On low resolutions of $16\x16$, vanilla model attention is dispersed and not as concentrated as $224\x224$ \textbf{(\cref{fig:gram_cam_results})}. However, our method (w/ LR tokens) shows focus on the object which helps to learn better representations at low resolution.
\vspace{-4pt}

\section{Conclusion}

Our extensive evaluation of Visual-Language Foundation Models through the LR0.FM benchmark has highlighted critical limitations in their ability to generalize under low-resolution conditions, a prevalent issue in real-world scenarios. While larger models and higher-quality pre-training datasets offer increased robustness, our findings underscore the significant impact of fine-tuning and input resolution on performance. Importantly, we observed that low-resolution inputs primarily disrupt the early layers of these models, leading to degraded performance. To address these challenges, we introduced the LR-TK0 strategy, which improves model robustness to low-resolution inputs without altering pre-trained weights, offering a practical solution for real-world applications. Additionally, our proposed Weighted Aggregated Robustness metric provides a more comprehensive evaluation of model resilience, addressing the limitations of existing metrics.

\section{Acknowledgment}
The authors thank Steven Dick (UCF High-Performance Computing) and Rohit Gupta (UCF CRCV) for their help in generating the synthetic data set.

\bibliography{iclr2025_conference}
\bibliographystyle{iclr2025_conference}

\newpage
\appendix
\begin{center}
    \Large{\textbf{LR0.FM: Low-Resolution Zero-Shot Classification\\ benchmark for Foundation Models\\
(Appendix)}}
\end{center}

\section{Dataset Description}
This paper presents a comprehensive benchmarking of zero-shot image classification on low-resolution images utilizing 15 diverse datasets, each representing prominent computer vision challenges as depicted in \Cref{tab:dataset}. Among them, ImageNet~\cite{deng2009imagenet} stands out as a significant repository, containing 50,000 (in test-set) labeled images and serving as a standard for evaluating image classification models. 
Caltech101~\cite{griffin2007caltech}, with its 6,085 test-set images spanning 101 object categories, is widely used for object recognition tasks.
The Describable Textures Dataset (DTD)~\cite{cimpoi2014describing}, comprising over 1,880 texture images in the test-set, facilitates texture analysis.
Food101 provides 25,250 test-set images across 101 food categories, supporting food recognition tasks. SUN397’s~\cite{zhou2014learning} 19,850 annotated test-set images aid scene recognition in understanding diverse environments. Stanford Cars~\cite{kramberger2020lsun} and FGVC Aircraft~\cite{maji2013fine} datasets focus on fine-grained classification tasks for vehicles and aircraft, respectively. Oxford Pets~\cite{parkhi2012cats} offers a dataset for pet breed classification, while Flower102~\cite{liu2016flower} is dedicated to flower species recognition. Eurosat~\cite{helber2019eurosat} specializes in land use and cover classification using satellite imagery. UCF101~\cite{soomro2012ucf101}, containing over 1,794 video clips (in test-set), is pivotal for action recognition research, offering a diverse range of action sequences.  Moreover, we explore four ImageNet variants for natural distribution shifts, previously considered as out-of-distribution (OOD) data for ImageNet~\cite{radford2021learning, shu2022test}. ImageNet-V2~\cite{recht2019imagenet} provides an independent test set with 10,000 natural images collected from different sources across 1,000 ImageNet categories, while ImageNet-A~\cite{hendrycks2021nae} contains 7,500 challenging ``natural adversarial examples'' from 200 ImageNet categories misclassified by a standard ResNet-50~\cite{he2016deep}. Lastly, ImageNet-R~\cite{hendrycks2021many} adds further diversity by offering 30,000 artistic renditions across 200 ImageNet categories, and ImageNet-Sketch~\cite{wang2019learning} includes 50,000 black-and-white sketches covering 1,000 categories, collected independently from the original ImageNet validation set. The test dataset size, the number of classes, and dataset focus are further elaborated in \Cref{tab:dataset}.

\begin{table}[!h]
\begin{center}
  \caption{\textbf{Statistics of benchmark datasets for zero-shot image recognition.}}
  \label{tab:dataset}
  \begin{tabular}{@{}p{4.2cm}@{}|
@{}P{1.4cm}@{}|P{2cm}@{}P{2cm}@{}P{4cm}}
    \toprule
    Dataset & Year & Test Size & \# classes & Focus  \\  
    \midrule
    ImageNet-A (\citeyear{hendrycks2021nae}) & 2021 & 7500 & 200 & Generic\\ 
     ImageNet-V2 (\citeyear{recht2019imagenet}) & 2019 & 10,000 & 1000 & Generic  \\ 
     ImageNet (\citeyear{deng2009imagenet}) & 2009 & 50,000 & 1000  & Generic \\
     Caltech101 (\citeyear{griffin2007caltech}) & 2004 & 6,085 & 101 & Generic \\
     \hline 
     ImageNet-Sketch (\citeyear{wang2019learning}) & 2019 & 50,000 & 1000 & Edges \\ 
     ImageNet-R (\citeyear{hendrycks2021many}) & 2021 & 30,000 & 200 & Texture  \\ 
    EuroSAT (\citeyear{helber2019eurosat}) & 2019 & 5,000 & 10 & Texture \\ 
     DTD (\citeyear{cimpoi2014describing}) & 2014 & 1,880 & 47 & Edges, Texture \\
     \hline 
     Food101 (\citeyear{bossard2014food}) & 2014 & 25,250 & 101 & Fine-grained \\  
     Stanford Cars (\citeyear{kramberger2020lsun}) & 2013 & 8,041 & 196 & Fine-grained\\
     FGVC-Aircraft (\citeyear{maji2013fine}) & 2013 & 3,333 & 100 & Fine-grained \\ 
     Oxford Pets (\citeyear{parkhi2012cats}) & 2012 & 3,669 & 37 & Fine-grained \\ 
     Oxford Flowers102 (\citeyear{liu2016flower}) & 2008 & 6149 & 102 & Fine-grained  \\ 
    \hline 
     SUN397 (\citeyear{zhou2014learning}) & 2010 & 19,850 & 397 & Scene understanding\\
     UCF101 (\citeyear{soomro2012ucf101}) & 2012 & 1,794 & 101 & Scene understanding\\
  \bottomrule
  \end{tabular}
  \end{center}
\end{table}

\textbf{\Cref{tab:dataset_template}: Dataset templates:} As the main paper outlines, we adopt CLIP~\cite{radford2021learning} evaluation protocol for all models to ensure a fair comparison of low-resolution robustness. To generate the text embedding for a given image, we utilize dataset-specific templates, such as ``a photo of a [label]'', ``a low-resolution photo of a [label]'', \emph{etc} as detailed in \Cref{tab:dataset_template}. For each class label, we generate multiple text embeddings by inserting the label into $n$ prompt templates and then average these $n$ embeddings. For instance, consider an image of a cat from the Imagenet dataset. With 1000 class labels and 80 prompt templates, we insert the label ``cat'' into the templates, generate 80 corresponding text embeddings, and compute their average to represent the cat class in text space. This process yields 1000 text embeddings, one for each class. The dot product between the image embedding and these 1000 text embeddings produces class logits, where the highest logit score determines the predicted class. In \Cref{tab:dataset_template}, we present data-specific prompt template samples along with the total number of such prompts.

\begin{table}[!t]
\begin{center}
  \caption{\textbf{Benchmark Datasets Templates} Zero-shot image classification. Here \textbf{[L]} is the class name (labels). These templates are taken from CLIP~\citep{radford2021learning} and OPENCLIP~\citep{ilharco_gabriel_2021_5143773}
  }
  \label{tab:dataset_template}
  \begin{tabular}{@{}l@{ }|@{ }l@{ }|@{}P{1.2cm}@{}}
    \toprule
    Dataset & Sample prompt template & \# 
 Prompts \\  
    \midrule
     ImageNet & a low resolution photo of a \textbf{[L]}, a photo of a small \textbf{[L]}, art of a \textbf{[L]}, \etc & 80\\
     ImageNet-SK & a sketch of the \textbf{[L]}, a rendering of a \textbf{[L]}, a drawing of a \textbf{[L]}, \etc & 80\\ 
     ImageNet-A & a sculpture of a \textbf{[L]}, a close-up photo of the \textbf{[L]}, the cartoon \textbf{[L]} \etc  & 80\\ 
     ImageNet-V2 & a black and white photo of a \textbf{[L]}, a \textbf{[L]} in a video game, a toy \textbf{[L]}, \etc 
 & 80\\ 
     ImageNet-R  & a cropped photo of the \textbf{[L]}, a blurry photo of the \textbf{[L]}, graffiti of a \textbf{[L]}, \etc  & 80\\ 
     
     Caltech101  & a photo of a \textbf{[L]}, a painting of a \textbf{[L]}, the origami \textbf{[L]}, the toy \textbf{[L]}, \etc & 34\\
     DTD & a photo of a \textbf{[L]} texture, a photo of a \textbf{[L]} pattern, \etc & 8\\
     Food101 & a photo of \textbf{[L]}, a type of food & 1\\  
     SUN397 & a photo of a \textbf{[L]}, a photo of the \textbf{[L]} & 2\\
     Cars & a photo of a \textbf{[L]}, a photo of my new \textbf{[L]}, a photo of my dirty \textbf{[L]}, \etc & 8\\
     Aircraft & a photo of a \textbf{[L]}, a type of aircraft \& a photo of the \textbf{[L]}, a type of aircraft & 2\\ 
     Pets & a photo of a \textbf{[L]}, a type of pet & 1\\ 
     Flowers102 & a photo of a \textbf{[L]}, a type of flower & 1\\ 
     EuroSAT & a centered satellite photo of the \textbf{[L]}, a centered satellite photo of a \textbf{[L]}, \etc & 3\\ 
     UCF101 & a video of a person doing \textbf{[L]}, a example of a person practicing \textbf{[L]}, \etc & 48\\
  \bottomrule
  \end{tabular}
  \end{center}
\end{table}

\section{Performance Drop}
\begin{figure}[!th]
\centering
\label{fig:performance_drop_lr}
\subfloat[\centering Oxford Pets]
{
\includegraphics[height=3.1cm]{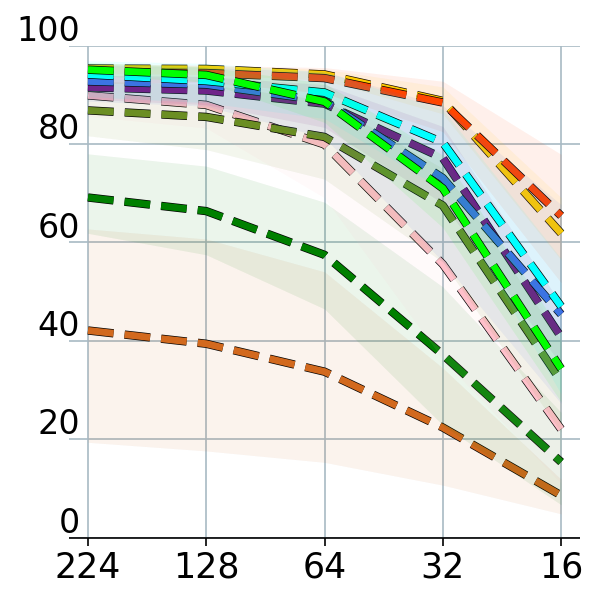}
\label{fig:pets_drop}
}
\subfloat[\centering UCF101]
{
\includegraphics[height=3.1cm]{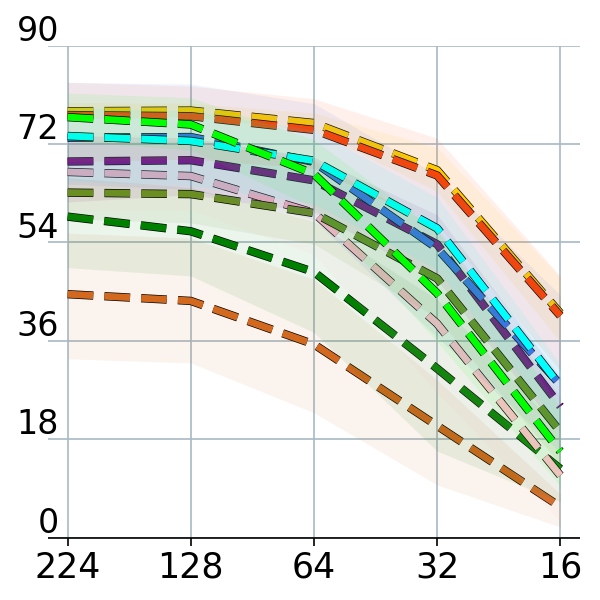}
\label{fig:ucf_drop}
}
\subfloat[\centering Stanford Cars]
{
\includegraphics[height=3.1cm]{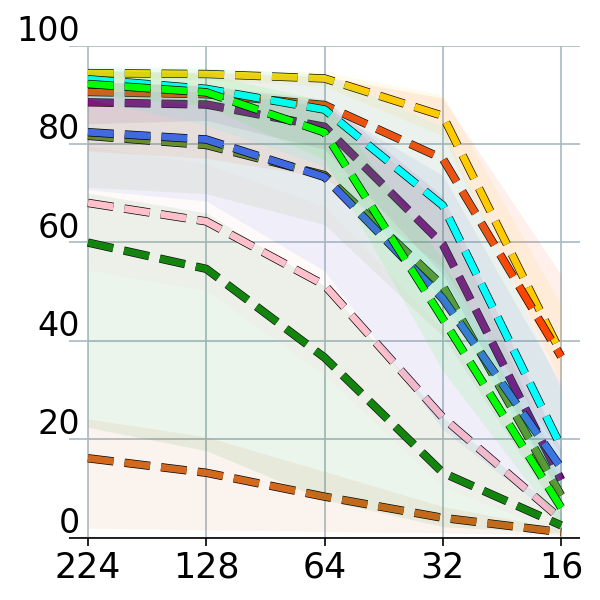}
\label{fig:cars_drop}
}

\subfloat[\centering ImageNet]
{
\includegraphics[height=3.1cm]{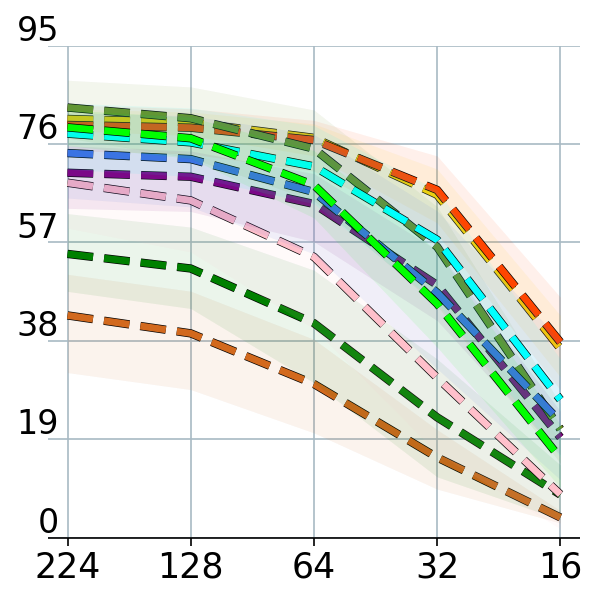}
\label{fig:imagenet_drop}
}
\subfloat[\centering ImageNet-v2]
{
\includegraphics[height=3.1cm]{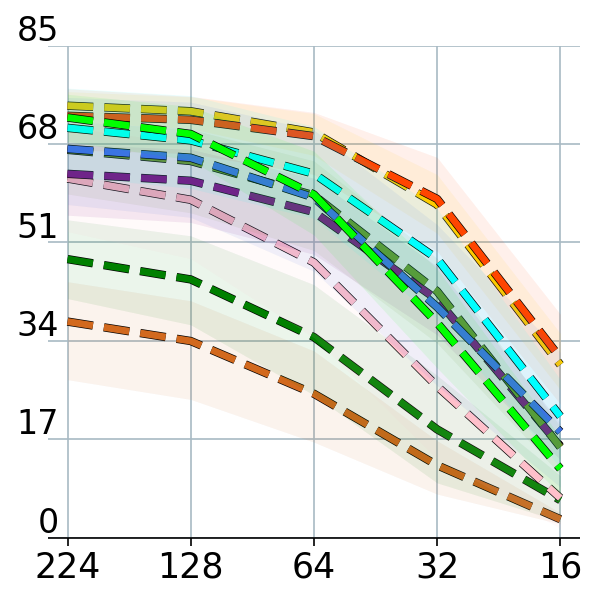}
\label{fig:imagenet_v2_drop}
}
\subfloat[\centering ImageNet-R]
{
\includegraphics[height=3.1cm]{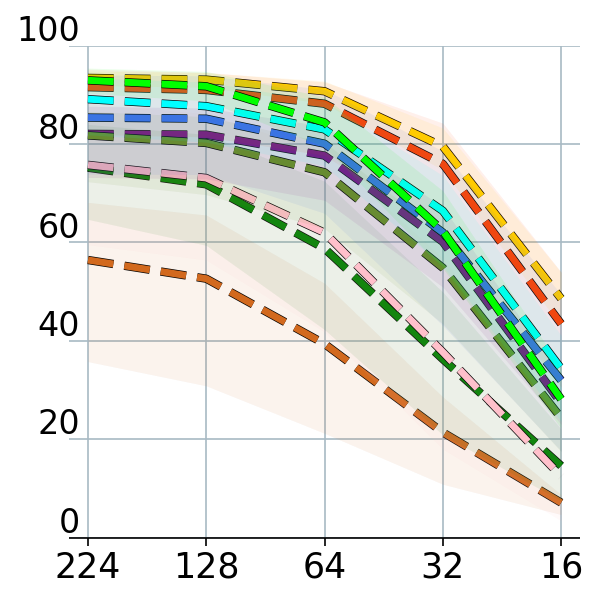}
\label{fig:imagenet_R_drop}
}
\quad
\subfloat[\centering ImageNet-Sketch]
{
\includegraphics[height=3.1cm]{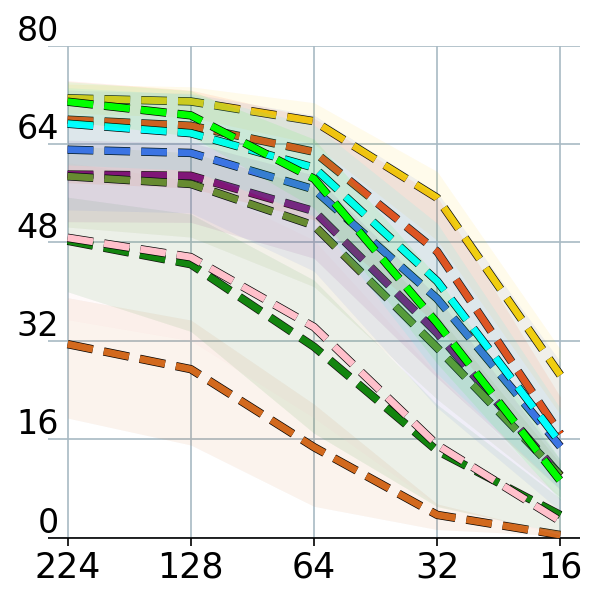}
\label{fig:imagenet_sketch_drop}
}
\subfloat[\centering SUN397]
{
\includegraphics[height=3.1cm]{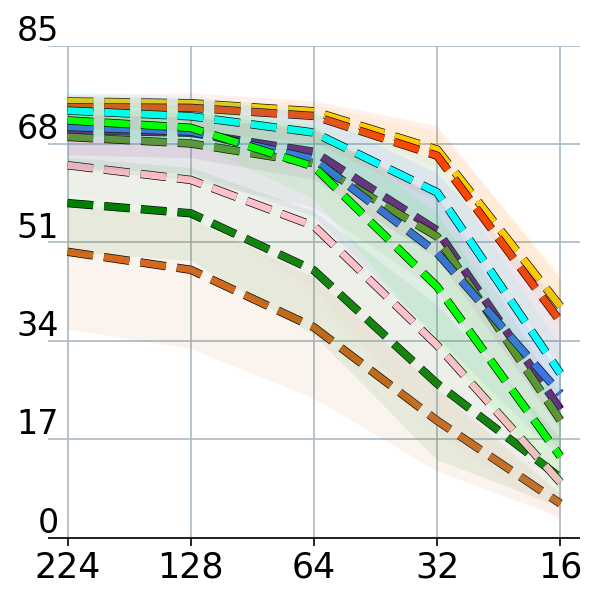}
\label{fig:sun397_drop}
}
\subfloat[\centering Food101]
{
\includegraphics[height=3.1cm]{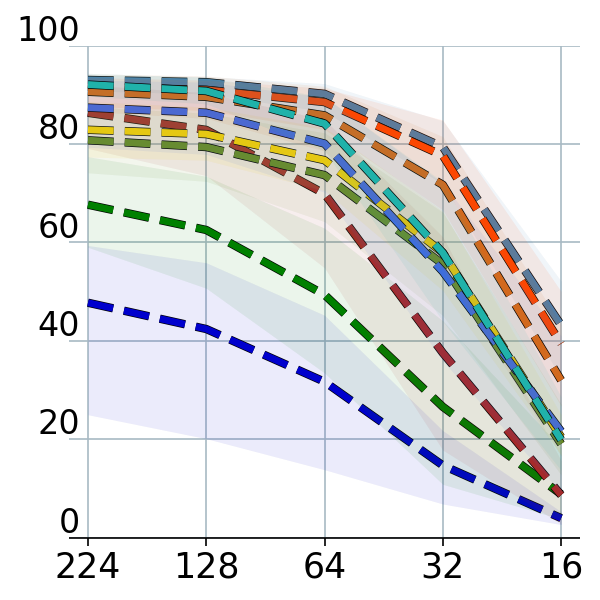}
\label{fig:food_drop}
}
\quad
\subfloat[\centering CalTech101]
{
\includegraphics[height=3.1cm]{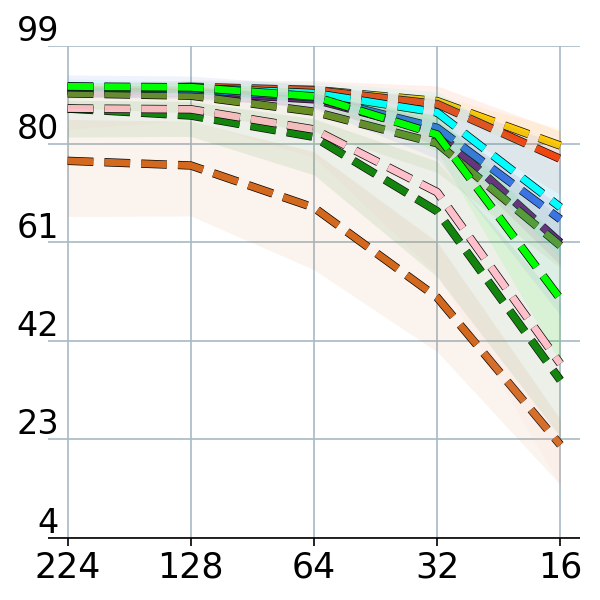}
\label{fig:caltech_drop}
}
\subfloat[\centering DTD Split 1]
{
\includegraphics[height=3.1cm]{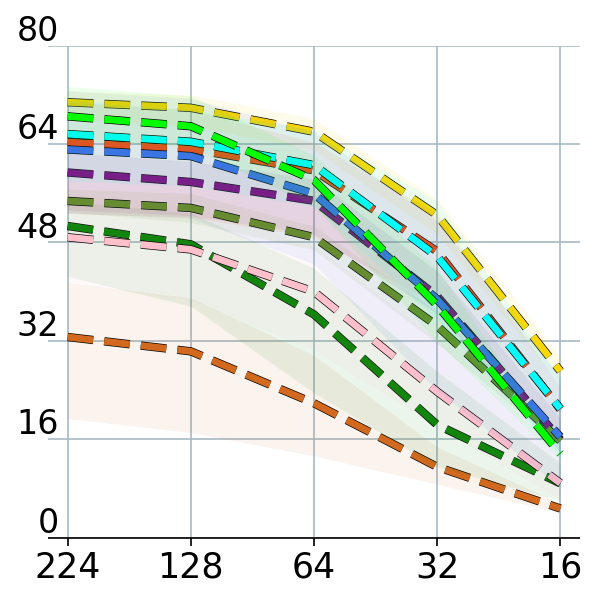}
\label{fig:dtd_drop}
}
\subfloat[\centering EuroSat]
{
\includegraphics[height=3.1cm]{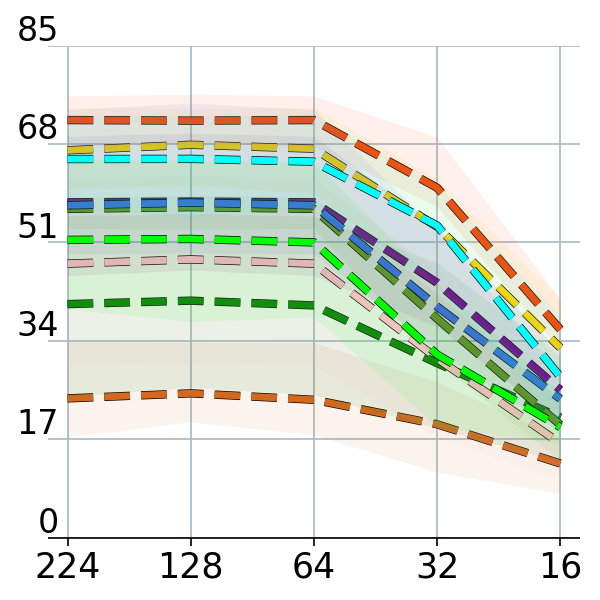}
\label{fig:eurosat_drop}
}
\quad
\subfloat[\centering FGVC Aircraft]
{
\includegraphics[height=3.1cm]{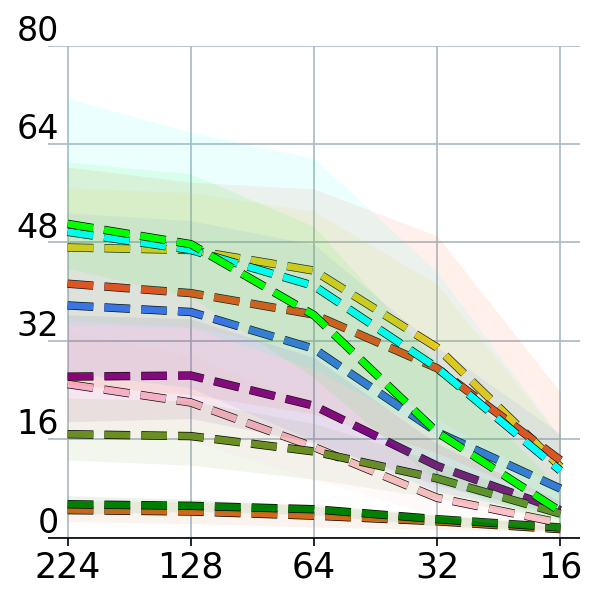}
\label{fig:aircraft_drop}
}
\subfloat[\centering Flowers102]
{
\includegraphics[height=3.1cm]{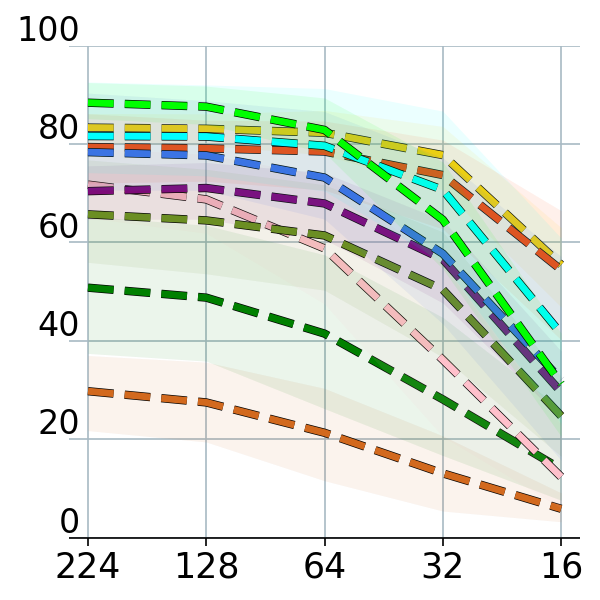}
\label{fig:flower_drop}
}
\subfloat[\centering ImageNet-A]
{
\includegraphics[height=3.1cm]{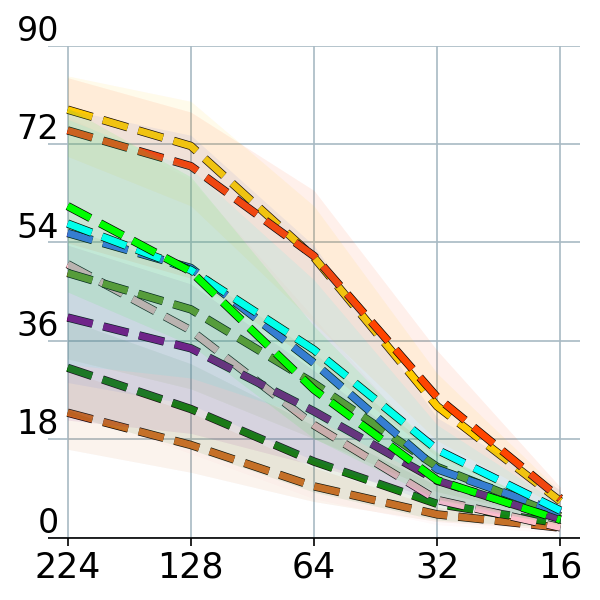}
\label{fig:imagenet_A_drop}
}
\caption{\textbf{Top-1 Accuracy drop}: Drop in accuracy for all models for all the datasets. The color scheme same as Figure 1 from the main submission. 
\\ \\ \\ \\ \\  \\ 
}
\label{fig:supp_per_drop}
\end{figure}

\textbf{\Cref{fig:supp_per_drop}: Zero-shot Classification vs Resolution:} This figure is an extension of \Cref{fig:acc_drop} in the main paper, highlighting a major objective of our study: the relationship between resolution and model performance. As the resolution decreases, we observe a pronounced decline in the performance of all foundational vision-language models when compared to their high-resolution counterparts ($224\times224$), as illustrated in \Cref{fig:supp_per_drop}. Our analysis reveals that this performance drop is consistent across 15 widely used computer vision benchmark datasets, affecting all model backbones. Notably, a performance decline begins at a resolution of $64\times64$, with a more substantial degradation occurring as the resolution falls below $32\times32$.


\section{Weighted Aggregated Robustness (WAR)}

The improved relative robustness is computed as:
\begin{align}
    \Gamma^{D}_{n} &= \gamma^{D}_{n} \times (1 -  e^{- \alpha (\mathcal{E}_D)^2}) \hspace{0.5cm} \mid \hspace{0.5cm} \alpha >> 1 \hspace{0.5cm} \& \hspace{0.5cm} 0 \le \mathcal{E}_D \le 1 \label{eq:improved_robsutness_again}
\end{align}

The additional factor $(1 -  e^{- \alpha \mathcal{E}_D^2})$ is shown in \Cref{fig:new_gamma_old_gamma} in the main paper for $\alpha=200$. It remains close to 1 for the majority values of x or  $\mathcal{E}_D$ but steeps to 0 as $\mathcal{E}_D$ reaches 0. In accuracy terms, as $A_{224}$ comes closer to $A_{rand}$, relative robustness starts dropping to 0. This is shown in \Cref{tab:abnormal_relative} where the normal relative robustness score is high $\!\sim\!20-40$\% for almost random predictions, given the highest resolution accuracy is also close to random prediction. Our weighing term bring these scores to approximately $<12$\%.

\begin{table}[!th]
\begin{center}
  \caption{\textbf{Abnormally high relative robustness for random predictions}. Our All numbers in percentage (100\%). 
  We have shown results only for Easiness $\mathcal{E}_D < 0.15$, \ie highest resolution accuracy ($A_{224}$) is close to random predictions. 
  Our $\hat{\gamma}_{R,D}^{n}$ is plotted for $\alpha=200$, \ie $\hat{\gamma}_{R,D}^n= \gamma_{R,D}^n \x (1 -  e^{- 200 \mathcal{E}_D^2})$.
  $\text{A\textsubscript{rand}} = \frac{1}{\text{\# of classes}}$. High robustness scores within $2\x$ A\textsubscript{rand} are bold. Lines are drawn for easy readability.  }
  \label{tab:abnormal_relative}
  \begin{adjustbox}{angle=90}
  \begin{tabular}{
  |p{3.5cm}@{}|@{}P{1.5cm}@{}| 
@{}P{0.95cm}@{}| @{}P{0.95cm}@{}|| 
@{}P{0.95cm}@{}|@{}P{0.95cm}@{}|@{}P{0.95cm}@{}|| 
@{}P{0.95cm}@{}|@{}P{0.95cm}@{}|@{}P{0.95cm}@{}|| 
@{}P{0.95cm}@{}|@{}P{0.95cm}@{}|@{}P{0.95cm}@{}|| 
@{}P{0.95cm}@{}|@{}P{0.95cm}@{}|@{}P{0.95cm}@{}|| 
}
    \toprule
    Model & Dataset & A\textsubscript{rand} & A\textsubscript{224} 
    & A\textsubscript{16} &$\gamma_{R,D}^{16}$ & $\Gamma^{D}_{16}$ \textit{(\textbf{Our})} &
    A\textsubscript{32} &  $\gamma_{R,D}^{32}$ & $\Gamma^{D}_{32}$ \textit{(\textbf{Our})} & 
    A\textsubscript{64} &  $\gamma_{R,D}^{64}$ & $\Gamma^{D}_{64}$ \textit{(\textbf{Our})} & 
    A\textsubscript{128} &  $\gamma_{R,D}^{128}$ & $\Gamma^{D}_{128}$ \textit{(\textbf{Our})}
    \\  
    \midrule

ALBEF (4M) & Cars & 0.5 & 2.0 &  
0.6 & \textbf{28.8} & 1.2 & 
1.0 & \textbf{51.5} & 2.2 &
1.3 & 66.2 & 2.8 &
1.6 & 83.3 & 3.5\\
\hline 
\makecell[l]{ALBEF \\(14M+flickr\_finetuned)} & Aircraft   & 1.0 & 5.7 
& 1.8 & \textbf{31.7} & 11.3 &    
3.5 & 60.8 & 21.7 &    
4.0 & 70.9 & 25.2 &    
5.2 & 91.5 & 32.6\\
\hline 
\makecell[l]{BLIP-ViT-B/16 \\ (129M + COCO)}
& Aircraft        & 1.0 & 5.3 &  
1.4 & \textbf{26.6} & 8.3 &    
3.4 & 63.3 & 19.8 &  
5.0 & 94.9 & 29.7 &  
5.1 & 95.5 & 29.9\\
\hline 
ALBEF (14M) & Aircraft   & 1.0 & 3.6 &  
1.2 & \textbf{33.6} & 4.2 &    
2.2 & 61.3 & 7.7 &    
3.1 & 86.6 & 10.9 &  
3.8 & 105.9 & 13.3\\
ALBEF (4M) & Aircraft 
& 1.0 & 2.7 &  
1.0 & \textbf{37.1} & 2.1 &   
1.4 & \textbf{50.6} & 2.8 &     
1.8 & \textbf{66.3} & 3.7 &  
2.3 & 87.6 & 4.9\\
BLIP-ViT-B/16 (129M) & Aircraft & 1.0 & 3.8 
&  1.2 & \textbf{31.7} & 4.6 &   
1.9 & \textbf{51.6} & 7.5 &  
3.8 & 100.0 & 14.5 &   
4.3 & 115.1 & 16.7\\
\hline 
\makecell[l]{ALBEF  \\ (14M + coco\_finetuned)}
& Aircraft    
& 1.0 & 6.1 &  
1.6 & \textbf{27.1} & 11.0 &  
3.8 & 61.6 & 25.0 &  
5.5 & 90.1 & 36.7 &   
5.8 & 95.1 & 38.7\\
\hline 
BLIP-ViT-B/16 (4M)     
& Aircraft   
& 1.0 & 6.6 & 
1.6 & \textbf{23.6} & 11.1 &  
2.9 & 43.2 & 20.2 &   
4.8 & 72.7 & 34.1 &   
6.0 & 90.5 & 42.4\\
\hline 
\makecell[l]{BLIP-ViT-L/16 \\ (129M + Flickr)}
  & Aircraft   
  & 1.0 & 5.9 & 
  1.9 & \textbf{31.8} & 12.4 &  
  2.8 & 47.5 & 18.4 &  
  4.7 & 79.3 & 30.8 &   
  4.9 & 82.3 & 32.0\\
\hline 
\makecell[l]{BLIP-ViT-B/16 \\ \& CapFilt-L (129M)}  
 & Aircraft    
 & 1.0 & 5.0 & 
 1.4 & \textbf{27.5} & 7.6 &  
 2.6 & 52.1 & 14.4 &   
 3.8 & 75.4 & 20.9 &  
 4.3 & 86.2 & 23.9\\
\hline 
\makecell[l]{BLIP-ViT-B/16 \\ (129M + Flickr)}
  & Aircraft   
  & 1.0 & 4.8 & 
  1.8 & \textbf{36.9} & 9.3 &   
  3.7 & 76.2 & 19.3 &  
  4.9 & 101.3 & 25.6 &  
  5.0 & 103.8 & 26.3\\
  \hline 
BLIP-ViT-L/16 (129M)             & Aircraft
& 1.0 & 5.3 & 
2.0 & \textbf{37.6} & 11.9 &  
3.4 & 64.0 & 20.3 & 
5.0 & 94.4 & 29.8 &  
5.8 & 107.9 & 34.1\\
ALBEF (14M)                      & EuroSAT              & 10.0 & 17.4 & 
17.3 & \textbf{99.1} & 66.3 &
15.7 & \textbf{89.7} & 60.1 &  
17.8 & \textbf{102.1} & 68.4 & 
20.7 & \textbf{118.8} & 79.5\\
ALBEF (4M)                       & EuroSAT              & 10.0 & 19.4 &  
7.7 & \textbf{39.8} & 32.9 &  
11.3 & \textbf{58.6} & 48.5 &  
19.0 & \textbf{97.9} & 81.0 &  
20.0 & \textbf{103.5} & 85.6\\
  \bottomrule
  \end{tabular}
  \end{adjustbox}
  \end{center}
\end{table}

\section{CNN vs ViT}
\textbf{\Cref{tab:cnn_vs_vit}:} While early research in multi-modal learning employed both CNN and ViT-based backbones (such as CLIP~\cite{radford2021learning} and OpenCLIP~\cite{ilharco_gabriel_2021_5143773}) -- new SOTA models solely leverage ViTs as their backbone. We explore the effectiveness of CNN (mainly ResNets-based) and ViTs-based backbone within the same model settings while low-resolution shift occurs. Here, we found that ViT-based backbones (such as ViT-B/32, ViT-B/16, and ViT-L/14) are much more robust and lower sensitive to LR shift as compared to CNN-based (such as RN50, RN101, RN50x4, RN50x16, and RN50x64) backbones. In Table~\ref{tab:cnn_vs_vit}, we report the SAR and WAR ($\Gamma^{D}_{n}$) scores of CLIP~\cite{radford2021learning} backbones across 15 datasets for different severity labels.

\begin{table}[!h]
  \caption{\textbf{Robustness analysis of CNN vs ViT-based backbones} of CLIP model across 15 datasets for different severity labels using $\Gamma^{D}_{n} (\uparrow)$.}
  \label{tab:cnn_vs_vit}
  \centering
  \footnotesize
  \setlength\tabcolsep{2pt}
  \begin{tabular}{l|c|c|cc|cc|cc|cc|cc}
    \toprule
    Backbones & \# Params $(\downarrow)$ & $A_{224}$&  
    \multicolumn{2}{c|}{$224\rightarrow128$} & 
    \multicolumn{2}{c|}{$224\rightarrow64$} & 
    \multicolumn{2}{c|}{$224\rightarrow32$} & 
    \multicolumn{2}{c|}{$224\rightarrow16$}
    & \multicolumn{2}{c}{Avg. $(\uparrow)$} \\ 
    & & WAR & SAR & WAR & SAR & WAR & SAR & WAR & SAR & WAR & SAR & WAR\\ 
    \midrule
     RN50 & 102M & 99.90 & 92.54 & 87.89 & 70.16 & 66.00 & 32.75 & 32.52 & 10.00 & 17.20 & 51.36 & 50.90\\
     RN101 & 120M & 99.95 & 94.78 & 92.09 & 75.66 & 70.99 & 39.18 & 38.11 & 10.49 & 13.99 & \textbf{55.03} & \textbf{53.80} \\
     
     \hline 
     RN50x4 & 178M & 99.99 & 92.46 & 88.82 & 70.94 & 66.50 & 34.77 & 30.64 & 10.04 & 11.88 & 52.05 & 49.46 \\  
     \hline

     RN50x16 & 291M & 100 & 91.41 & 85.08 & 73.09 & 64.42 & 37.72 & 32.06 & 10.90 & 14.46 & 53.28 & 48.76\\
     RN50x64 & 623M & 100 & 93.58 & 88.49 & 78.70 & 70.26 & 44.22 & 35.11 & 12.09 & 14.11 & 57.15 & 52.24\\
     \midrule
     ViT-B/16 & 150M & 100 & 96.35 & 93.93 & 83.39 & 77.89 & 53.03 & 44.89 & 21.01 & 19.90 & 63.45 & 59.15\\
     ViT-B/32 & 151M & 99.98 & 96.62 & 94.88 & 82.67 & 77.68 & 52.39 & 44.49 & 19.41 & 16.91 & 62.77 & 58.49\\ 
     ViT-L/14 & 428M & 100 & 97.12 & 95.36 & 87.05 & 80.35 & 63.40 & 51.38 & 25.68 & 18.20 & \textbf{68.31} & \textbf{61.32}\\ 
     ViT-L/14@336px & 428M & 100& 96.08 & 93.74 & 85.68 & 78.22 & 61.11 & 50.65 & 24.42 & 17.81 & 66.82 & 60.10\\ 
    \hline
  \bottomrule
  \end{tabular}
\end{table}

\section{Implementation Details}

\textbf{Dataset Weights:}
In Table~\ref{tab:dataset_weight_value}, we have shown the dataset-specific weight values used to compute weighted aggregated robustness for low-resolution. All models were trained on 2 48GB GPUs. 


\begin{table}[!h]
  \caption{Optimized dataset weight values for WAR-16, shown using pie chart in \Cref{fig:datasets_weights} (right) in the main paper.}
  \label{tab:dataset_weight_value}
  \centering
  \setlength\tabcolsep{2pt}
  \begin{tabular}{l|c|c|c}
    \toprule
    Dataset & Weight & SAR-16 Correlation & WAR-16 Correlation\\ 
    \midrule
    Imagenet & 0.15556157429688613 & 0.99269 & 0.93295 \\ 
    \rowcolor{mygray} 
    ImageNet-A & 0.970498446080589 & 0.55646 & \textbf{0.68070} \\
    ImageNet-V2 & 0.2854574367981364 & 0.99165 & 0.93733 \\ 
    ImageNet-R & 0.01 & 0.98201 & 0.90682 \\
    ImageNet-Sketch & 0.021456095637452655 & 0.95086 & 0.87241 \\
    Caltech101 & 0.01 & 0.97695 & 0.90853 \\
    DTD split-1 & 0.505922498560715 & 0.87676 & 0.82507 \\
    Food101 & 0.01 & 0.97771 & 0.91575 \\
    SUN397 & 0.407563119725743 & 0.98760 & 0.94531 \\
    Stanford Cars & 0.13583821249199218 & 0.96639 & 0.91721 \\
    FGVC Aircraft & 0.8229545014750042 & 0.89746 & 0.89016 \\
    Oxford Pets & 0.08995285864599148 & 0.97224 & 0.90114 \\
    Flowers102 & 0.08972060770047119 & 0.97073 & 0.91809 \\
    \rowcolor{mygray} 
    EuroSAT & 1.0 & 0.25753 & \textbf{0.49229} \\
    UCF101 & 0.01 & 0.97324 & 0.93516 \\

    \hline
  \bottomrule
  \end{tabular}
\end{table}

\noindent \textbf{Super Resolution Method Preprocessing:} 
Here, we present preprocessing steps for two pipelines \emph{i.e.} (i) \textbf{Vanilla Pipeline:} raw image $\rightarrow$ create a low-resolution image using \texttt{transforms.Resize}\:($\cdot$)\:$\rightarrow$ upscale it to the model resolution using \texttt{transforms.Resize}\:($\cdot$)\:$\rightarrow$ input to the model; and (ii) \textbf{Super Resolution Pipeline:} raw image $\rightarrow$ create a low-resolution image using \texttt{transform\_test}\:($\cdot$)$\rightarrow$ pass through Super Resolution models $\rightarrow$ get the model resolution using \texttt{sr\_transform\_test}\:($\cdot$) $\rightarrow$ input to the model. The detailed implementation of these two pipelines is illustrated in the code below:

\begin{lstlisting}[language=Python, caption=SR data preprocessing, label=code:transformation, basicstyle=\ttfamily\footnotesize, keywordstyle=\color{blue}, commentstyle=\color{green!60!black}, stringstyle=\color{red}]

# org_res is the original model resolution
# low_res is the low resolution 
# normalize (mean, std) is the normalization specific to the model

# Pipeline-1: Vanilla
transform_test = transforms.Compose([
    transforms.Resize(low_res,interpolation=InterpolationMode.BICUBIC),
    transforms.Resize(org_res,interpolation=InterpolationMode.BICUBIC),
    transforms.CenterCrop(size=(org_resolution, org_resolution)),
    _convert_image_to_rgb, # converts img to RBG using PIL
    transforms.ToTensor(),
    normalize,
])

EVA_INPUT = transform_test(RGB_IMG)
...
# Pipeline-2: SUPER RESOLUTION
# RAW Image --> SR model 
transform_test = transforms.Compose([
    transforms.Resize(low_res,interpolation=InterpolationMode.BICUBIC),
    transforms.CenterCrop(size=(low_res, low_res)),
    _convert_image_to_rgb,
    transforms.ToTensor(),
    normalize,
])

# SR Image --> EVA model 
mean = (0.48145466, 0.4578275, 0.40821073)
std = (0.26862954, 0.26130258, 0.27577711)
sr_transform_test = transforms.Compose([
    transforms.Resize(org_res,interpolation=InterpolationMode.BICUBIC),
    transforms.CenterCrop(size=(org_res, org_res)),
    _convert_image_to_rgb,
    transforms.ToTensor(),
    transforms.Normalize(mean, std),
])

SR_INPUT = transform_test(RGB_IMG)
SR = SR_MODEL (SR_INPUT)
SR = transforms.functional.to_pil_image(normalize(SR), mode=None)
EVA_INPUT = sr_transform_test(SR)
....

\end{lstlisting}

\noindent \textbf{RobustSAM implementation for Classification:}
We use the official code\footnote{URL: \url{https://robustsam.github.io/}} to replace the mask token with the vision class token. 
Robust SAM is a segmentation model. We remove all its segmentation mask components and mask prediction step. The vision transformer encoder's last block is used instead of the decoder, and all the mask component is stripped away. 
Vanilla Transformer is treated as a teacher. In the student model, the class token is replaced with a \textit{learnable token}. 
This new learnable token is passed through each transformer block. After the first block, we treat this as ``early\_feature" as mentioned in the official github. 
Using RobustSAM denoising trainable modules, we generate `complementary\_features' of these early features. 
After the final block, we use the new learnable token to generate 
`final\_image\_embeddings' using the `self.fourier\_last\_layer\_features\:(image\_embeddings, clear=CLEAR)'.

`robust\_features = complementary\_features + final\_image\_embeddings'.

MSE makes noisy and clear class token and robust features similar.

\noindent \textbf{VPT Implementation:}
VPT is the same as ours, instead of adding on top of spatial tokens, trainable 50 tokens are concatenated to frozen spatial tokens before the first block. 
The decline in the performance at higher resolution indicates the need for introducing tokens at every layer instead of just once at the start. 

Both methods follow the same training environment as our LR-TK0 (multi-training paradigm and diffusion-based images 7k * 30).

\section{More Results}


\subsection{Dataset wise resolution vs. accuracy}

\begin{figure}[!t]
\centering
\subfloat[\centering Top-1 Accuracy $32\x32$]
{
\includegraphics[height=3cm]{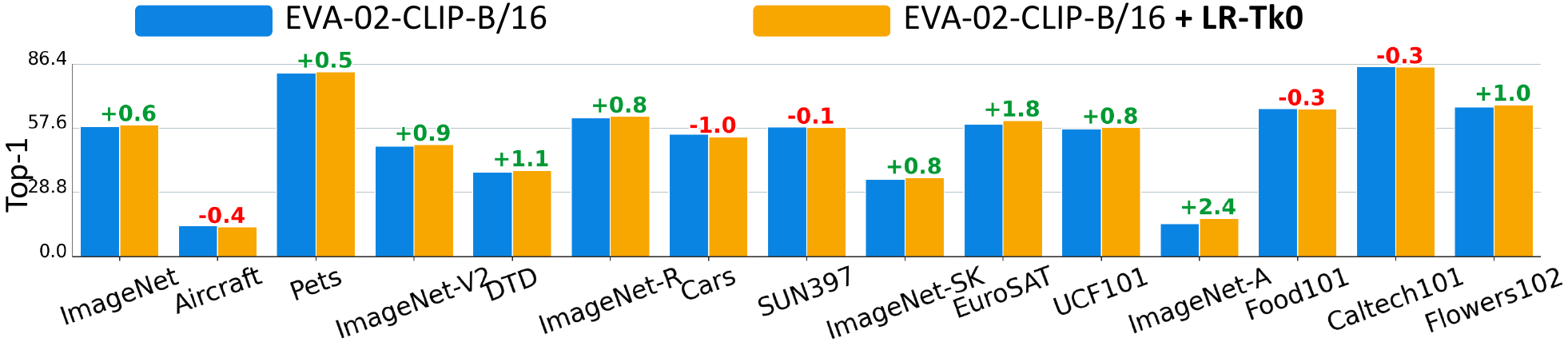}
}
\hfill
\subfloat[\centering Top-1 Accuracy $64\x64$]
{
\includegraphics[height=3cm]{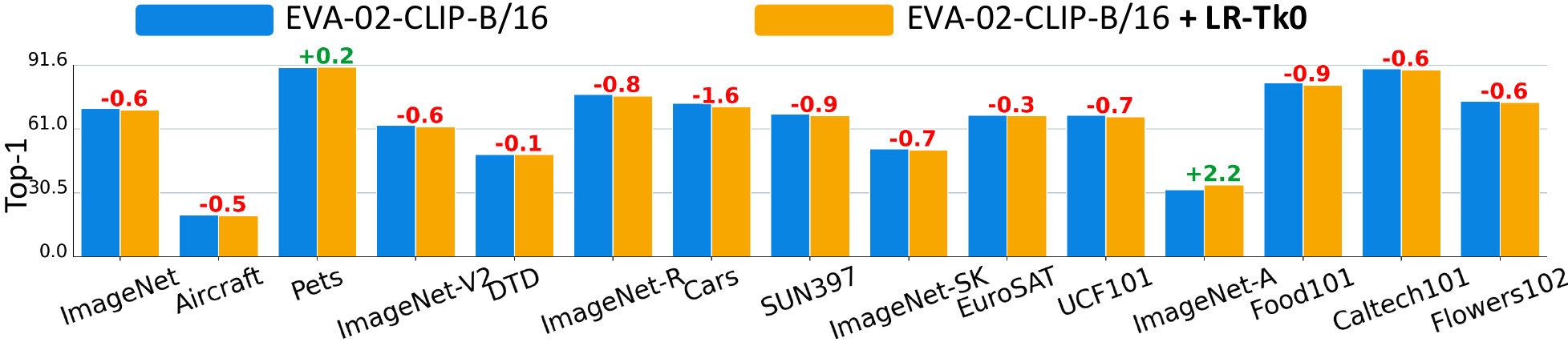}
}
\hfill
\subfloat[\centering Top-1 Accuracy $128\x128$]
{
\includegraphics[height=3cm]{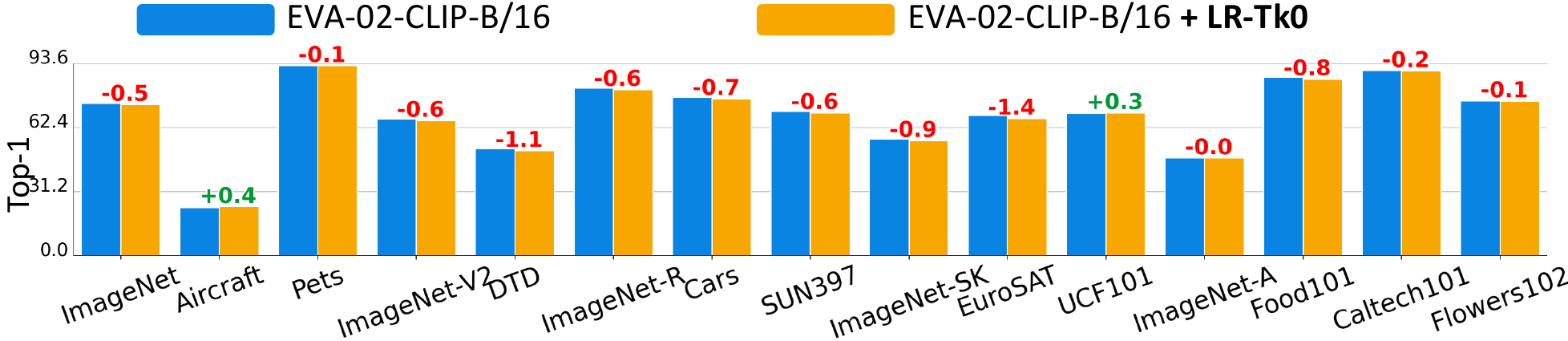}
}
\caption{\textbf{Vanilla vs LR-TK0 (Our)}: 
  Top-1 accuracy for \texttt{EVA-02-CLIP-B/16} model for different resolutions.
}    
\label{fig:dataset_wise_performance}
\end{figure}

In \textbf{\Cref{fig:dataset_wise_performance}}, we highlight the superior zero-shot low-resolution performance (\emph{i.e.} accuracy) of our proposed method, \textbf{LR-TK0}, compared to the vanilla \texttt{EVA-02-CLIP-B/16} model, while utilizing the same backbone across 15 datasets at varying resolutions: $32\times32$, $64\times64$, and $128\times128$. The main paper already demonstrates the results for the $16\times16$ resolution in \Cref{fig:performance_dataset}.

\begin{figure}[!t]
\centering
\subfloat[\centering $\gamma^{D}_{16} (\approx\Gamma^{D}_{16}$) Robustness on $16\x16$]
{
\includegraphics[height=3cm]{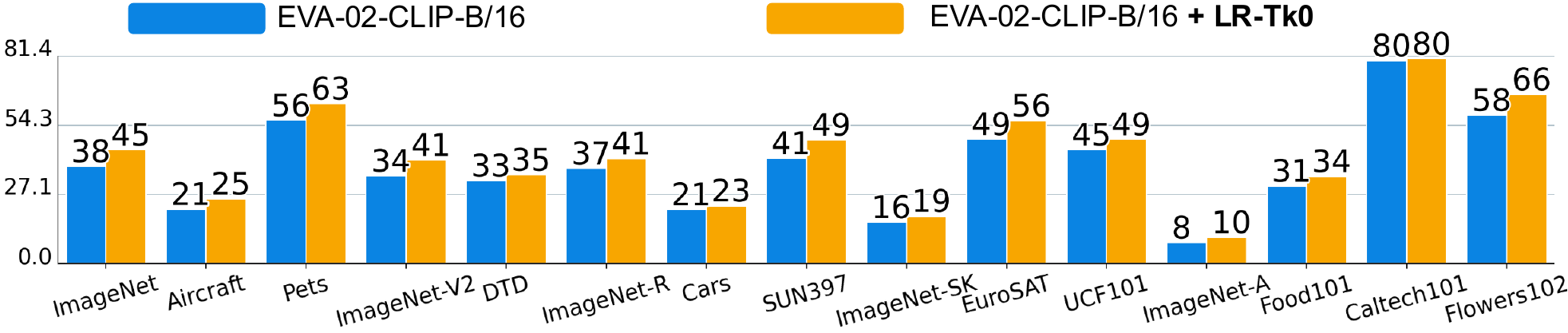}
}
\hfill
\subfloat[\centering $\gamma^{D}_{32} (\approx\Gamma^{D}_{32}$) Robustness on $32\x32$]
{
\includegraphics[height=3cm]{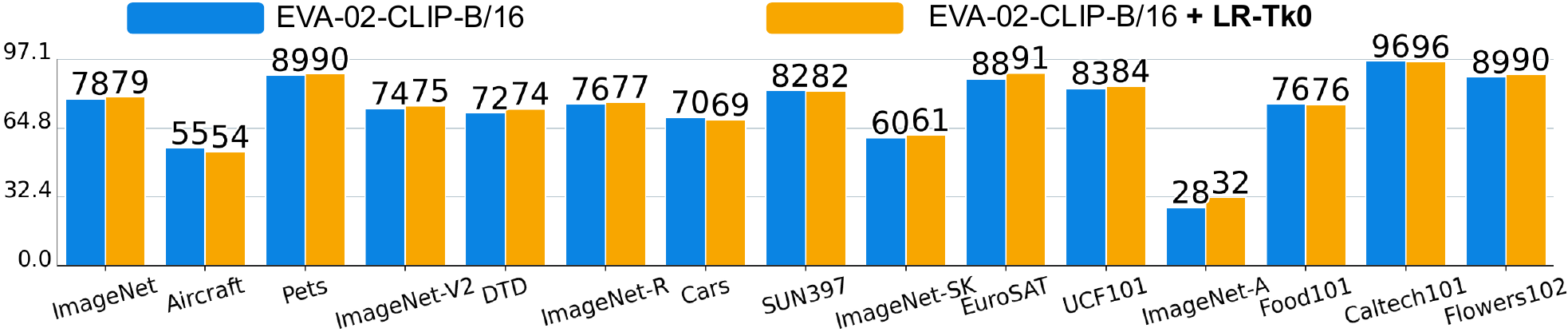}
}
\hfill
\subfloat[\centering $\gamma^{D}_{64} (\approx\Gamma^{D}_{64}$) Robustness on $64\x64$]
{
\includegraphics[height=3cm]{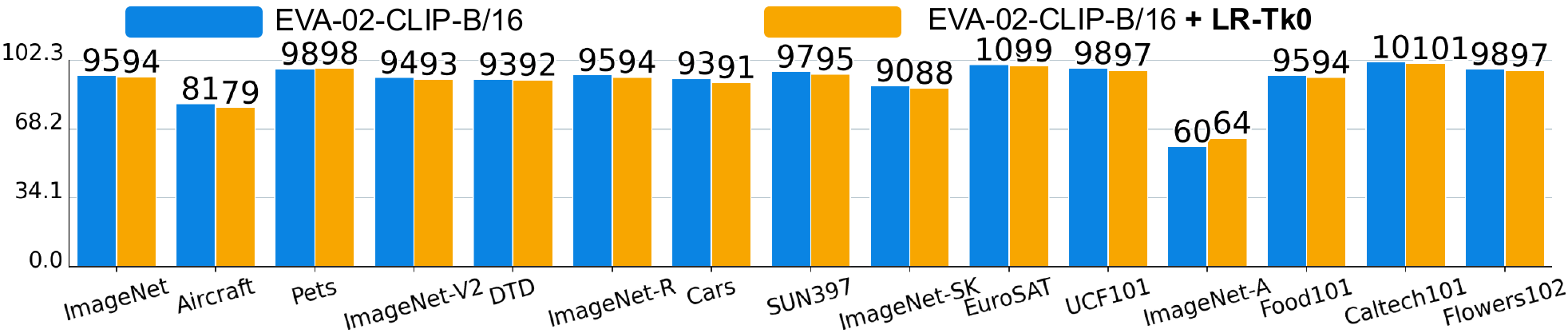}
}
\hfill
\subfloat[\centering $\gamma^{D}_{128} (\approx\Gamma^{D}_{128}$) Robustness on $128\x128$]
{
\includegraphics[height=3cm]{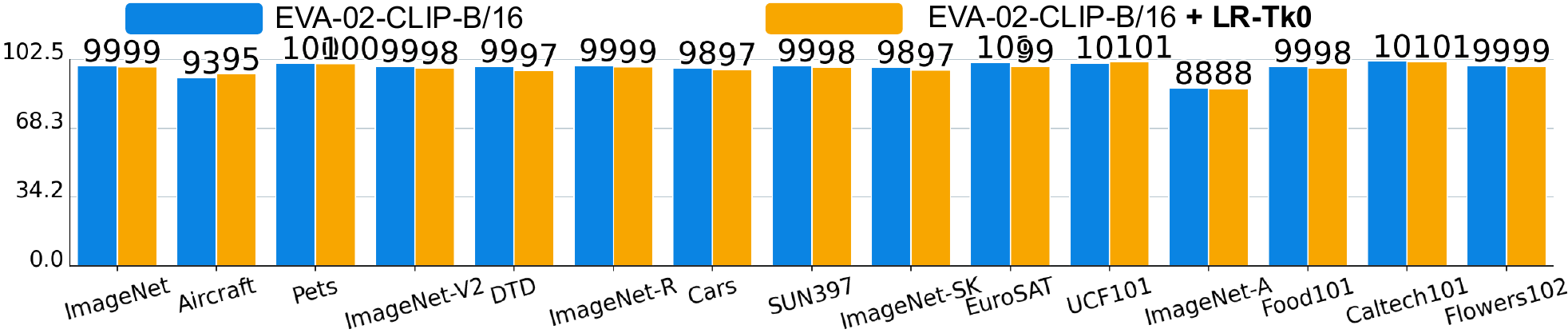}
}
\caption{\textbf{Vanilla vs LR-TK0 (Our)}: 
  Gamma Robustness for \texttt{EVA-02-CLIP-B/16} model for different resolutions on each dataset.
}    
\label{fig:gamma_variation}
\end{figure}

Since EVA performs far superior to random prediction, we present a detailed dataset-specific breakdown of gamma robustness, denoted as $\Gamma^{D}_{n} \approx \gamma^{D}_{n}$ for our proposed method compared with the vanilla \texttt{EVA-02-CLIP-B/16} across resolutions $n = 16, 32, 64,$ and $128$. These results are detailed in \textbf{Figure~\ref{fig:gamma_variation}}. It should be noted that robustness is the absolute value and in Figure~\ref{fig:gamma_variation}, robustness exceeds 100 only when the model's accuracy at lower resolutions surpasses its accuracy at the original 224 resolution.


\begin{table}[!t]
\centering
\caption{\textbf{Comparison with SR}: 
EVA-B/16 results, with different data preprocessing (for SR).
}
\label{tab:comparison_with_SR_supp}
\setlength\tabcolsep{2.2pt}
\renewcommand{\arraystretch}{1.2}
\scalebox{0.98}{
\centering
\begin{tabular}{
l|
c|c|c|    c|c|c|    c|c|c|   c|c|c
}
\specialrule{1pt}{0pt}{0pt}
\rowcolor{mygray} 
& \multicolumn{3}{c|}{$16\x16$}
& \multicolumn{3}{c|}{$32\x32$}
& \multicolumn{3}{c|}{$64\x64$}
& \multicolumn{3}{c}{$128\x128$}
\\[-.4pt] \hhline{~------------}
\rowcolor{mygray} 
\multirow{-2}{*}{Method}
& SAR & WAR & Acc
& SAR & WAR & Acc
& SAR & WAR & Acc
& SAR & WAR & Acc
\\
\hline\hline
Baseline 
& 34.1 & 26.8 & 25.0 
& 71.8 & 59.0 & 51.2 
& \textbf{91.6} & 83.8 & \textbf{63.8 }
&\textbf{ 98.2} & \textbf{95.4} & \textbf{67.6} 
\\ 
\hline
BSRGAN~\SPCITE{zhang2021designing} 
& 12.4 & 12.2 & 8.8    
& 37.3 & 28.7 & 26.9 
& 70.1 & 58.0 & 49.4
& 88.9 & 77.2 & 61.9 
\\
ESRGAN~\SPCITE{wang2018esrgan}
& 14.2 & 15.1 & 10.0 
& 40.3 & 32.6 & 28.9 
& 74.4 & 61.8 & 52.4
& 90.8 & 79.7 & 63.2 
\\ 
Swinir~\SPCITE{liang2021swinir}
& 17.9 & 17.6 & 12.7 
& 47.7 & 38.3 & 34.3
& 79.2 & 68.9 & 55.6 
& 92.7 & 84.6 & 64.2
\\ 
AddSR~\SPCITE{xie2024addsr} 
& 20.5 & 16.8 & 15.0 
& 48.3  & 36.0 &  35.2 
& 73.5 & 57.5 & 52.3 
& 83.6 & 69.4 & 58.7 
\\ 
\hline
\rowcolor{mygray} 
Our 
& \textbf{38.9} & \textbf{29.5} & \textbf{28.4} 
& \textbf{73.1} & \textbf{62.0} & \textbf{52.0}
& 91.4  &  \textbf{85.5 }& 63.6 
&  97.6 & 95.2 & 67.3  
\\
\specialrule{1pt}{0pt}{0pt}
\end{tabular}
}
\end{table}

\subsection{ALL SR Results}

In \Cref{tab:comparison_with_SR_supp}, we present a comparison of our proposed \textbf{LR-TK0} method against the baseline and several state-of-the-art super-resolution methods, including BSRGAN~\cite{zhang2021designing}, ESRGAN~\cite{wang2018esrgan}, SwinIR~\cite{liang2021swinir}, and AddSR~\cite{xie2024addsr}. All super-resolution methods were employed in a zero-shot setting to ensure a fair comparison. Our method significantly outperformed these super-resolution techniques across all resolutions and demonstrated a substantial improvement over the baseline at resolutions of $16\times16$ and $32\times32$. Furthermore, it exhibited comparable robustness at resolutions of $64\times64$ and $128\times128$ with the baseline method.

\subsection{SR results for IDM and Inf-DiT}
\textbf{\Cref{tab:idm_inf_dit}}: 
IDM generalized Zero shot weights do not match their GitHub implementation. Hence we use their weight for cat datasets. 
We evaluate IDM on the pets dataset which is the closest to its pretrained weights. 
For uniformity, we compare Inf-DiT on the pets dataset as well. Both diffusion-based models take around ~4-5 mins per batch of 10 images, making large-scale dataset evaluation impossible. 

\subsection{Grad CAM results}
\begin{figure}[!t]
\centering
\subfloat{
\includegraphics[height=4.5cm]{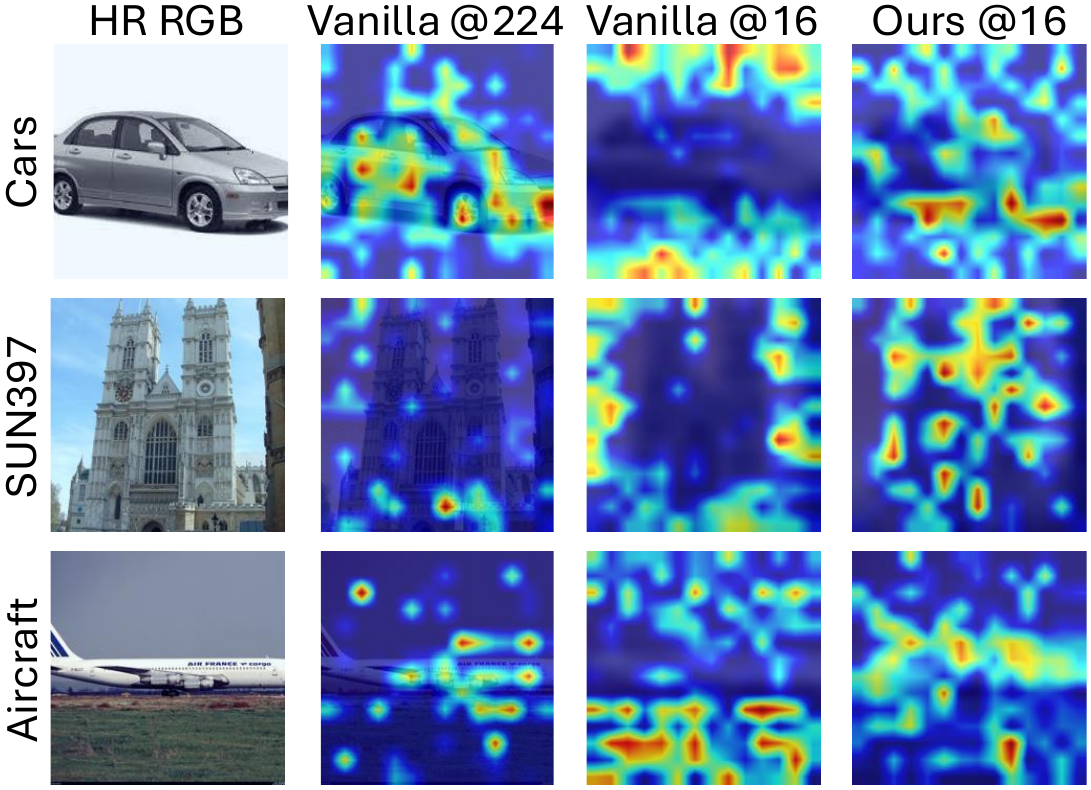}
}
\hfill
\subfloat
{
\includegraphics[height=4.5cm]{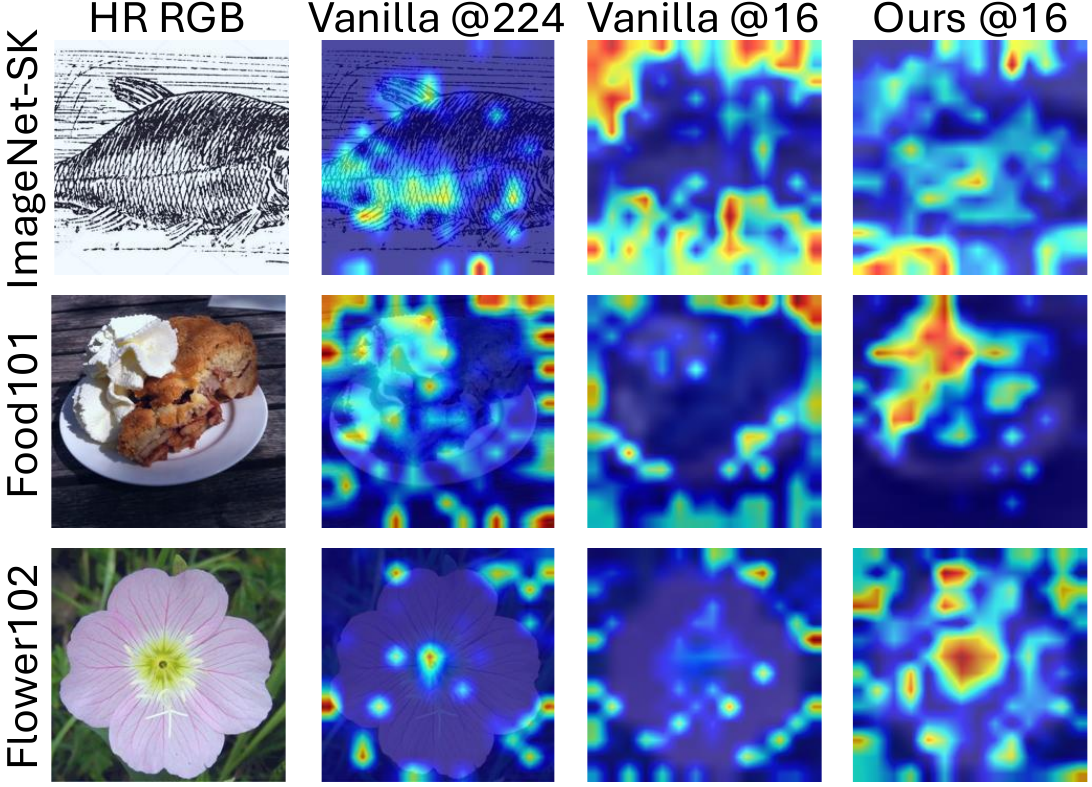}
}
\caption{ 
\textbf{Effect of LR token}: `@' is input resolution. Vanilla model attention is scattered at $16\x16$ (compared to $224\x224$), while our LR tokens focus on the object, capturing fine-grained details.
}
\label{fig:gram_cam_results_supp}
 \vspace{-8pt}
\end{figure}

\textbf{\Cref{fig:gram_cam_results_supp}}, an extension of  \Cref{fig:gram_cam_results} in the main paper, presents the Grad CAM visualization of the vanilla model and proposed method, showcasing the effect of proposed LR tokens.

\begin{table}[!t]
\centering
\renewcommand{\arraystretch}{1.2}
\setlength\tabcolsep{6pt}
\centering
\captionof{table}{
IDM \& Inf-DiT performance on Pets dataset. 
}
\label{tab:idm_inf_dit}  
\scalebox{1}{
\centering
\begin{tabular}{c|c| c| c| c }
\specialrule{1pt}{0pt}{0pt}
\rowcolor{mygray} 
Method &  Top -1 $16\x16$ & Top -5 $16\x16$ & Top-1 $32\x32$ & Top-5 $32\x32$\\
\hline\hline
Eva-B/16 & 51.840	& 84.710 & 82.530 & 98.530 \\ 
Eva-B/16 + LR-Tk0 & 57.92	 & 88.66 & 83.07 & 98.36\\
IDM + Eva-B/16 & 7.2 & 29.03 & 7.88 & 30.25\\ 
Inf-DiT + Eva-B/16 & 29 & 60.94 & 73.43 & 94.36\\ 
\specialrule{1pt}{0pt}{0pt}
\end{tabular}
}
\end{table}

\section{Ablations}

\begin{figure}[!t]
\centering
\subfloat{
\includegraphics[height=4.5cm]{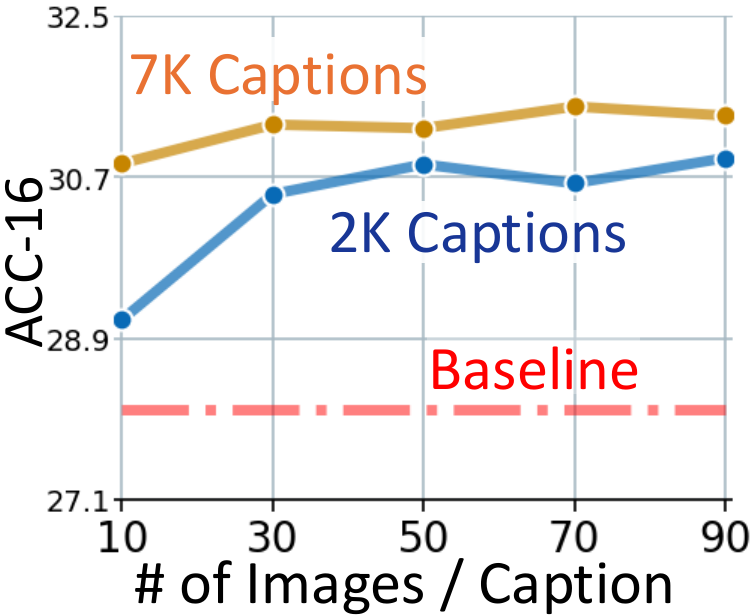}
}
\hfill
\subfloat
{
\includegraphics[height=4.5cm]{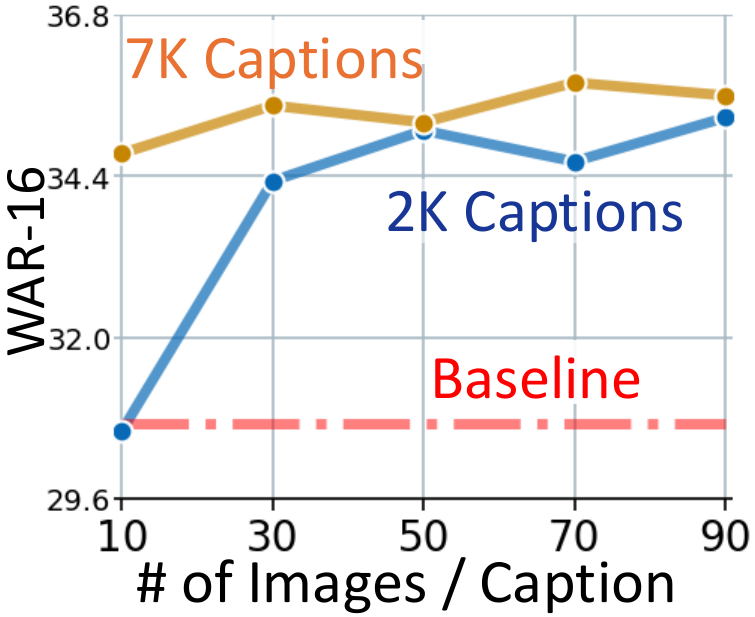}
}
\caption{ 
\textbf{Images/ Caption }: For ACC, and WAR, evaluation metrics on $16\x16$. SAR in the main paper.
}
\label{fig:ldm_gsize_per}
\end{figure}

\noindent \textbf{Number of Images Per Caption:} In the main paper, \Cref{fig:img_per_caption} presents the number of generated (by diffusion model) images (captions) with SAR-16 metric to emphasize how it helps to improve the model robustness. Here, in \textbf{\Cref{fig:ldm_gsize_per}}, we extend this by including ACC-16 and WAR-16 evaluation metrics, while varying the number of generated images.

\begin{figure}[!t]
\centering
\subfloat{
\includegraphics[height=2.8cm]{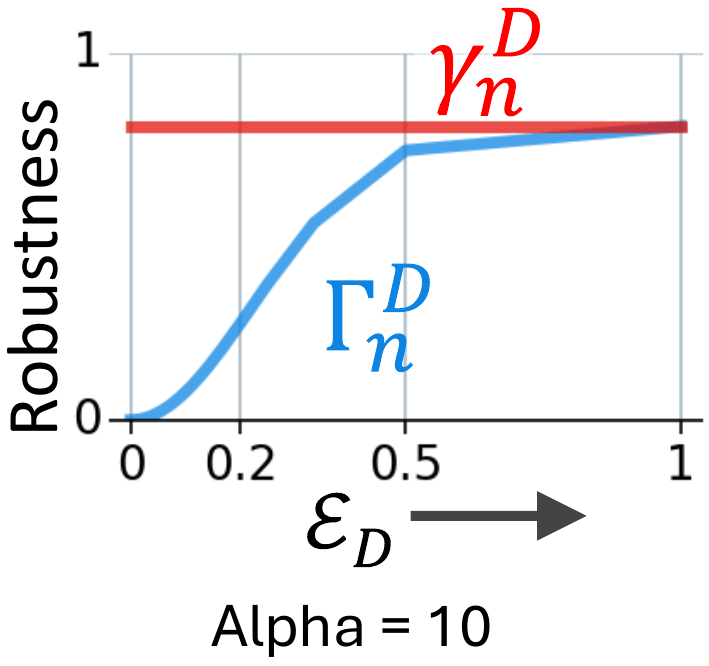}}
\subfloat{
\includegraphics[height=2.8cm]{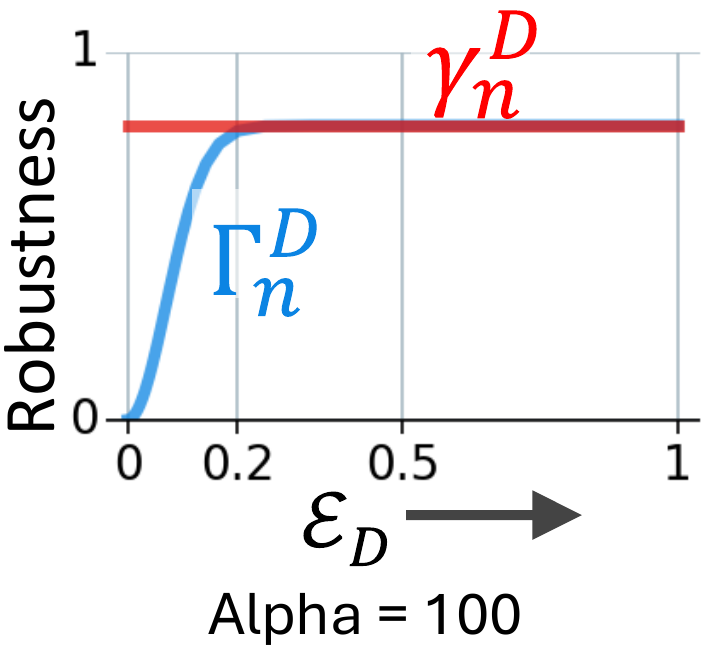}}
\subfloat{
\includegraphics[height=2.8cm]{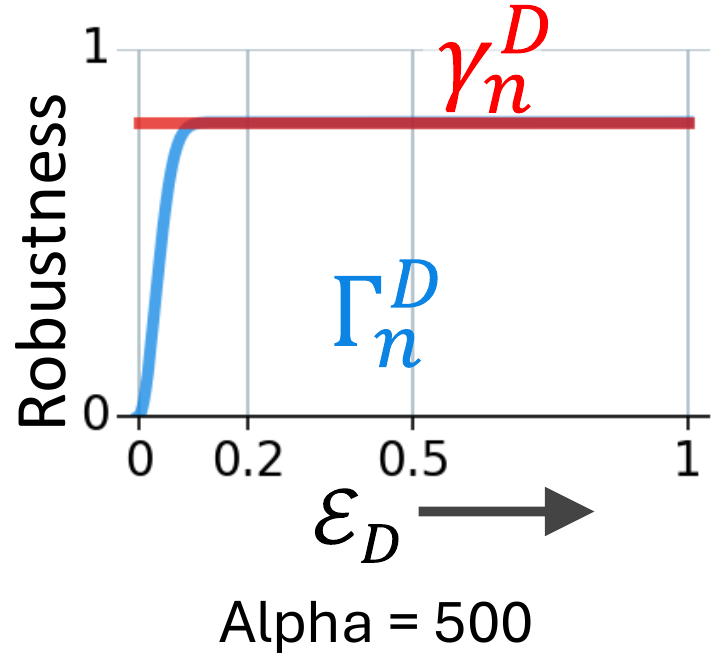}}
\subfloat{
\includegraphics[height=2.8cm]{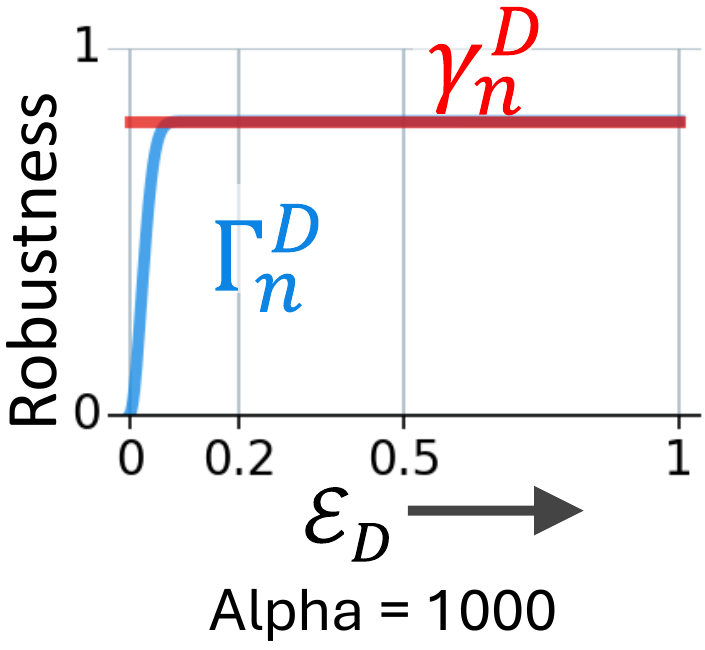}}
\caption{ 
\textbf{Rate of robustness declines as accuracy approaches random prediction.
}
}
\label{fig:alpha_hyperparam_supp}
\end{figure}

\noindent \textbf{Hyperparameter $\mathbf{\alpha}$} signifies the rate of robustness declines as accuracy approaches random prediction. In \textbf{\Cref{fig:alpha_hyperparam_supp}}, we varied the $\alpha$ value with robustness and considered $\alpha=200$ for our experiments as shown in \Cref{fig:hyperparam_war} (left) of the main paper.

\begin{figure}[!t]
\centering
\subfloat{
\includegraphics[height=4.5cm]{Images/position_lr_token-SAR.pdf}
}
\hfill
\subfloat
{
\includegraphics[height=4.5cm]{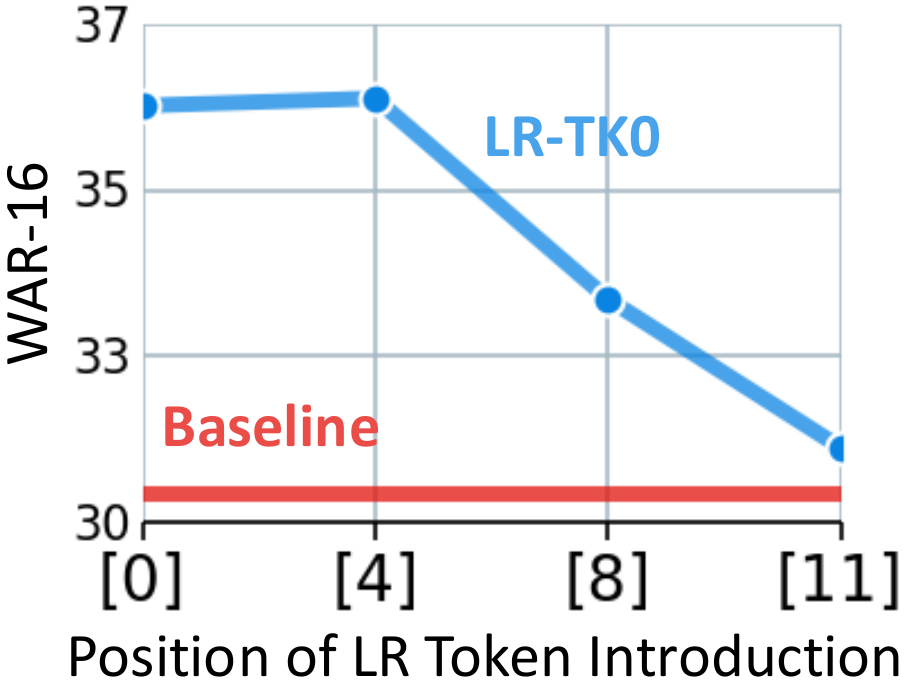}
}
\caption{ 
\textbf{Position of LR tokens introduction}: No tokens were added after the position embedding stage. [i]-th indicates the block from which LR tokens were introduced. Performance metrics variants of \Cref{fig:pos_lr_tokens}.
}
\label{fig:pos_lrtk_supp}
\end{figure}

\noindent \textbf{LR token position:} In the main paper, \Cref{fig:pos_lr_tokens} shows the performance (in terms of SAR-16, 16 is for resolution) with respect to the position of LR tokens being introduced in the form of a line chart. Here, in \textbf{\Cref{tab:pos_lr_tokens}}, we detailed the corresponding numerical values of \Cref{fig:pos_lr_tokens} for better clarity.
Furthermore, we present a side-by-side comparison between the LR token introduction for WAR-16 and SAR-16 metrics in \textbf{\Cref{fig:pos_lrtk_supp}}.

\begin{table}[!t]
\centering
\renewcommand{\arraystretch}{1.2}
\setlength\tabcolsep{2pt}
\centering
\captionof{table}{\textbf{LR token Position (Pos):} 
$[i]$ means LR tokens after $i^{th}$ block (and no token after patchification). 
}
\label{tab:pos_lr_tokens}  
\scalebox{1.0}{
\centering
\begin{tabular}{c|c| c| c| c }
\specialrule{1pt}{0pt}{0pt}
\rowcolor{mygray} 
Pos. &  SAR-16 & WAR-16 & SAR-32 & WAR-32 \\
\hline\hline
$[0]$ & \textbf{42.4} & \textbf{35.4} & 75.3 & 66.4 \\ 
$[5]$ & 41.4 & 35.3 & \textbf{75.4} & \textbf{67.0} \\
$[8]$ & 39.6 & 33.3 & 74.8 & 65.5 \\ 
$[11]$ & 38.4 & 31.3 & 74.5 &  64.6 \\ 
\specialrule{1pt}{0pt}{0pt}
\end{tabular}
}
\end{table}

\begin{figure}[!t]
\centering
\subfloat[\centering Spearman correlation for $32\x32$]
{
\includegraphics[height=5.5cm]{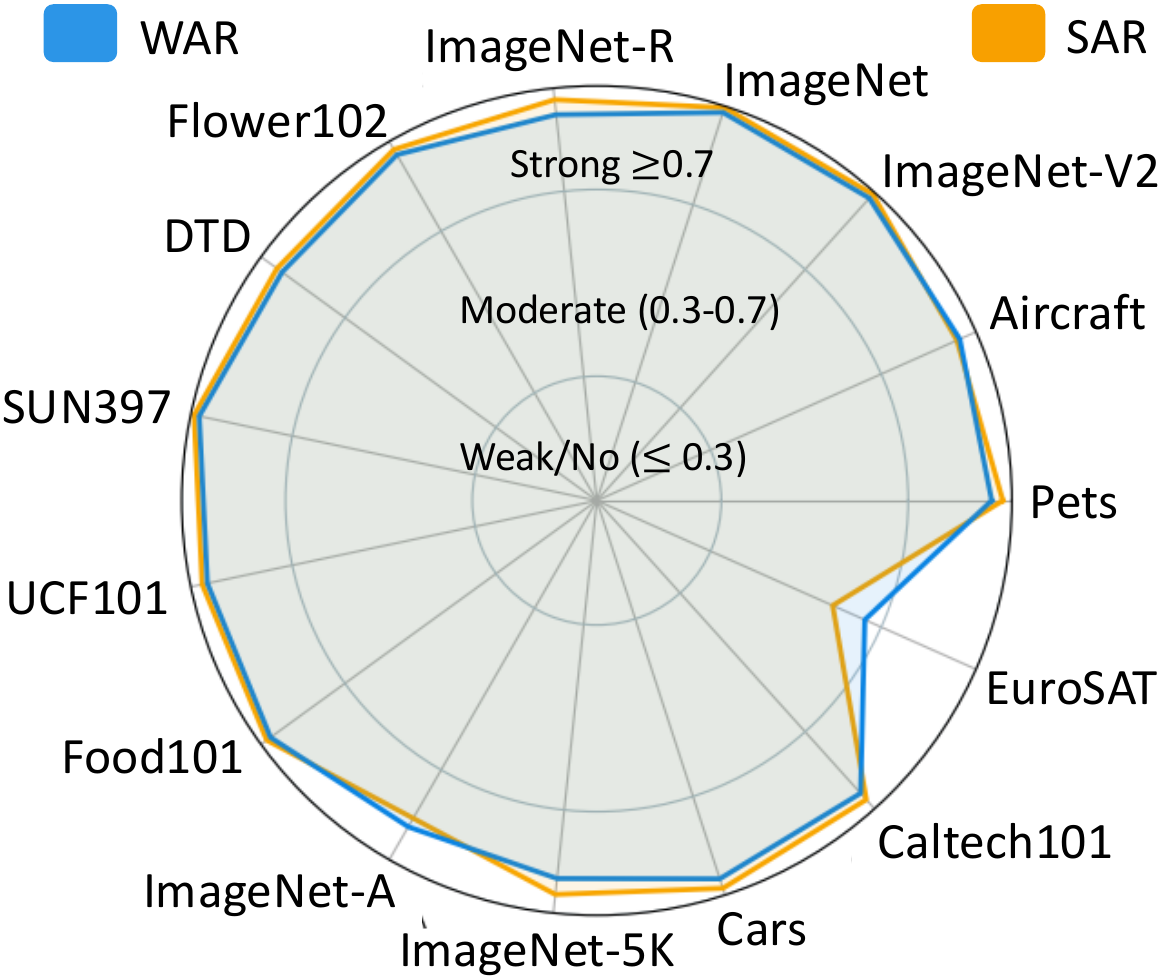}
}
\hfill
\subfloat[\centering Spearman correlation for $64\x64$]
{
\includegraphics[height=5.5cm]{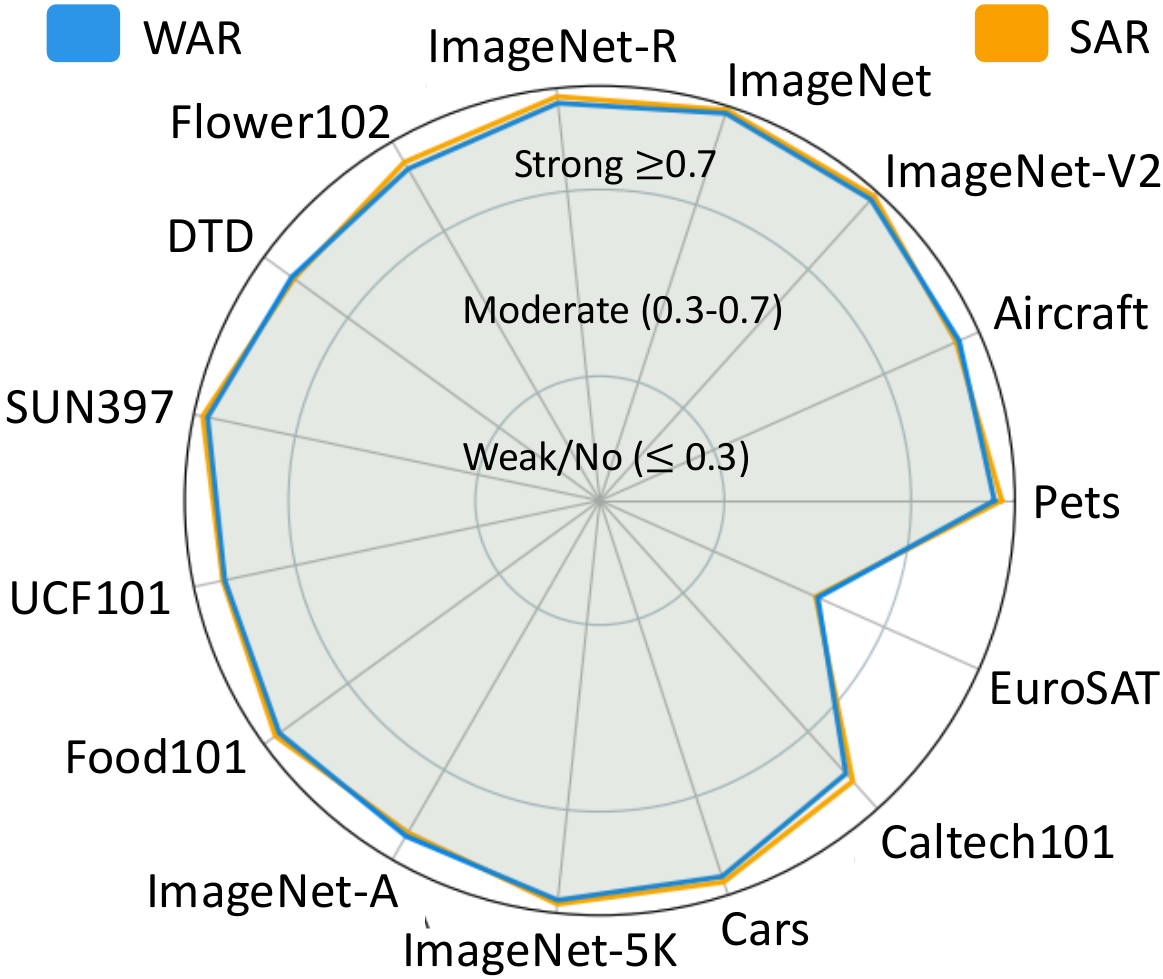}
}
\hfill
\subfloat[\centering Spearman correlation for $128\x128$]
{
\includegraphics[height=5.5cm]{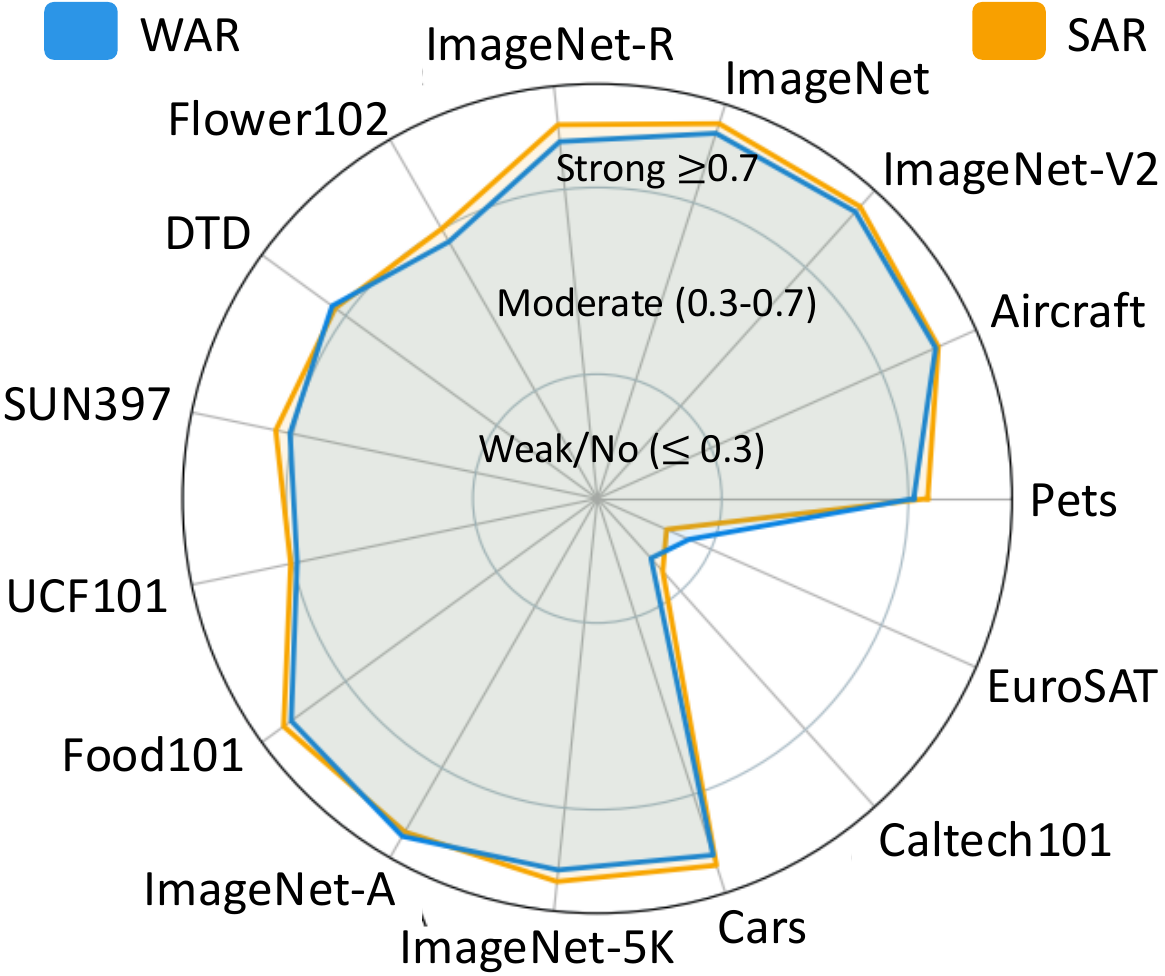}
}
\caption{\textbf{Spearman Correlation} for weights derived for $16\x16$ for higher resolutions.
}    
\label{fig:Sr_images_supp_16}
\end{figure}

\begin{figure}[!t]
\centering
\subfloat[\centering All dataset weights as 1]
{
\includegraphics[height=5.5cm]{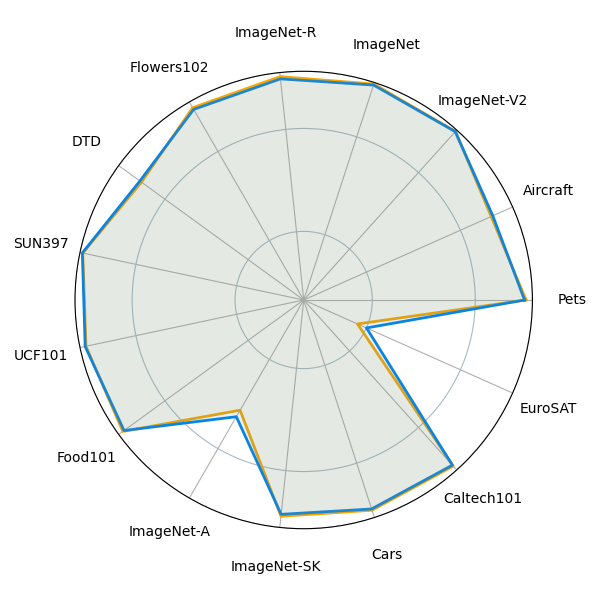}
}
\subfloat[\centering Imagenet: 1, ImageNet-A:1, EuroSAT:1]
{\includegraphics[height=5.5cm]{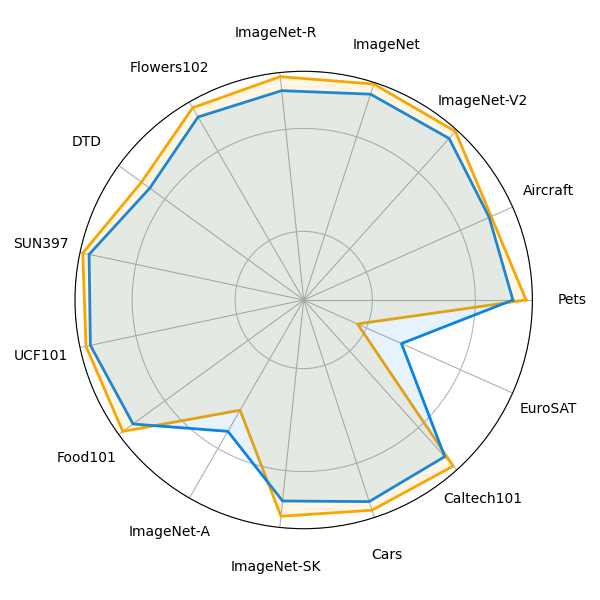}
}\\
\subfloat[\centering Imagenet: 1, EuroSAT:1]
{\includegraphics[height=5.5cm]{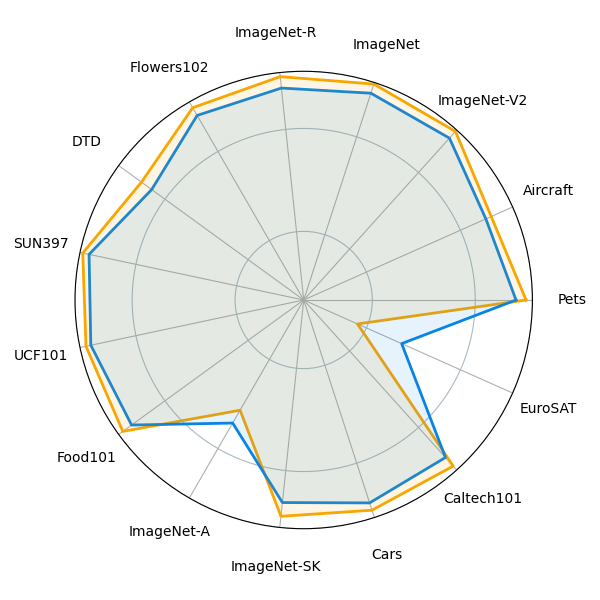}
}
\subfloat[\centering Imagenet: 0.95, EuroSAT:1]
{\includegraphics[height=5.5cm]{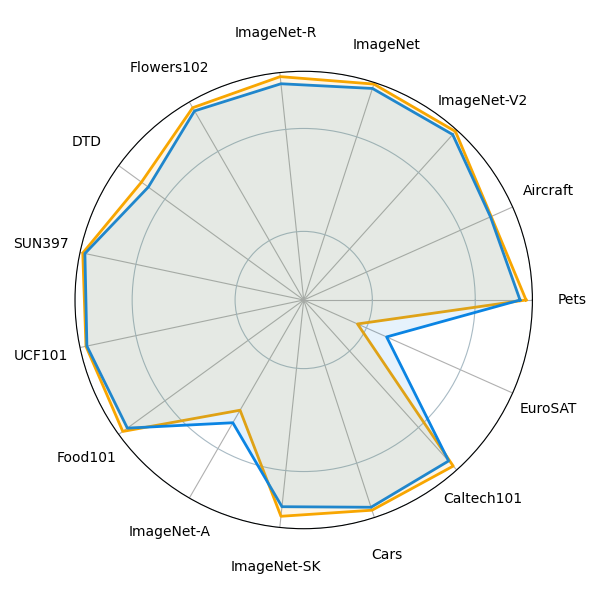}
}
\caption{\textbf{Spearman Correlation} for different optimization function.
The optimization objective is to maximize the mentioned dataset Spearman correlation (SC). 
For Example `Imagenet: 0.95, EuroSAT:1' means: SC (Imagenet) $\x$ 0.95 + SC (EuroSAT) $\x$ 1. 
}    
\label{fig:diff_optimization_configs}
\end{figure}

\noindent \textbf{Spearman correlation for other resolutions:}
In \textbf{\Cref{fig:Sr_images_supp_16}}, weights derived for $16\x16$ are used for other models. Weights for $16\x16$ hold for $32\x32$ but degrade for $64\x64$ and $128\x128$ becoming identical to SAR.
\textbf{\Cref{fig:diff_optimization_configs}} shows different configurations for obtaining dataset weights.

\begin{figure}[!t]
\centering
\subfloat{
\includegraphics[height=4.5cm]{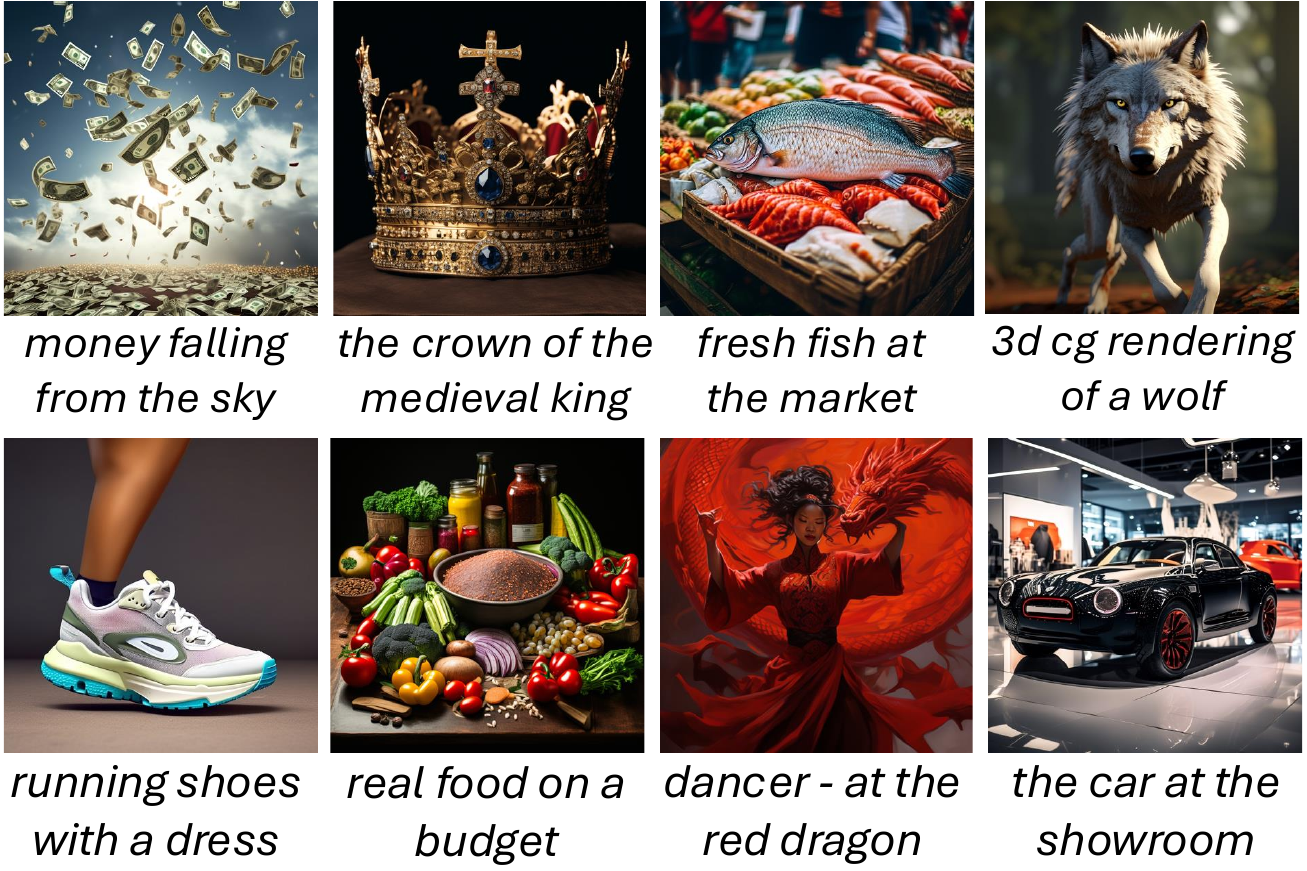}
}
\hfill
\subfloat
{
\includegraphics[height=4.5cm]{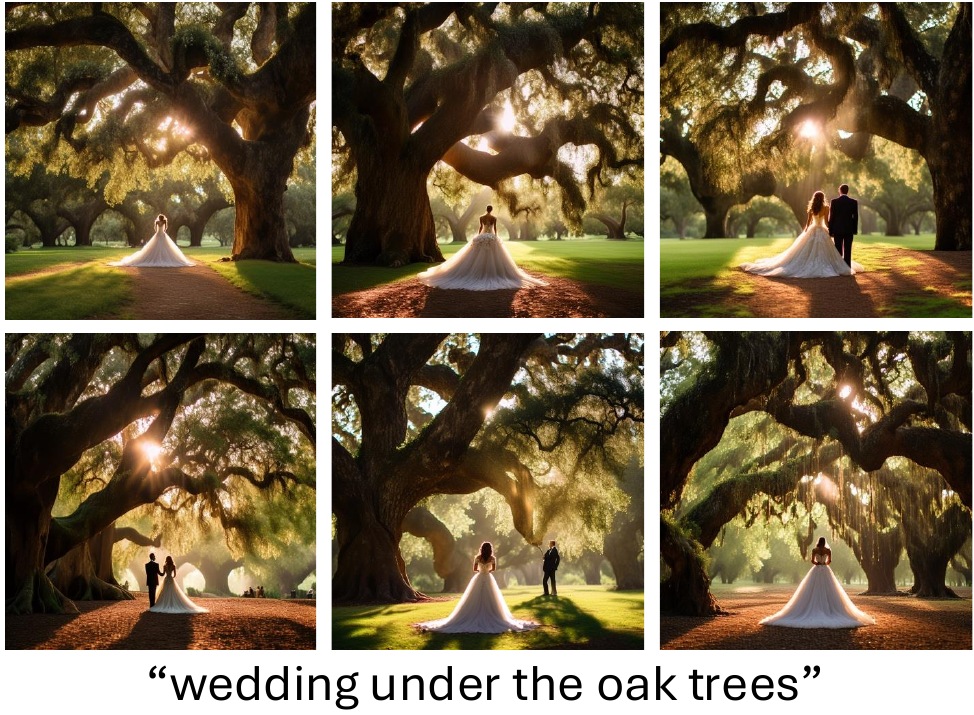}
}
\caption{ 
\textbf{Synthetic Images}: Images generated using PIXART-$\alpha$~\citep{chen2023pixartalpha} using the captions randomly sampled from Conceptual Captions~\citep{sharma2018conceptual}.
\textit{Left}: Sample Images, while \textit{right} shows multiple images per caption generated via different seeds.
More examples of ~\Cref{fig:Sr_images} (in main paper).
}
\label{fig:ldm_images_1}
\vspace{-2pt}
\end{figure}

\begin{figure}[!t]
\centering
\subfloat{
\includegraphics[height=5.5cm]{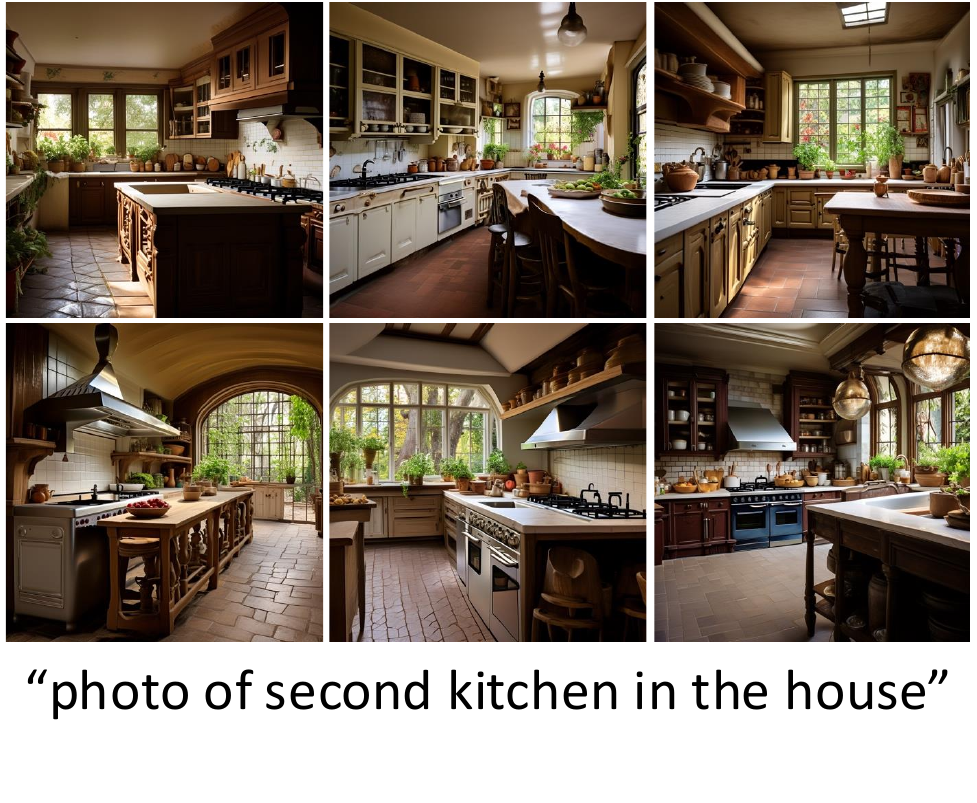}
}
\hfill
\subfloat
{
\includegraphics[height=5.5cm]{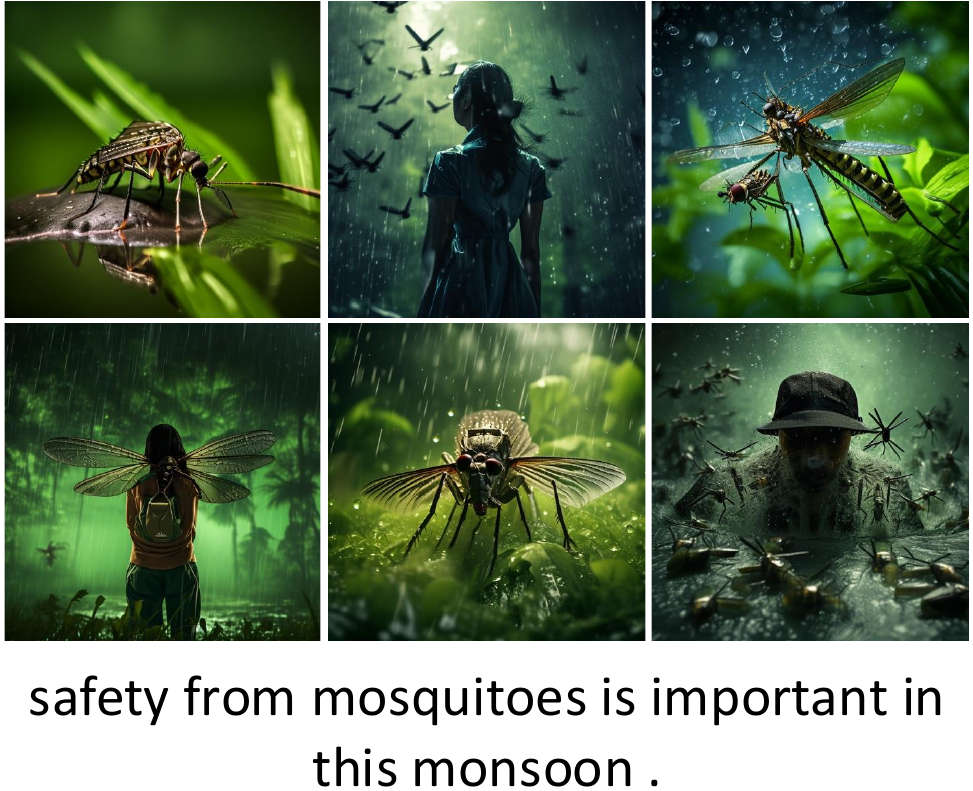}
}
\caption{ 
\textbf{Synthetic Images2}: Mutliple Images / Caption. More examples of ~\Cref{fig:Sr_images} (in the main paper)
}
\label{fig:ldm_images_2}
\vspace{-2pt}
\end{figure}

\noindent \textbf{Samples of Diffusion Generated Images:}
In \textbf{\Cref{fig:ldm_images_1}} and \textbf{\Cref{fig:ldm_images_2}}, we showcase a few sample images generated using PIXART-$\alpha$. These plots are an extension of \Cref{fig:Sr_images} presented in the main paper.

\section{More Observations}

\begin{figure}[!t]
\centering
\subfloat{
\includegraphics[width=0.95\linewidth]{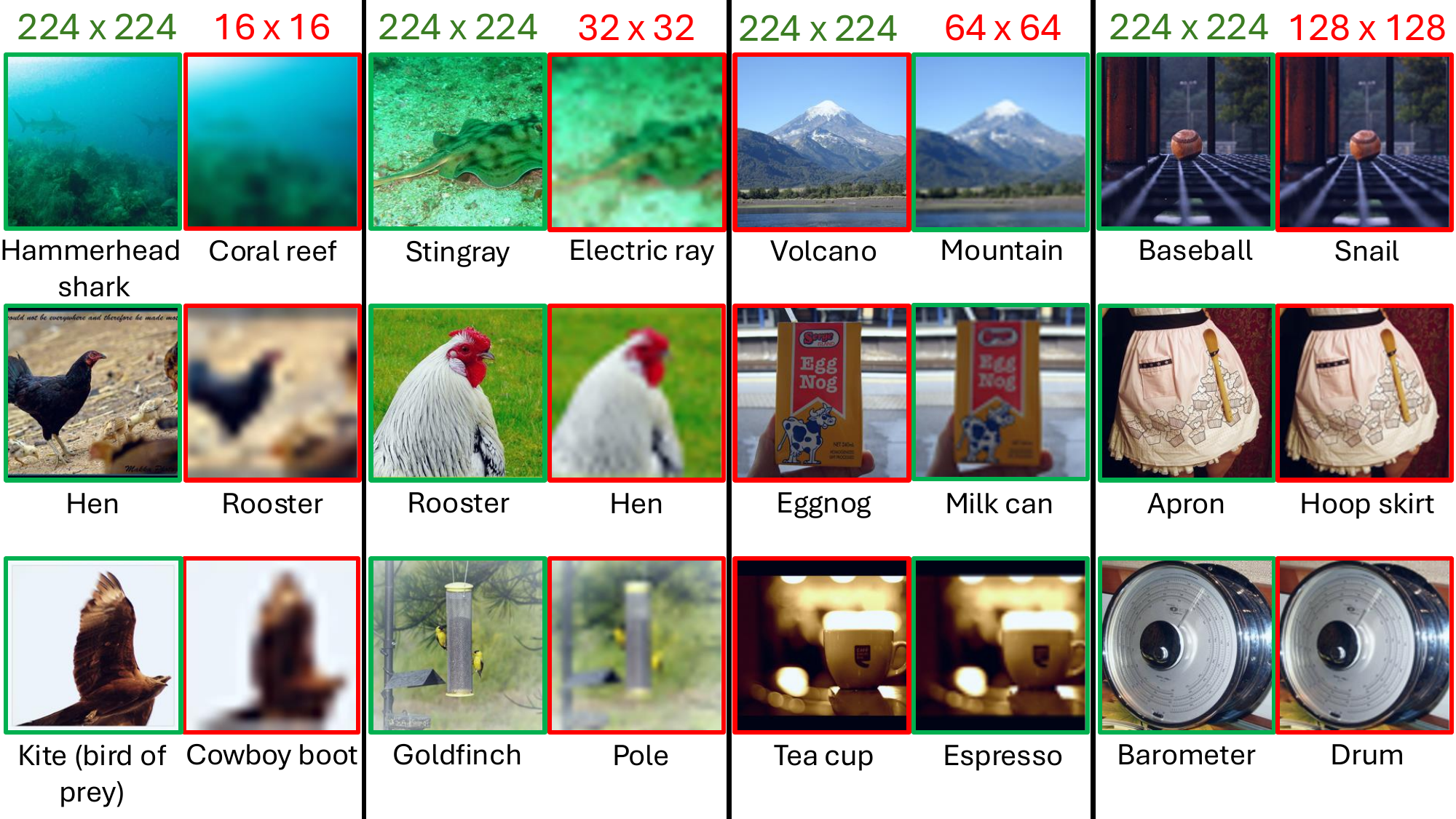}
}
\caption{ 
\textbf{Semantically Correct Predictions}: More examples of \Cref{fig:img_ex} in the main paper. Visually different examples were chosen to show the usefulness of the pre-trained weights even in low resolution. 
}
\label{fig:semantically_resonable_prediction}
\end{figure}

\noindent \textbf{Semantically correct mispredictions:}
As described in \Cref{fig:img_ex} of the main paper, misclassified low-resolution images are still assigned reasonable semantic predictions. Here in \textbf{\Cref{fig:semantically_resonable_prediction}}, we showcase more such examples where the above phenomenon holds.

\begin{figure}[!t]
\centering
\subfloat{
\includegraphics[width=0.95\linewidth]{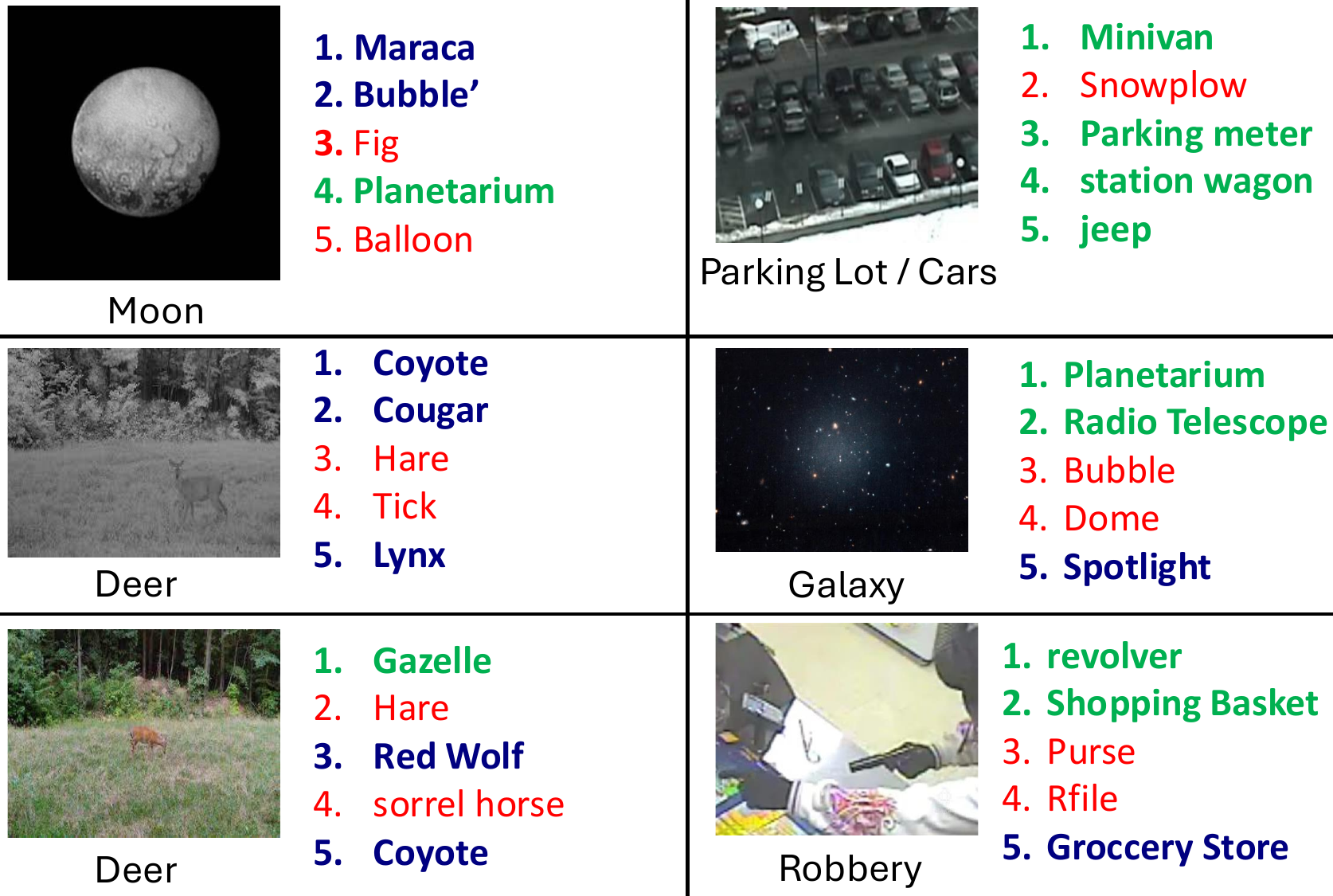}
}
\caption{ 
\textbf{Real World low-resolution images}: Top-5 predictions for images taken from Google (true label shown, below image). \textbf{Blue} indicates Semantically reasonable prediction, \textbf{Green} indicates correct prediction, and \textbf{Red} means wrong prediction. \texttt{EVA-02-CLIP-B/16} model predictions for unknown resolution (real-world footage). Labels/templates are chosen from the \textbf{ImageNet dataset}.
}
\label{fig:semantically_resonable_prediction2}
\end{figure}

\noindent \textbf{Real World low-resolution images:} We have taken a few real-world low-resolution sample images from Google as shown in \textbf{\Cref{fig:semantically_resonable_prediction2}} to see the considered model's performance. Here, we have considered the top-5 predictions of the model and see which indicates (i) correct predictions, (ii) semantically reasonable predictions, and (iii) wrong predictions. The ground labels (or templates) for considered images are chosen from Imagenet.

\section{Limitation}

A key limitation of our study is the lack of detailed analysis of the pre-training datasets, which could provide deeper insights into model performance, particularly regarding how dataset quality impacts robustness. However, due to the scale and unavailability of certain datasets, conducting such an analysis is challenging, though it remains a promising direction for future work with available resources and accessible datasets.

\end{document}